\documentclass[manuscript,nonacm]{acmart}
\acmJournal{TOMACS}
\setcopyright{none}
\acmDOI{}
\usepackage{algorithm,bm}
\usepackage{algorithmicx}
\usepackage{algpseudocode}
\usepackage{comment,makecell,mdframed,enumitem}
\usepackage{booktabs,threeparttable,siunitx,tabularx}
\usepackage{rotating,fancyvrb,tikz,xcolor,subcaption}
\DeclareMathOperator*{\argmax}{arg\,max}

\newcommand{\p}{\mathfrak{prom}}
\definecolor{bluea}{RGB}{192,188,248}
\usepackage{pgfplots}
\pgfplotsset{compat=newest}
\usepgfplotslibrary{polar}
\theoremstyle{acmplain}
\newtheorem{theorem}{Theorem}
\newtheorem{proposition}{Proposition}
\newtheorem{assumption}{Assumption}
\theoremstyle{acmdefinition}

\begin{document}
\title[Budgeted Prompt-Set Expansion and Refinement]{Budgeted Prompt-Set Expansion and Refinement for Language Models via Simulation Optimization}
\author{Haoting Zhang}
\author{Jinghai He}
\author{Rhonda Righter}
\author{Zeyu Zheng}
\affiliation{%
  \department{Department of Industrial Engineering and Operations Research}
  \institution{University of California Berkeley}
  \city{Berkeley}
  \state{California}
  \country{USA}
}
\renewcommand{\shortauthors}{Zhang et al.}
\begin{abstract}
A prompt is an instruction or description provided by the user to guide a generative language model toward outputs that meet application requirements. When these requirements change, existing prompts may need to be adapted and improved. Evaluating candidate prompts requires time and computational resources. We therefore study how to expand and refine an initial set of prompts under a limited evaluation budget through simulation optimization, aiming to maximize a pre-defined task score for a fixed language model. Specifically, we propose a two-stage framework. In the first stage, we expand the initial set of prompts into a finite candidate set, where each prompt is represented by a moderate-dimensional vector. In the subsequent stage of evaluation and selection, we construct a surrogate model of the score for moderate-dimensional vectors representing the prompts. We propose sequentially selecting prompts for evaluation using the constructed surrogate model. We prove the consistency of the sequential evaluation procedure in our framework. We further introduce a refinement procedure that uses the collected scores to search beyond the initial candidate set. We also conduct numerical experiments to demonstrate the advantages of our method over baseline prompt selection methods, with higher prompt scores, especially when the budget for calling language models to evaluate prompts is limited.
\end{abstract}
\maketitle

\section{Introduction}

In recent years, the proliferation of the so-called ``language models", such as the GPT (Generative Pre-trained Transformer) series from \cite{openai2023chatgpt}, has marked a significant advancement in technology and industry \citep{chowdhery2023palm}. Some of most influential language models, such as GPT, are pre-trained so that users can directly use the language models without further training. In this work, when we use the term  \textit{language models}, we refer to the language models that have been already pre-trained and made available to the public, either closed-source or open-source, rather than referring to the training procedures of the language models. 

These language models, largely developed by large technological companies, once made publicly available, can serve as an important tool for smaller businesses and non-profit organizations. Studies have shown that small businesses leverage language models for cost-effective customer service solutions, automating responses, and increasing efficiency with limited resources \citep{de2022increasing}. Non-profit organizations, as highlighted in recent case studies \citep{donorbox2023ai}, have successfully used generative language models to enhance donor engagement and analyze social impact data, leading to more targeted and effective efforts \citep{kanter2022smart}. These examples illustrate the growing relevance and potential of generative language models for smaller-scale operations, highlighting the need for effective use of language models for these organizations.

\begin{figure}
    \centering
    \includegraphics[width=1
\textwidth]{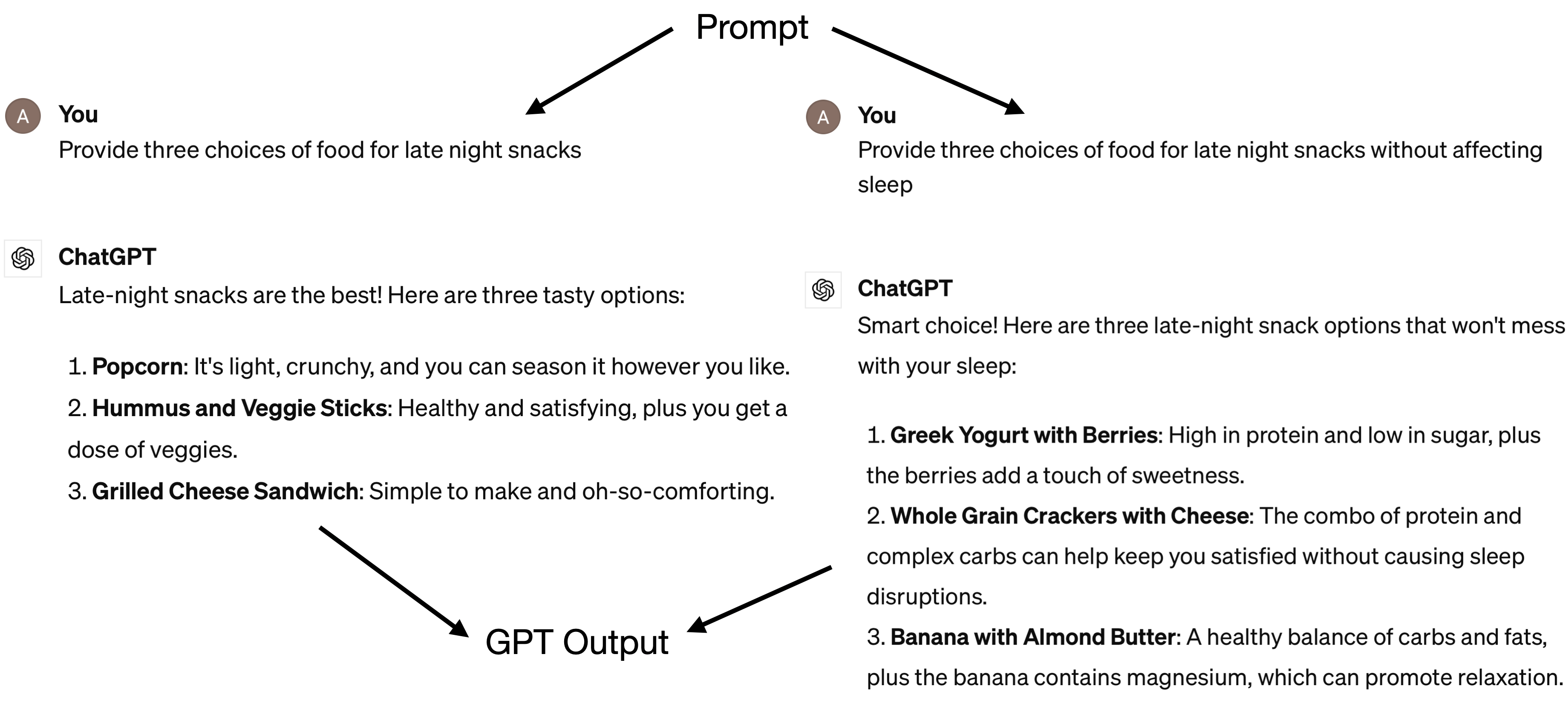}
    \caption{An illustration of different prompts leading to different outputs on a common subject.}
    \label{fig:expro}
    \Description{Two prompt-and-response examples about late-night snacks. Adding the requirement that snacks should not affect sleep changes the three food recommendations.}
\end{figure}

One of the cornerstones for effectively leveraging generative language models in smaller enterprises and non-profit organizations emerges in the selection of prompts, which are user-input instructions to guide the generation of outputs from generative language models \citep{giray2023prompt}. As an example in {Figure \ref{fig:expro}}, the prompt input by the user describes the requirement for the language model to generate the output. The way a prompt is phrased significantly influences the relevance of the context output by the generative language model. Studies \citep{power} have shown that well-crafted prompts can even outperform models specifically fine-tuned for applications. Since smaller enterprises and non-profit organizations can suffer from the prohibitive cost of collecting datasets and the extensive computational resources required for training generative language models, prompt selection emerges as a promising alternative for these organizations with constrained generative language model development capabilities, underscoring the need for research and development in prompt selection. 

{Moreover, in some application scenarios, the underlying tasks recur with modest variations, and existing prompts may need to be adapted and improved when task requirements change. For instance, organizations increasingly rely on time series anomaly detection to monitor business metrics such as revenue, inventory, and customer-support volumes, as reported by some enterprise service companies \citep{Microsoft_AI_Anomaly_Detector, Microsoft_Anomaly_Detector_TN, AWS_Lookout_Anomalies}. Recent work shows that language models can integrate both numerical signals and unstructured context to detect and explain anomalies, with their effectiveness depending sensitively on how the detection task is instructed by the prompts \citep{su2024large,li2024anomaly,gu2025argos}. Another trend in language model applications is solving optimization problems in operational management, such as supply chain inventory management, production planning, and logistics \citep{orlm2024,jiang2025llmopt}. In these scenarios, the management-related problems requiring decision-making are provided as textual descriptions to the language models for seeking solutions. This strategy avoids the need for extensive human effort to translate business context into mathematical problems and solve them, while existing prompts may need to be refined when the business context changes. In addition, organizations routinely maintain proprietary human feedback datasets \citep{boyer_quality_2025,wu2023bloomberggpt}; leveraging these datasets to select and update prompts is an ongoing process rather than a one-shot decision. Taken together with the dependence of language model performance on the prompt, these recurring application scenarios further underscore the need for efficient and cost-effective methods for prompt selection, particularly for smaller enterprises and non-profit organizations with limited resources.

}

The selection and design of prompts for generative language models often rely on human input and subjective guidance \citep{zamfirescu2023johnny,hou2024prompt}. These approaches capitalize on the nuanced understanding and creativity inherent in human cognition, allowing for the creation of prompts that tend to be relevant and useful. For a given application, human input can provide an initial set of prompts that a user seeks to improve. Expanding this set creates additional candidates to consider, while a limited evaluation budget restricts how much can be learned about each. Because the observed scores are noisy, the user must decide which candidates to evaluate and how often.

In this work, we use simulation optimization to study how an initial set of prompts can be expanded and refined under a limited evaluation budget. With the target language model held fixed, we construct candidate prompts and use their observed scores to guide sequential evaluation and subsequent refinement. In general, simulation optimization handles problems in which each sample is costly to obtain (either through an expensive computer simulation of complex systems or from customers/patients in classical or modern clinical trials). Part of the goal of simulation optimization's algorithm design is to save the sampling costs to achieve a certain requirement of accuracy. We next summarize the prompt selection problem addressed in this work and explain how it may be viewed as a simulation optimization problem.

In this work, we describe the process of prompt selection as follows: Initially, a prompt is input into the generative language model along with other relevant input contexts, leading to the generation of an output context. Then a score of the prompt is revealed to quantify the effectiveness of the selected prompt. The calculation of the score employs two components: a pre-defined set of baseline output contexts tailored to the task, and a task-specific score function that compares the generated output with the reference output. Regarding a specified task, paragraph revision, for example, a baseline set comprises pairs of pre- and post-revision paragraphs. These baseline sets can be collected from a variety of sources including academic publications' draft and final versions, edit histories of collaborative writing platforms like Wikipedia, or datasets of student essays with teachers' feedback \citep{toutanova2016dataset,spangher2022newsedits}. For paragraph revision, the score function assesses the similarity between the baseline output and the output generated by the language model \citep{chandrasekaran2021evolution}. A higher score indicates better performance under the chosen task metric. The performance of the prompt is then determined by this score. Thus, when the prompt selection is cast as a simulation optimization problem, 1) the decision variable is prompt in text form, 2) the objective to maximize is the expected score of the prompt, and 3) each prompt evaluation through the language model is treated as a sample from a stochastic system.

Applying simulation optimization to budgeted prompt-set expansion and refinement presents two challenges. First, although an initial set of prompts is available, a broader finite candidate set must be constructed for sequential evaluation. Second, even when the feasible set is restricted to a finite number of prompts, selection under a limited evaluation budget remains challenging. Considering the implicit dependence of the mean score on the prompt, structural properties used by some simulation optimization methods, such as convexity and Lipschitz continuity \citep{eckman2022plausible}, cannot readily be assumed for text prompts. Surrogate models address the lack of transparency in the objective function \citep{xie2020global,wang2023gaussian}. Applying these models to text prompts requires suitable numerical representations through which observations can inform the evaluation of other candidates. Simulation optimization methods are also available for a finite feasible set without inherent structure \citep{hong2021review}. However, as the candidate set grows, repeatedly evaluating each prompt becomes expensive, considering both the computational time to generate contexts and the cost of invoking generative language models. This motivates selective allocation of the evaluation budget and sharing information across candidates.

\subsection{Introduction to Our Method and Results}
In this work, we propose a simulation optimization framework for prompt-set expansion and refinement under a limited evaluation budget. Our framework is composed of two stages. \textbf{1. Search Stage:} This stage uses randomized search to expand an initial set of prompts into a finite candidate set. The procedure begins by utilizing a text autoencoder, a specialized machine learning model for natural language processing, to transform a few initial example prompts into vectors. These vectors serve as numerical representations of the original text prompts, which are generally high-dimensional. Subsequently, these initial vectors are perturbed, leading to the generation of a larger set of prompts, each uniquely represented by its high-dimensional vector. The next step applies principal component analysis to these vectors to reduce the dimensionality and acquire moderate-dimensional vectors, termed ``prompt representations". The set of prompt representations serves as the feasible set for the prompt selection problem. \textbf{2. Evaluation and Selection Stage:} This stage sequentially evaluates the prompt representation decided in the previous search stage. Specifically, we propose constructing a surrogate model of the mean score regarding the moderate-dimensional prompt representation using a Bayesian parametric model. The surrogate model accounts for the uncertainty in the observed scores for prompts, which are assumed to follow a Gaussian distribution. In addition, the surrogate model provides not only the approximated mean score for each prompt representation but also the approximation uncertainty quantification. We then propose an acquisition function based on the surrogate model, accounting for both exploitation and exploration. In this way, the sequential evaluation of the prompt is decided by maximizing the acquisition function. When the new observations of scores are collected, the surrogate model and the acquisition function are updated. %

We further introduce an optional refinement procedure that uses the collected scores to search beyond the finite candidate set.

Our results are summarized as follows:
\begin{enumerate}
    \item We formulate budgeted prompt-set expansion and refinement for a fixed language model as a simulation optimization problem. We propose a framework that constructs a finite candidate set from an initial set of prompts, then allocates evaluations across that set. A subsequent refinement procedure uses the collected scores to search beyond the finite candidate set.

    \item We propose a surrogate-based simulation optimization algorithm for a finite feasible set, where the surrogate model selection is flexible. We design an acquisition function for which the trade-off between exploration and exploitation is explicitly addressed by a user-specified hyperparameter. Under the stated assumptions, we prove asymptotic consistency of the sequential evaluation over a fixed candidate set using a Bayesian parametric model and our proposed acquisition function. Additionally, the optimization of the acquisition function involves approximating its value at each decision variable within the feasible set using simulated samples. Despite the feasible set being finite, this procedure becomes computationally expensive with a large number of decision variables. Therefore, we also consider employing a probabilistic reparameterization method. This method transforms the optimization of the acquisition function from a discrete to a continuous optimization problem. Consequently, the stochastic gradient ascent method is used for optimizing the acquisition function, which is more efficient to implement when the number of decision variables is large. We also prove that the probabilistic reparameterization to speed up the acquisition function optimization does not alter the optimal solution.

\item {We conduct numerical experiments to examine how candidate-set size and evaluation allocation affect the score of the selected prompt, and when refinement improves it. In experiments, we consider tasks including 1) time series anomaly detection, 2) solving optimization problems, and 3) judging human preferences between responses. We compare our framework with baseline methods, including 1) classical simulation optimization algorithms for sequential budget allocations, 2) token-space black-box optimization methods, and 3) prompt engineering methods driven by language models. Comparisons with the baselines identify settings in which our method achieves higher scores under limited evaluation budgets. In addition, we also include the discussions of the limitations of our method.}

\end{enumerate}

Our work may provide the following managerial relevance: First, our framework offers a method for expanding and refining prompt sets under a limited evaluation budget, improving task performance without updating the target language model's weights (which can incur high costs associated with data collection and extensive computational resources). Because of this, our proposed prompt selection approach may be particularly beneficial for small businesses and non-profit organizations aiming to leverage generative language models better. Secondly, while our focus in this work is on prompt-set expansion and refinement, the proposed framework may be adapted to other management applications with the use of language models. For example, language models have been utilized to serve low-income communities by providing accessible, instant `question and answer (Q\&A)' services that offer guidance on rights, healthcare information, and social services. In such applications, our framework could support prompt evaluation against a specified task score. We also take the view that prompt selection and prompt engineering have the potential to enhance the accessibility of operations technology to a broader range of the population.

\subsection{Literature Review}

{

Our work is relevant to prompt engineering, which is an emerging field within the domain of generative language models. The concept involves crafting effective prompts, which serve as the instructions and task descriptions fed into generative language models to acquire desired outputs. As documented in \cite{giray2023prompt} and \cite{power}, it has been extensively observed that the prompt fed into the language model significantly influences the model's output in terms of relevance, creativity, and accuracy. The methodologies of prompt engineering can be classified into two categories: white-box and black-box. White-box prompt engineering considers the scenarios when the language model is fully accessible and primarily involves adding tunable numerical layers to the language model to represent the prompt \citep{shin2020autoprompt}. Given a dataset to evaluate the language model's output, these white-box methods only fine-tune (by gradient descent) the added layers to facilitate prompt engineering \citep{lester2021power,liu2022p,zhou2022learning,liu2024gpt}. In contrast, we focus on black-box language models in our work, which are called via a text-in-text-out user Application Programming Interface (API). In these scenarios, neither 1) tunable layers can be appended to the language model nor 2) the gradients are available. Black-box prompt engineering methods typically first numericalize the human-readable prompt and then optimize over its numerical representation \citep{diao2022black,chen2023instructzero,lin2024use}. Since gradients of language models are unavailable, black-box prompt engineering methods often rely on derivative-free optimization algorithms (e.g., evolutionary strategies) \citep{sun2022black,prasad2023grips}. \cite{cheng2024black} also trains a sequence-to-sequence model as a general-purpose prompt optimizer. In addition, there is a recent trend to employ language models themselves for prompt engineering because of their proven capability of in-context learning to understand task requirements \citep{zhou2022large,yang2023large}.

Within this literature, we study the expansion and refinement of an initial set of prompts under a limited evaluation budget. Our framework couples candidate-set construction with sequential evaluation, using a surrogate model to share score information across candidates and the collected observations to support further refinement.

}

Because each prompt evaluation is costly and yields a noisy score, we approach this problem through simulation optimization. Simulation experiments are widely used to evaluate the performance of complex systems, and application scenarios include but are not limited to finance \citep{gordy2010nested,jiang2020online}, queue management \citep{ibrahim2010delay,ibrahim2012modeling,ata2023drift,ata2023singular}, etc. In addition to evaluating system performance, simulation experiments have also been employed to optimize system performance with different inputs to the simulation models, which is referred to as simulation optimization; see \cite{eckman2023diagnostic,eckman2023simopt}.

When the feasible set contains finite decision variables with no inherent structure defined, the simulation optimization problem is generally cast as ranking and selection (R\&S) \citep{luo2015fully,dong2016three,ni2017efficient,hong2021review,pei2022parallel,wu2022data,hong2022solving,li2023surprising}. In the context of R\&S, prominent methods include but are not limited to indifference zone approaches \citep{fan2016indifference,fan2020distributionally,tsai2023adaptive} and optimal computing budget allocation \citep{he2007opportunity,fu2007simulation}. 

For finite alternatives with contextual information, \cite{cakmak2024contextual} use Gaussian process posteriors to guide sequential budget allocation.

Another stream of literature on simulation optimization focuses on addressing stochastic optimization problems, where the objective function is the expectation of a random function. To estimate this objective function, simulation experiments are conducted to generate samples of the random functions, as discussed in works by \cite{wu2018bayesian,lam2022general}, and \cite{blanchet2022optimal}. In addition to considering the objective function represented by the mean of a random function, simulation optimization also takes into account the risk measures of random functions, such as the quantile and the conditional value at risk, as seen in \cite{deo2023achieving} and \cite{he2023adaptive}. Additionally, simulation experiments are widely employed to estimate the gradient of the objective function, with detailed discussions found in \cite{ahamed2006adaptive,zhu2021constructing,peng2022new}, and \cite{wang2023large}.

For agent-based simulation, \cite{andelfinger2022towards} uses smooth approximations of discrete model elements to enable automatic differentiation and gradient-based optimization.

Our work benefits from surrogate models in simulation \citep{chen2012effects,dong2018unbiased,wang2018adaptive}, which are statistical models employed to approximate the simulation output and alleviate the expensive execution of simulation experiments. Owing to this advantage, surrogate models have wide applications in simulation fields, including simulation input uncertainty analysis \citep{barton2014quantifying,xie2014bayesian} and simulation optimization \citep{l2019gaussian,xie2020global,semelhago2021rapid,hong2021surrogate,wang2023gaussian}.

Local surrogate approaches include STRONG, which combines response-surface methodology with trust-region search \citep{chang2013strong}, and ASTRO-DF, which adapts sampling to balance sampling error and model error \citep{shashaani2018astro}. For high-dimensional discrete simulation optimization, \cite{li2024projected} use projected Gaussian Markov random fields to guide search through lower-dimensional region and solution layers.

\section{Problem Description}
We consider a fixed language model and a user-specified task. Given an initial set of prompts and a limited evaluation budget, the goal is to obtain a prompt with a high expected task score through candidate expansion, sequential evaluation, and refinement. For example, a user may seek a prompt that guides the language model in revising paragraphs. The set of potential prompts is $\tilde{\mathcal{P}} = \left\{\p_1,\p_2,\ldots\right\},$ where $\p_n$ denotes a human-readable prompt in text form (e.g., ``Revise the following paragraph:''). The prompt and the input context (e.g., the paragraph requiring refinements) are fed into the generative language model and the model generates the output as
\begin{equation}
\label{eq.llm}
    \hat{\bm{y}} = \mathtt{LM}\left(\bm{x},\p\right).
\end{equation}
Here, $\bm{x}$ denotes the input context, $\mathtt{LM}$ represents the generative language model, and $\hat{\bm{y}}$ is the output context generated by the generative language model. We note that, given a fixed pair of input contexts and the prompt $\left(\bm{x},\p\right)$, the output context $\hat{\bm{y}}$ is a random object that can exhibit variability in different trials. Additionally, the language model $\mathtt{LM}$ is a black box called via a text-in-text-out user API.

To quantify the performance of a prompt, our framework incorporates two components. Firstly, there is a baseline set, consisting of baseline contexts tailored to the task, providing a standard for comparison. Secondly, a score function is employed to quantitatively assess the quality of the output generated by the generative language model \citep{chandrasekaran2021evolution}. We define the baseline set as $
\mathcal{B} = \left\{\left(\bm{x}_1,\bm{y}_1\right),\left(\bm{x}_2,\bm{y}_2\right),\ldots,\left(\bm{x}_M,\bm{y}_M\right)\right\}.$ Each pair $\left(\bm{x}_m,\bm{y}_m\right)$ represents a baseline input-output context. For example, regarding the task of paragraph revision, $\bm{x}_i$ denotes the initial paragraph before revision and $\bm{y}_i$ denotes the revised paragraph. Such datasets can be collected from academic publications' draft and final versions, edit histories of collaborative writing platforms like Wikipedia, or datasets of student essays with teachers' feedback \citep{toutanova2016dataset,spangher2022newsedits}. To evaluate a prompt, say $\p$, we input $\left(\bm{x}_m,\p\right)$ into the language model $\mathtt{LM}$, generating the output $\hat{\bm{y}}_m$. The score of the prompt $\p$ is calculated by $h\left(\hat{\bm{y}}_m, \bm{y}_m\right)$, where $h(\,\cdot\,,\,\cdot\,)\in \mathbb{R}$ is the employed score function comparing the generated output $\hat{\bm{y}}_m$ with the baseline $\bm{y}_m$. {The selection of the score function is flexible in our framework and depends on the tasks that language models are implementing. For example, if the task is text-based (e.g., paragraph revision), the score function can be the cosine similarity. When the task is anomaly detection, the score function is the F1 score calculated with the binary labels. The description of the score functions is included in the numerical experiments with different tasks.} A higher score indicates better performance on the task, measured by comparing $\hat{\bm{y}}_m$ with the reference $\bm{y}_m$. In this way, given a baseline set $\mathcal{B}$ and a score function $h(\,\cdot\,,\,\cdot\,)$, the performance of a prompt is evaluated by 
\begin{equation*}
\label{eq.promptevaluation}
\begin{aligned}
    \widehat{v}\left(\p\right) = v(\p)+\epsilon(\p).
\end{aligned}
\end{equation*}
Here, $\widehat{v}(\p)$ is the observed score, $v(\p)\doteq\mathbb{E}_{(\bm{x}_m,\bm{y}_m)\sim \mathcal{D}}\left [ h(\mathtt{LM}(\bm{x}_m,\p),\bm{y}_m )\right ] $ is the mean score of the prompt, with $\mathcal{D}$ the uniform distribution over $\mathcal{B}$ and the expectation also taken over language-model randomness, and $\epsilon(\p)$ is the uncertainty when observing the score of $\p$. We note that the uncertainty $\epsilon(\p)$ comes from two aspects: 1) the random selection of a pair of $\left(\bm{x}_m,\bm{y}_m\right)\in \mathcal{B}$ with equal probabilities, and 2) the phenomenon that language model $\mathtt{LM}$ generates different output contexts even when the input context and the prompt are fixed. In this work, we assume the uncertainty $\epsilon$'s are Gaussian random variables that are independent across different evaluations. The same notation applies when an evaluation uses a batch of baseline examples and $h$ is calculated over the batch.

With the mean score of a prompt defined, the selection is formulated as an optimization problem:
\begin{equation}
\label{eq.select}
\begin{aligned}
    \p^* \in \arg\max_{\p\in \mathcal{\tilde{P}}} v(\p).
\end{aligned}
\end{equation}
That is, for a fixed baseline set $\mathcal{B}$, the objective is to maximize the mean score $v\left(\,\cdot\,\right)$. The evaluation budget limits the number of score observations available to guide the search, which begins with candidates expanded from the initial set of prompts. Considering the implicit dependence of the mean score on the selected prompt, we cast the selection (\ref{eq.select}) as a \textit{simulation optimization} problem, regarding each evaluation of a prompt as simulating a sample from a stochastic system. On the other hand, utilizing classical simulation optimization methods for selecting the prompt encounters challenges:
\begin{enumerate}
    \item \textbf{Candidate set construction:} Regarding the prompt selection problem (\ref{eq.select}), a finite candidate set must first be constructed by expanding the available initial set of prompts. %
    
    \item \textbf{Unknown objective structure \& expensive evaluations:} The prompt selection problem remains challenging even when the feasible set is restricted to be finite. Firstly, structural properties of the score as a function of the prompt, such as convexity or Lipschitz continuity, cannot readily be assumed. This poses challenges to simulation optimization methods that rely on such structure \citep{fan2018surrogate,eckman2022plausible}. Secondly, while surrogate-based simulation optimization methods can optimize objective functions lacking explicit structural properties \citep{quan2013simulation,xie2020global,wang2023gaussian}, applying them to prompts requires suitable numerical representations through which observations can inform the evaluation of other candidates. Lastly, in cases where the feasible set includes decision variables without inherent structure, the problem can be cast as a ranking and selection (R\&S) problem \citep{hong2021review}. When the candidate set is large relative to the evaluation budget, obtaining enough observations to distinguish among prompts becomes difficult. In the context of prompt selection, such extensive evaluation is expensive, considering the computational time and costs associated with engaging the generative language model to generate contexts.
\end{enumerate}

\begin{figure}
    \centering
    \includegraphics[width=\textwidth]{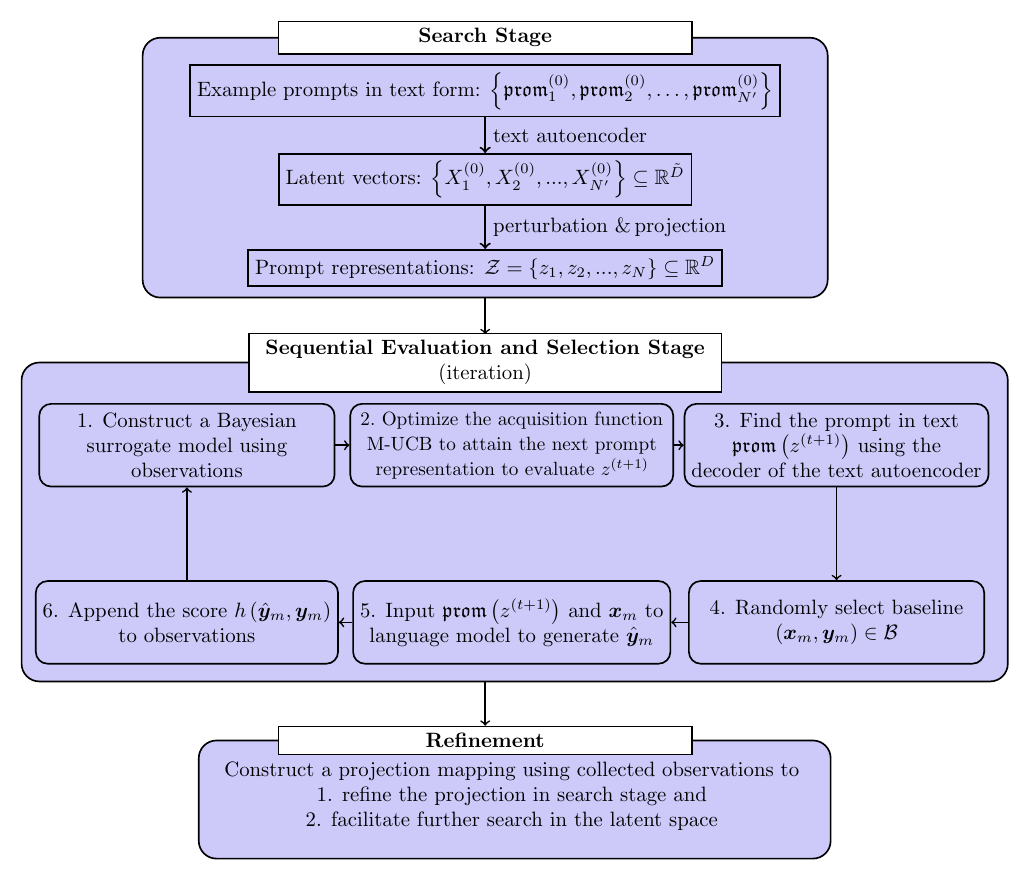}
    \caption{Framework for budgeted prompt-set expansion and refinement.}
    \label{fig:framework}
    \Description{Flowchart of the two-stage framework. Initial text prompts are encoded, perturbed, and projected into prompt representations. Sequential evaluation alternates surrogate updating, acquisition maximization, decoding, language-model evaluation, and score collection. The refinement procedure uses collected scores to adjust the projection and generate further candidates.}
\end{figure}

To address these challenges, we propose a framework for prompt-set expansion and sequential evaluation. The framework is composed of two sequential stages: \textbf{1. the search stage} and \textbf{2. the sequential evaluation and selection stage}. In the initial search stage, we use randomized search to construct a set of candidate prompts to be evaluated and represent these human-readable prompts with real-valued and moderate-dimensional vectors. We first transform a few prompts in text form to high-dimensional vectors in a latent space, using the text autoencoder, a specialized machine learning model that numericalizes texts. We then perturb these vectors to attain a larger set of prompts and apply principal component analysis to reduce the dimensionality of the vectors representing prompts. Then, in the evaluation and selection stage, with a finite set of prompts represented by these moderate-dimensional vectors, we sequentially allocate the evaluation budget across prompts using a Bayesian parametric model. Specifically, we construct a surrogate model using the observed scores regarding vector-represented prompts to approximate the mean score of prompts and propose an acquisition function based on the constructed surrogate model. In each round, we select the prompt to be evaluated by maximizing the acquisition function, which accounts for both exploitation (evaluating prompts that have evidence for high scores) and exploration (evaluating prompts with high uncertainty). When the total budget of prompt evaluation is reached, the prompt that yields the highest mean observed score is selected. We also propose a refinement procedure that uses the observations collected during sequential evaluation and selection. This involves constructing a surrogate model and a projection mapping from the high-dimensional latent space in the initial search stage to a moderate-dimensional subspace. This surrogate model, along with the associated projection mapping, is then used to search for prompts that achieve higher scores than those represented by prompt representations. We present a summary illustration of our framework in {Figure \ref{fig:framework}}. Section \ref{sec.searching} presents the search stage in detail; Section \ref{sec.evaluation} describes the sequential evaluation and selection stage; and the refinement procedure is in Section \ref{sec.refinment}.

\section{Search Stage}
\label{sec.searching}
We describe the search stage of our proposed prompt selection framework. We propose a procedure to expand an initial set of prompts into candidates for the prompt selection problem (\ref{eq.select}) and represent these candidates as moderate-dimensional vectors. The construction uses similarity between prompt texts, without evaluating their task performance with the target language model. We name these moderate-dimensional vectors \textit{prompt representations}. These vectors serve as inputs to the subsequent surrogate model; the target language model receives the corresponding text prompts. In Section \ref{sec.embed}, we describe the procedure to represent a prompt in text form with a high-dimensional vector, which we name \textit{latent vector}. In Section \ref{sec.sampling}, we describe the procedure for constructing a set of prompt representations based on the latent vectors, serving as the feasible set for the prompt selection problem.

\subsection{Prompt Vector Representation}
\label{sec.embed}
Our procedure begins with a few example prompts that have been recognized as plausible for the specified task. For example, in terms of the paragraph revision task to be accomplished by ChatGPT, these example prompts include: 1. ``Please revise the paragraph:'', 2. ``Can you revise the paragraph to be more formal and in the third person? Here's the original paragraph:'', 3. ``Could you please revise the paragraph to fix any grammatical errors and enhance its style? The paragraph is as follows:'', etc. Instead of selecting from a limited number of these example prompts, it is anticipated that other prompts could lead to higher scores for the revision task. This anticipation is supported by existing prompt engineering procedures accomplished by human efforts; see \cite{zamfirescu2023johnny,hou2024prompt}. They found that the performance of prompts will be affected by adding descriptions and adverbials, changing word order, etc. In our framework, the example prompts serve as initial reference points. The scope of the feasible set $\tilde{\mathcal{P}}$ in the prompt selection problem (\ref{eq.select}), extends beyond these initial examples. To expand the initial set of prompts automatically, we transform these prompts into vectors and conduct exploration sampling in the vector space.

Denote the initial set of prompts by $\tilde{\mathcal{P}}^{(0)} = \left\{\p^{(0)}_1,\p^{(0)}_2,\ldots,\p^{(0)}_{N^{(0)}}\right\}$, where $\p^{(0)}_n$ is an example prompt for the specified task and $N^{(0)}$ denotes the number of the example prompts. The choice of representation and decoding methods can be adapted to the application. Here, we employ an auxiliary text autoencoder model \citep{li2015hierarchical} of the form
\begin{equation}
\label{eq.ae}
    \mathtt{AE}(\operatorname{text}_{\text{in}}) = \mathtt{Dec}\left (   \mathtt{Enc}\left (  \operatorname{text}_{\text{in}}\right ) \right ) = \operatorname{text}_{\text{out}}.
\end{equation}
Here $\mathtt{AE}\left(\,\cdot\,\right)$ represents the autoencoder model, which is composed of two parts: an encoder and a decoder. The encoder model $\mathtt{Enc}\left(\,\cdot\,\right)$ is a nonlinear mapping that maps text into a fixed-dimensional real-valued vector, which serves as the bottleneck of the autoencoder. The decoder model $\mathtt{Dec}\left(\,\cdot\,\right)$ takes this vector as input and outputs text. Training aims to reconstruct the input text through this vector representation; approximate reconstruction is sufficient for initializing the search.

In our implementation, we fine-tune both the encoder and decoder of a pre-trained text autoencoder on the initial set of prompts while keeping the target language model fixed; details are given in Appendix \ref{sec.promptimplementation}. After fine-tuning, we encode the initial prompts once and use the resulting latent vectors as starting points for search. The trained decoder is then held fixed and used with deterministic decoding, so that each latent vector maps to a fixed candidate prompt. New candidates are generated by perturbing latent vectors and passing them directly to this decoder, as described in Section \ref{sec.sampling}. We let $X^{(0)}_n\doteq\mathtt{Enc}\left(\p_{n}^{(0)}\right) $ and name it the latent vector. In this manner, the latent vector set is $\mathcal{X}^{(0)} = \left\{X_1^{(0)},X_2^{(0)},\ldots,X^{(0)}_{N^{(0)}}\right\},$ where $X^{(0)}_n = \mathtt{Enc}\left(\p_{n}^{(0)}\right)  \in \mathbb{R}^{\tilde{D}}$. We denote $\tilde{\mathcal{X}}$ as the \textit{latent space}, which is a subset of the $\tilde{D}$-dimensional vector space. In our implementation, we use $\tilde{\mathcal{X}} = \left[-1,1\right]^{\tilde{D}}$.

\subsection{Prompt Representation Set Construction}
\label{sec.sampling}
We now present the procedure of extending the latent vector set and then projecting these vectors to a subspace with a moderate dimension. Recall that we have the set of latent vectors $\mathcal{X}^{(0)} =\left\{X^{(0)}_1,X^{(0)}_2,\ldots,X^{(0)}_{N^{(0)}}\right\}$, where each $X^{(0)}_n\in \mathbb{R}^{\tilde{D}}$ represents a candidate prompt $\mathtt{Dec}\left(X^{(0)}_n\right)$ for the pre-specified task. Instead of directly exploring the language space containing combinations of words and sentences, we search for other prompts by sampling new vectors in the latent space. That is, $\forall X^* \in \mathbb{R}^{\tilde{D}}$, the decoder $\mathtt{Dec}\left(X^*\right)$ will generate a text output, which might serve as a potential prompt. To screen the prompt associated with a latent vector $X^*$, we compare $\mathtt{Dec}\left(X^*\right)$ with a baseline prompt $\p_{b}$. {The comparison is supported by a search band function $g\left(\,\cdot\,,\,\cdot\,\right)$, which quantifies the similarity between two texts. In this manner, we retain the latent vector when $g\left(\mathtt{Dec}\left(X^*\right),\p_{b}\right)\in\left(r_1,r_2\right)$, where $r_1,r_2$ are two user-selected thresholds. In our implementation, $r_1,r_2$ are set to the 10\% and 90\% sample quantiles of the pairwise cosine similarities between the original example prompts, and we use the cosine similarity as the search band function.} The selected thresholds are intended to allow variation in the new prompt $\mathtt{Dec}\left(X^*\right)$ while retaining some similarity to the baseline prompt. Task performance is evaluated in the next stage. We sample new latent vectors based on the initial latent vectors. We first set $\mathcal{X} = \left\{X_1,X_2,\ldots,X_{N^{(0)}}\right\}$, where $X_{n} = X^{(0)}_n$. Then the latent vector set is sequentially extended as follows:
\begin{enumerate}
    \item Calculate a perturbation matrix $\bm{\Sigma}_{p}\in \mathbb{R}^{\tilde{D}\times{\tilde{D}}}$ based on the current set $\mathcal{X} = \left\{X_1,X_2,\ldots,X_{L}\right\}$. In our method, we select the sample covariance matrix for perturbation. That is, $\bm{\Sigma}_{p} = \frac{1}{L} \sum _{n=1}^{L}\left ( X_n-\bar{X} \right ) \left ( X_n-\bar{X} \right ) ^{\top},$ where $\bar{X} = \frac{1}{L} \sum _{n=1}^{L}X_n$.

    \item Randomly select a latent vector $X_l\in \mathcal{X}$ with probability $p_l = \frac{\exp{\left ( -\tilde{n}_l \right ) }}{C_L}$, where $\tilde{n}_l$ denotes the number of times $X_l$ has been selected as a perturbation anchor and $C_L =\sum_{n=1}^{L}\exp{\left ( -\tilde{n}_n \right ) } $ is a normalizing constant. Then a new latent vector $X_{\text{cand}} \sim \mathcal{N}\left( X_l,\bm{\Sigma}_{p}\right)$.

    \item Accept the new latent vector $X_{\text{cand}}$ when 1) $X_{\text{cand}}\in\tilde{\mathcal{X}}=[-1,1]^{\tilde{D}}$ and 2) \[
g\left(\mathtt{Dec}\left(X_{\text{cand}}\right),\p_{b}\right)\in\left(r_1,r_2\right)
\], where $\p_{b} = \mathtt{Dec}\left(X_l\right)$. If the latent vector $X_{\text{cand}}$ is not accepted, go back to Step 1. If the latent vector $X_{\text{cand}}$ is accepted, then append it to the latent vector set $\mathcal{X}$ and go back to Step 1.
    
\end{enumerate}
The procedure terminates when the number of elements in the latent vector set reaches a pre-specified threshold, $N$, and we attain an extended set of latent vectors, $\mathcal{X} = \left\{X_1, X_2, \ldots, X_N\right\}$. As mentioned in Section \ref{sec.embed}, the dimensionality of the latent vectors, \(\tilde{D}\), is generally high, which poses challenges to the subsequent evaluation and selection stage, where we construct a surrogate model of the mean value regarding prompt's numerical representations. High-dimensional vectors can increase the number of evaluations needed to fit the surrogate model, making estimation difficult under a limited budget. Therefore, we apply a dimension reduction algorithm to the latent vector set \(\mathcal{X}\), transforming each \(X_n \in \mathcal{X}\) into a vector \(z_n \in \mathbb{R}^D\) with a moderate dimension \(D\), named as a prompt representation. Specifically, we employ the method of principal component analysis (PCA); see \cite{bro2014principal}. This projection is computed from the latent vectors alone, without task scores.

\begingroup\emergencystretch=3em
Upon completion of dimensionality reduction, we attain a prompt representation set $\mathcal{Z} = \left\{z_1,z_2,\ldots,z_N\right\}\subseteq\mathbb{R}^{D}$. {Each prompt representation $z_n$ is stored together with its original latent vector $X_n$. The corresponding prompt text is obtained as $\mathtt{Dec}\left(X_n\right)$ using the decoder described in Section \ref{sec.embed}.} In this manner, we propose a procedure that constructs a feasible set $\mathcal{Z}$ for the prompt selection problem from the initial example prompts. The constructed feasible set $\mathcal{Z}$ consists of a finite number of candidate representations for subsequent evaluation and selection.\par\endgroup

{We analyze the search stage under a stylized model that represents text similarity by a symmetric function of latent vectors that is continuous in each argument.
\begin{assumption}
\label{assumption.search}
    \begin{enumerate}
        \item The latent space is $\tilde{\mathcal{X}}=[-1,1]^{\tilde{D}}$, with $\tilde{D}\geq 1$. The similarity function $F:\tilde{\mathcal{X}}\times\tilde{\mathcal{X}}\to\mathbb{R}$ is symmetric, $F(x,y)=F(y,x)$, and for every fixed $x$, the map $y\mapsto F(x,y)$ is continuous. For fixed thresholds $r_1<r_2$, each band shell
\[
\mathcal{A}(x):=\left\{y\in\tilde{\mathcal{X}}:r_1<F(x,y)<r_2\right\}
\]
is nonempty.
        \item The perturbation matrix $\bm{\Sigma}_p$ is fixed and positive definite. At proposal step $n+1$, a candidate $Y_{n+1}\sim\mathcal{N}(X_n,\bm{\Sigma}_p)$ is accepted if $Y_{n+1}\in\mathcal{A}(X_n)$; otherwise the state remains at $X_n$. The resulting states form a time-homogeneous Markov chain with transition kernel
\begin{equation}
\label{eq.mckernel}
    K(x,\,\mathrm{d}y)=\mathbf{1}\{y\in\mathcal{A}(x)\}\varphi_{\bm{\Sigma}_p}(y-x)\,\mathrm{d}y+\left(1-\int_{\mathcal{A}(x)}\varphi_{\bm{\Sigma}_p}(u-x)\,\mathrm{d}u\right)\delta_x(\mathrm{d}y),
\end{equation}
where $\varphi_{\bm{\Sigma}_p}$ is the centered Gaussian density with covariance $\bm{\Sigma}_p$.
    \end{enumerate}
\end{assumption}
For a fixed initial state $X_0=x_0\in\tilde{\mathcal{X}}$, define the reachable set by
\begin{equation}\label{eq.set}
\mathcal{X}_{\infty}(x_0):=\left\{x\in\tilde{\mathcal{X}}:\forall\varepsilon>0,\ \mathbb{P}_{x_0}\left(\exists t\geq 0,\ X_t\in\mathcal{B}_{\varepsilon}(x)\cap\tilde{\mathcal{X}}\right)>0\right\},
\end{equation}
where $\mathcal{B}_{\varepsilon}(x)$ is the open ball of radius $\varepsilon$ centered at $x$, and $\mathbb{P}_{x_0}$ is the probability law given $X_0=x_0$. We write $\mathcal{X}_{\infty}$ when the initial state is understood.

In this model, each proposal perturbs the current state using a fixed covariance. The original sampling procedure instead selects an anchor from the accumulated latent vectors and updates the covariance. The following result concerns the simplified procedure in Assumption \ref{assumption.search}.
\begin{proposition}
\label{proposition.search}
    Under Assumption \ref{assumption.search}, for every initial state $x_0\in\tilde{\mathcal{X}}$, the reachable set is $\mathcal{X}_{\infty}(x_0)=\tilde{\mathcal{X}}$. Both the sequence of states $\{X_n\}_{n\geq 0}$ and the subsequence at accepted-proposal times are dense in $\tilde{\mathcal{X}}$ almost surely.
\end{proposition}
The proof is provided in Appendix \ref{sec.proof_search}.

Proposition \ref{proposition.search} describes the asymptotic coverage of the simplified search procedure. The reachable set $\mathcal{X}_{\infty}$ serves as a feasible set within the latent space. This feasible set is high-dimensional, making it challenging to optimize over, given the black-box nature and cost of language models. Thus, we consider a finite set of moderate-dimensional prompt representations $\mathcal{Z}$. In Theorem \ref{thm.infinite} of Section \ref{sec.evaluation}, we study the consistency of optimization over this finite set. The subsequent limit as the candidate set grows is addressed in Theorem \ref{thm.end2end} under its stated assumptions.

}

\section{Evaluation and Selection Stage}
\label{sec.evaluation}
In this section, we describe the procedure of allocating evaluations across the constructed candidate set and selecting a prompt for the fixed language model. The prompts to be evaluated and selected from are represented by moderate-dimensional vectors, termed prompt representations, $z_n\in\mathcal{Z} = \left\{z_1,z_2,\ldots,z_N\right\}\subseteq\mathbb{R}^{D}$, where $N$ is the number of prompt representations and $D$ is the dimension of the vector-valued prompt representation. Each prompt representation $z_n$ is derived from a latent vector $X_n\in\mathcal{X}\subseteq\mathbb{R}^{\tilde{D}}$ by a dimension reduction procedure as in Section \ref{sec.sampling}. We store pairs of latent vectors and prompt representations $\left(X_n,z_n\right)$ so that each prompt representation $z_n$ is associated with its latent vector $X_n$, and the latent vector is transformed into prompt text by applying the decoder $\mathtt{Dec}\left(X_n\right)$. In this way, each prompt representation represents a candidate prompt denoted by $\p\left(z_n\right)$.

To observe a score of the prompt representation, our framework involves two components: 1) a baseline set \[
\mathcal{B} = \left\{\left(\bm{x}_1,\bm{y}_1\right),\left(\bm{x}_2,\bm{y}_2\right),\ldots,\left(\bm{x}_M,\bm{y}_M\right)\right\}
\], where each $\left(\bm{x}_m,\bm{y}_m\right)$ denotes a pair of baseline input-output contexts for comparison; and 2) a score function $h\left(\,\cdot\,,\,\cdot\,\right)$ to assess the generated output against the reference output for the task. In this way, given a prompt representation $z_n$, we attain the candidate prompt $\p\left(z_n\right)$ in text form. We then randomly select a pair of baseline input-output contexts $\left(\bm{x}_m,\bm{y}_m\right)$ from the baseline set $\mathcal{B}$ with equal probabilities. After feeding the prompt $\p\left(z_n\right)$ and the input context $\bm{x}_m$ to the language model, we have the generated output $\hat{\bm{y}}_m\left(z_n\right) = \mathtt{LM}\left(\bm{x}_m,\p\left(z_n\right)\right)$ as in (\ref{eq.llm}), where $\mathtt{LM}$ represents the generative language model. By comparing the generated output $\hat{\bm{y}}_m\left(z_n\right)$ and baseline output context $\bm{y}_m$ by the score function, we then observe a score  
\begin{equation}
\label{eq.generating}
\begin{aligned}   \widehat{v}_{n,m} = h\left(\hat{\bm{y}}_m\left(z_n\right),\bm{y}_m\right) = v\left(\p\left(z_n\right)\right) +\epsilon_{n,m}.
\end{aligned}
\end{equation}
Here, $v\left(\p\left(z_n\right)\right) = \mathbb{E}\left [ \widehat{v}_{n,m}  \right ] $ is the mean score of the prompt associated with the prompt representation $z_n$, which serves as the objective function to be maximized. Since the stored pair $(X_n,z_n)$ associates each prompt representation with a candidate prompt $\p\left(z_n\right)$, we let $v\left(z_n\right)\doteq v\left(\p\left(z_n\right)\right)$ for notational simplicity. In addition, $\epsilon_{n,m}\sim \mathcal{N}\left(0, \sigma_{n}^2\right)$ represents the uncertainty contained in the observed score, which is independent across evaluations and identically distributed across trials for each fixed $z_n$.%

To search for a prompt with a high mean score $v\left(z_n\right)$ under a limited evaluation budget, we propose a sequential selection procedure that employs a Bayesian parametric model as the surrogate model for the mean score with respect to prompt representations. Specifically, the procedure first selects some prompt representations to observe the scores during the warm-up step (Appendix \ref{sec.warm}). The observed scores are used to construct a surrogate model (Section \ref{sec.model}) and an associated acquisition function (Section \ref{sec.acquisition}). Then, during the sequential evaluation step, the next prompt representation to be evaluated is determined by maximizing the acquisition function. After observing new scores, the surrogate model and the acquisition function are updated and used to guide the selection of the next prompt to be evaluated. The procedure terminates when the budget of evaluating prompts $T$ runs out.

In the warm-up step, we evaluate a subset $\mathcal{Z}_W\subset\mathcal{Z}$, using the prompts corresponding to the initial examples as a practical choice, and collect $R$ observations for each. These observations initialize the surrogate model and provide estimates of the observation variances $\sigma_n^2$ across the candidate set. The warm-up cost $T_W=|\mathcal{Z}_W|R$ is included in the total budget $T$. We treat the estimated variances as fixed inputs to the subsequent model and ignore their estimation error. Warm-up selection and variance estimation are described in Appendix \ref{sec.warm}.

\subsection{Bayesian Parametric Surrogate Model}
\label{sec.model}
We assume that the mean performance $v$ with respect to the prompt representation $z$ is represented by a parametric model $v(z) = \bm{f}(z;\bm{W})$, where the form of the model $\bm{f}(\,\cdot\,;\,\cdot\,)$ is known and $\bm{W}\in \mathbb{R}^{p}$ is an unknown parameter. In this work, the selection of the parametric model is flexible, and we propose a procedure to sequentially select the prompt representation $z_n$ to evaluate using Bayesian inference. Specifically, we regard the unknown parameter as a random vector with prior distribution $\pi(\bm{W})$. Suppose we have evaluated the prompts for $t$ rounds (including the observations in the warm-up step), and have collected the observed scores $\mathcal{S}_t \doteq \left\{\left(z^{(1)},\widehat{v}^{(1)}\right), \ldots, \left(z^{(t)},\widehat{v}^{(t)}\right)\right\}.$ Here $z^{(\tau)}\in \mathcal{Z} = \left\{z_1,z_2,\ldots,z_N\right\}$ denotes the selected prompt representation in the $\tau$-th round, and $\widehat{v}^{(\tau)}$ is the observed score associated with $z^{(\tau)}$ as in (\ref{eq.generating}). We note that an equivalent representation of the dataset is $\mathcal{S}_t =\left \{ \left ( z_n,\widehat{v}_{n,m}  \right )  \right \} $ for $m \in\left\{ 1, 2,\ldots,r_n(t)\right\}$ and $n\in\left\{1,2,\ldots,N\right\}$. Here $\widehat{v}_{n,m}$ denotes the $m$-th observed score for prompt representation $z_n$ as in (\ref{eq.generating}), and $r_n(t)$ denotes the number of evaluations at $z_n$ up to time $t$. That is, $\sum_{n=1}^N r_n(t) = t$. Conditional on the dataset $\mathcal{S}_t$, the posterior distribution of $\bm{W}$ is updated by
\begin{equation}
\label{eq.posterior}
    p\left(\bm{W} \mid \mathcal{S}_t\right)  = \frac{p\left(\mathcal{S}_t\mid \bm{W}\right)\pi\left ( \bm{W} \right )}{p\left(\mathcal{S}_t\right) } \propto p\left(\mathcal{S}_t\mid \bm{W}\right)\pi\left ( \bm{W} \right ),
\end{equation}
where 
\begin{equation*}
    p\left(\mathcal{S}_t\mid \bm{W}\right) = \prod_{n=1}^N \prod_{m=1}^{r_n(t)} \frac{1}{\sqrt{2 \pi \sigma_n^2}} \exp \left(-\frac{\left(\widehat{v}_{n, m}-\bm{f}\left(z_n ; \bm{W}\right)\right)^2}{2 \sigma_n^2}\right)
\end{equation*} is the likelihood of the observed scores. In this way, inference for the mean score function is based on $\bm{f}\left(z_n;\widehat{\bm{W}}\right),\widehat{\bm{W}}\sim p\left(\bm{W} \mid \mathcal{S}_t\right)$. For parametric models in which the exact posterior is intractable, Markov Chain Monte Carlo (MCMC) methods \citep{asmussen2007stochastic} or variational inference (VI) \citep{blei2017variational} can be employed to approximate and generate samples from the posterior distribution. Posterior computation is described in Appendix \ref{sec.mcimplementation}.

Finite-rank Gaussian processes and Bayesian neural networks are examples of the Bayesian parametric models used in this framework. The surrogate model is chosen according to the task and the available observations. Model constructions and additional examples are given in Appendix \ref{sec.surrogates}.

\subsection{Acquisition Function \& Consistency}
\label{sec.acquisition}
In this section, we present the acquisition function in the sequential evaluation step, which is used to guide the selection of the next prompt representation $z^{(t+1)}$ to be evaluated. Based on the surrogate model in each round $\bm{f}\left(z;\widehat{\bm{W}}\right),\widehat{\bm{W}}\sim p\left(\bm{W}\mid\mathcal{S}_t\right)$, we propose an acquisition function named \textit{Modified Upper Confidence Bound (M-UCB)}, defined on $z_n\in \mathcal{Z} = \left\{z_1,z_2,\ldots,z_N\right\}$,
\begin{equation}
\label{eq.acquisition}
    \alpha_t\left(z_n\right) =\mu_{t}\left ( z_n \right ) +\beta_t\left(\sigma_{t}\left(z_n\right)+\gamma\left(r_{n}\left(t\right)\right)\right).
\end{equation}
Here $\mu_{t}\left ( z_n \right ) = \mathbb{E}\left[\bm{f}\left(z_n;\widehat{\bm{W}}\right)\mid \mathcal{S}_t\right]$ and $\sigma_t\left(z_n\right) = \left\{\operatorname{Var}\left[\bm{f}\left(z_n;\widehat{\bm{W}}\right)\mid \mathcal{S}_t\right]\right\}^{1/2}$ are the posterior mean and standard deviation of the surrogate model at $z_n$, and $r_n(t)$ denotes the number of evaluations at $z_n$ up to time $t$. In addition, $\left\{\beta_\tau\in \mathbb{R}\right\}_{\tau =1,2,\ldots,t}$ is a user-selected non-decreasing sequence, and $\gamma(\,\cdot\,)\in \mathbb{R}$ is a pre-specified decreasing function defined on $\mathbb{N}= \{0,1,2,\ldots\}$ satisfying $\lim_{n \to \infty} \gamma\left(n\right) = 0$. The selection of $\left\{\beta_\tau\in \mathbb{R}\right\}_{\tau =1,2,\ldots,t}$ and $\gamma(\,\cdot\,)$ is discussed in Appendix \ref{sec.hyperparameters}.

The prompt representation to be evaluated in the next round is selected by maximizing the acquisition function. That is, $z^{(t+1)} \in \arg\max_{z_n\in \mathcal{Z} }\alpha_{t}\left ( z_n \right ),$ where $z^{(t+1)}$ is the prompt representation to be evaluated in the next round. After observing the score $\widehat{v}^{(t+1)}$, the surrogate model is updated via updating the posterior distribution (\ref{eq.posterior}) with $\mathcal{S}_{t+1} = \left\{\left(z^{(t+1)},\widehat{v}^{(t+1)}\right)\right\}\cup\mathcal{S}_t$. The sequential selection procedure iterates until the number of rounds $t$ meets the total budget $T$. The procedure of the entire evaluation and selection stage is summarized in {Algorithm \ref{alg.1}}.

\begin{algorithm}[ht!]
\caption{General Procedure of the Evaluation and Selection Stage with M-UCB}
\label{alg.1}
\begin{algorithmic}[1]
\Require{
The feasible set of prompts $\mathcal{Z} = \left\{z_1,\ldots,z_N\right\}$, a parametric model $\bm{f}(\,\cdot\,;\bm{W})$, a prior distribution of unknown parameters $\pi(\bm{W})$, the total budget of evaluating prompts $T$, and the number of evaluations at each prompt representation $R$ in the warm-up stage.}
\Ensure{The selected prompt representation $\widehat{z}^*_T$.}

\Statex{\textbf{\textit{Warm-Up Step:}}}
\State Decide the set of prompts $\mathcal{Z}_W$ to be evaluated in the warm-up step as in Appendix \ref{sec.warm}.
\For{$\forall z_{W;n}\in \mathcal{Z}_W$}

\State Evaluate $z_{W;n}$ for $R$ times.
\State Record the sample variances $\hat{\sigma}^2_{W;n}$.

\EndFor
\State{Approximate $\sigma^2_n$ for $n\in\left\{1,2,\ldots,N\right\}$.}
\State{Set $T_W = \left|\mathcal{Z}_W\right|R$.}

\Statex{\textbf{\textit{Sequential Evaluation Step:}}}
\State{Let $ t= T_W$ and $\mathcal{S}_t = \left\{\left(z^{(1)},\widehat{v}^{(1)}\right), \ldots, \left(z^{(t)},\widehat{v}^{(t)}\right)\right\}$}.
\While{$t< T$}
\State Update the posterior distribution $p\left(\bm{W}\mid \mathcal{S}_t\right)$ as in (\ref{eq.posterior}).
\State{Selecting the next prompt representation $z^{(t+1)}$ by maximizing the M-UCB function 
\begin{equation*}
z^{(t+1)} \in \arg\max_{z_n\in\mathcal{Z}}\alpha_t\left(z_n\right).
\end{equation*}
}

\State{Observe $\widehat{v}^{(t+1)}$ with $z^{(t+1)}$ as in (\ref{eq.generating}).}
\State{Update the historical dataset $\mathcal{S}_{t+1} = \left\{\left(z^{(t+1)},\widehat{v}^{(t+1)}\right)\right\}\cup\mathcal{S}_t$ and let $t\gets t+1$.}
\EndWhile
\vspace{-1.2mm}
\State
\Return{$n^* = \argmax_{n\in\{1,\ldots,N\}:\,r_n(T)>0}\frac{1}{r_n(T)}\sum_{m=1}^{r_n(T)}\widehat{v}_{n,m}$ and $\widehat{z}^*_T = z_{n^*}$.}

\end{algorithmic}
\end{algorithm}

The acquisition function M-UCB balances the exploitation-exploration trade-off. Specifically, the posterior mean represents the model's approximation of the mean scores of prompt representations, guiding the exploitation of known high-performing areas. Exploration accounts for both the posterior standard deviation $\sigma_t\left(z_n\right)$ and the number of evaluations $r_n\left(t\right)$. The former measures uncertainty in the surrogate prediction, while the count-based term gives additional weight to prompts with fewer evaluations. We provide the selection of $\beta_t$ and $\gamma\left(\,\cdot\,\right)$ in our numerical experiments. We now present the consistency of {Algorithm \ref{alg.1}}. We make the following assumptions.

\begin{samepage}
\begin{assumption}
    \label{assumption.1}
\begin{enumerate}
\item[]
\item For each prompt representation $z_n\in\mathcal{Z}$, the squared mean score is integrable with respect to the prior:\linebreak[4] $\mathbb{E}_{\bm{W}\sim\pi\left(\bm{W}\right)}\left[\bm{f}^2\left(z_n;\bm{W}\right)\right]<\infty$.
\item For the user-selected hyperparameter $\beta_t$, $\lim_{t\rightarrow\infty}\beta_t = \infty$.
\item The count bonus satisfies $\gamma(m)>0$ for every $m\in\{0,1,2,\ldots\}$ and $\lim_{m\to\infty}\gamma(m)=0$.
\end{enumerate}
\end{assumption}
\end{samepage}

\begin{theorem}\label{thm.infinite}
    For a fixed finite, nonempty candidate set $\mathcal{Z}$, let $z^*\in\arg\max_{z_n\in\mathcal{Z}}v\left(z_n\right)$ be a prompt with the highest mean score and $\widehat{z}^*_T$ be the prompt representation selected by Algorithm \ref{alg.1} with a budget $T$. Suppose the algorithm uses exact posterior moments under the Bayesian model in Section \ref{sec.model} and exactly maximizes M-UCB. Under Assumption \ref{assumption.1},
    \begin{equation*}
        \lim_{T\to \infty}v\left(\widehat{z}^*_T\right) \stackrel{w.p.1}{=} v\left(z^*\right),
    \end{equation*}
    where probability is with respect to the joint law of $\bm{W}\sim\pi$ and the evaluation noise.
\end{theorem}

Theorem \ref{thm.infinite} establishes consistency of the sequential evaluation step over the fixed candidate set as the evaluation budget grows. The second condition in Assumption \ref{assumption.1} is commonly employed in the literature related to Gaussian process and neural network bandit algorithms \citep{srinivas2009gaussian,zhou2020neural}. The posterior variance vanishes for any candidate evaluated infinitely often. The positive count bonus then forces every candidate to be evaluated infinitely often, so selection by the highest observed average score is consistent. The detailed proof is given in Appendix \ref{sec.proof_consistency}.
{

Moreover, Theorem \ref{thm.infinite} focuses on the second stage and provides consistency over a finite set of prompt representations. Under the simplified search assumptions of Section \ref{sec.searching}, we next connect this result to the reachable latent set.

\begin{theorem}
    \label{thm.end2end}
    Let $\mathcal{Z}^{(N)}$ consist of the indexed prompt representations $z_1,\ldots,z_N$ associated with the first $N$ accepted latent proposals $X_1,\ldots,X_N$ from one realization of the simplified search procedure in Assumption \ref{assumption.search}. Write $v(X):=v(\mathtt{Dec}(X))$, so that $v(z_n)=v(X_n)$ for each stored pair $(X_n,z_n)$. Suppose that $v$ is bounded above and lower semicontinuous on $\mathcal{X}_{\infty}$ almost surely.

    Define all runs on a joint probability space, with $\mathbb{E}[v(X_n)^2]<\infty$ for every $n$. For each $N$, run Algorithm \ref{alg.1} on $\mathcal{Z}^{(N)}$ under Assumption \ref{assumption.1}, with exact M-UCB maximization and exact conditional posterior moments given the candidate locations, their prompt representations, and the observed history. For each candidate $n\leqslant N$, assume the evaluation noises $\{\epsilon_{N,n,m}\}_{m\geqslant1}$ are independent and identically distributed as $\mathcal{N}(0,s_{N,n}^2)$ under this joint law, where the variances are deterministic and $s_{N,n}^2\leqslant b$ for all $N$ and $n\leqslant N$, with a single constant $0<b<\infty$. Use $\gamma(m)=2/\sqrt{m+1}$.

    Let $\varsigma_N>0$ be a decreasing deterministic sequence satisfying $\varsigma_N\to0$ and $\varsigma_N^2\log(N+1)\to0$. With $T_W^{(N)}$ denoting the finite warm-up cost in run $N$, define
    \begin{equation}\label{eq.tdefinition}
        T_N:=\inf\left\{t\geqslant T_W^{(N)}+1:\max_{z_n\in\mathcal{Z}^{(N)}}\left\{\sigma_t(z_n)+\gamma(r_n(t))\right\}\leqslant\varsigma_N\right\}.
    \end{equation}
    Here $t$ counts all evaluations, including warm-up, and the run index $N$ is suppressed in $\sigma_t$ and $r_n$. Then $T_N<\infty$ for all $N$ simultaneously with probability one, and
    \begin{equation*}
        \lim_{N\to\infty}v\left(\widehat{z}^*_{T_N}\right)\stackrel{w.p.1}{=}\sup_{X\in\mathcal{X}_{\infty}}v(X),
    \end{equation*}
    where $\widehat{z}^*_{T_N}$ is the prompt representation selected by Algorithm \ref{alg.1} with total budget $T_N$, and probability is with respect to the joint search, prior, and evaluation law.
\end{theorem}

Theorem \ref{thm.end2end} bridges Proposition \ref{proposition.search}, focusing on the latent space, and Theorem \ref{thm.infinite}, focusing on the finite prompt representation set. Our two-stage framework simplifies the search space from a high-dimensional, continuous latent space to a finite, prompt-representation set of moderate-dimensional vectors. The numerical experiments examine the performance of this construction under limited evaluation budgets (see Section \ref{sec.newexp}). Theorem \ref{thm.end2end} then extends consistency from the prompt representation set to the reachable latent set under the stated assumptions. The condition $\varsigma_N^2\log(N+1)\to0$ ensures that the evaluation counts grow fast enough to control estimation errors as the candidate set expands. The proof is given in Appendix \ref{sec.proof_end2end}. The initial set of prompts can affect the finite-budget search behavior and the resulting candidate set. How to select this initial set remains a question for future work.

}

\subsection{Computational Implementation}
\label{sec.acqimplementation}

When the posterior moments are not available in closed form, we draw parameter samples from the posterior and approximate $\mu_t\left(z_n\right)$ and $\sigma_t\left(z_n\right)$ by the sample mean and standard deviation of the corresponding surrogate predictions. These estimates are used to evaluate M-UCB. Appendix \ref{sec.mcimplementation} gives the estimators and their convergence result.

When the candidate set is large, approximating the acquisition function for every prompt in each round can be computationally expensive. To address this cost, we employ a probabilistic reparameterization method \citep{daulton2022bayesian} to optimize the acquisition function. That is, we assign a probability mass function to the feasible set of prompt representations $z_n\sim p\left(z;\theta\right)$, where $p\left(z;\theta\right)$ is a probability mass function defined on $\mathcal{Z}=\left\{z_1,z_2,\ldots,z_N\right\}$, and parametrized by a continuous parameter $\theta\in \Theta$. Specifically, we set $\Theta =[0,1]^{N}\setminus\{\mathbf{0}\}$ and let $z= z_i$ with probability $\frac{\theta^{(i)}}{\sum_{j=1}^{N}\theta^{(j)}}$, where $\theta^{(i)}$ denotes the $i$-th entry of $\theta$. We define
\begin{equation*}
    \theta^*_t\in \arg\max_{\theta\in \Theta}\left\{\widetilde{\alpha}_t(\theta)\doteq\mathbb{E}_{p(z;\theta)}\left[\alpha_t\left(z\right)\right]\right\},
\end{equation*}
which is a continuous optimization problem regarding $\theta\in \Theta$ that can be solved with gradient-based algorithms. After determining $\theta^*_{t}$, we generate $\widehat{z}^{(t+1)}\sim p\left(z;\theta^{*}_{t}\right)$ and then observe the score associated with $\widehat{z}^{(t+1)}$ as in (\ref{eq.generating}).

We name the acquisition function $\widetilde{\alpha}_t(\theta)$ \textit{Probabilistic Reparameterized Modified Upper Confidence Bound (PR-M-UCB)}, and provide the theoretical support of using PR-M-UCB.
\begin{theorem}
\label{thm.pr}
    For each $t>0$, let $\mathcal{H}_t^*=\arg\max_{z\in\mathcal{Z}}\alpha_t(z)$ and $\mathcal{J}_t^*=\arg\max_{\theta\in\Theta}\widetilde{\alpha}_t(\theta)$. Define
    \begin{equation*}
        \widehat{\mathcal{H}}_t^*=\bigcup_{\theta\in\mathcal{J}_t^*}\operatorname{support}\left(p(\,\cdot\,;\theta)\right).
    \end{equation*}
    Then, $\widehat{\mathcal{H}}_t^*=\mathcal{H}_t^*$.
\end{theorem}
Theorem \ref{thm.pr} states that the maximizers of PR-M-UCB lead to a probability distribution that generates the maximizers of M-UCB. Specifically, when the maximizer of M-UCB is unique, the maximizers of PR-M-UCB lead to a point mass distribution supported only at the unique maximizer of M-UCB.

The gradient estimator and the full procedure (Algorithm \ref{alg.detail}) are given in Appendix \ref{sec.primplementation}.

\section{Refinement}
\label{sec.refinment}
In this section, we provide an optional refinement procedure that uses the scores collected during evaluation and selection to support search beyond the candidate set constructed in the search stage. Specifically, we first refine the projection mapping from the latent space $\tilde{\mathcal{X}}\subseteq\mathbb{R}^{\tilde{D}}$. Recall that each prompt representation $z_n\in\mathcal{Z}\subseteq\mathbb{R}^{D}$ is attained by projecting a latent vector $X_n$ in the latent vector set $\mathcal{X}$ to $z_n =A X_n $, where $A$ is the projection matrix attained through principal component analysis, as described in Section \ref{sec.sampling}. This linear projection is constructed without task-score observations. We then refine the projection mapping based on the scores observed during the evaluation and selection stage, introducing a non-linear mapping $\tilde{A}^*\left(\,\cdot\,\right):\mathbb{R}^{\tilde{D}}\to \mathbb{R}^{D^*}$. This enhances the flexibility of the projection from the latent space. The dimension of the space after this refined projection, denoted as $D^*$, is not necessarily equivalent to the dimension of the prompt representation space $D$ before the refinement.%

We introduce a \textit{projection stochastic kriging} (PSK) model to facilitate the construction of the nonlinear projection mapping $\tilde{A}^*\left(X\right)$. We assume that the observed score regarding the latent vector is 
\begin{equation}
    \label{eq.skmodel}
    \widehat{v}_{m}\left(X\right) = v\left(X\right) +\epsilon_{m}\left(X\right), X\in\tilde{\mathcal{X}},
\end{equation}
where 
\begin{equation}
\label{eq.skmodel2}
    v \left(X\right)\sim \mathcal{GP}\left(0, \tilde{A}^*\left(X\right)^{\top}\tilde{A}^*\left(X'\right)\right).
\end{equation}
Here, $\widehat{v}_m(X)$ denotes a score that will be observed when evaluating the prompt $\p\left(X\right)\doteq \mathtt{Dec}\left(X\right)$. That is, given a vector in the latent space $X\in\tilde{\mathcal{X}}$, the decoder model described in Section \ref{sec.embed} outputs the corresponding prompt text. In addition, $v\left(X\right)$ is the mean score of $\p\left(X\right)$; and $\epsilon_m\left(X\right)\stackrel{i.i.d.}{\sim}\mathcal{N}\left(0,\sigma^2_{\epsilon}\left(X\right)\right)$ is the uncertainty of the scores to be observed when evaluating $\p\left(X\right)$. Furthermore, the mean score regarding the latent vector is a realization from a zero-mean Gaussian process, and the kernel function $\bm{K}\left(X,X'\right) = \tilde{A}^*\left(X\right)^{\top}\tilde{A}^*\left(X'\right)$ is defined on the latent space $\tilde{\mathcal{X}}$.

We use the observations $\mathcal{S}_T$ collected during evaluation and selection, replacing each prompt representation $z_n$ with its associated latent vector $X_n$. The data preparation and noise variance estimation are described in Appendix \ref{sec.pskimplementation}.

Suppose that the refined projection mapping $\tilde{A}^*\left(\,\cdot\,\right)$ is selected from a known class of mappings $\tilde{\mathcal{A}}$, such as a neural network model with fixed layers of nodes and unknown weight parameters, and is learned by maximizing the log marginal likelihood:
\begin{equation*}
    \tilde{A}^* = \arg\max_{\tilde{A}\in\tilde{\mathcal{A}}}\ell\left(\mathcal{S}_T;\tilde{A}\right),
\end{equation*}
where $\ell$ is the Gaussian log marginal likelihood of the observed mean scores, given in Appendix \ref{sec.pskimplementation}. The dimension $D^*$ is selected by cross-validation \citep{stone1974cross}; the fitting and validation details are also given there.

In addition to the refinement of the projection, the PSK model can also be used to select latent vectors $X\in\tilde{\mathcal{X}}$ when evaluation budget is allocated to refinement. This budget may be reserved from an overall evaluation budget or supplied after the initial evaluation and selection stage.

Using the initial observations and any refinement observations collected so far, the PSK model provides a predicted mean score $\widehat{\mu}(X)$ and prediction variance $\widehat{\sigma^2}(X)$ for any $X\in\tilde{\mathcal{X}}$. Their expressions are given in (\ref{eq.pskpredictor}) and (\ref{eq.predictionuncertainty}) in Appendix \ref{sec.pskimplementation}.

These predictions extend beyond the finite set $\mathcal{X}$ determined in the search stage. Thus, classical simulation experimental designs and simulation optimization methods, particularly those based on the Gaussian process model, can be directly applied to the latent space $\tilde{\mathcal{X}}\subseteq\mathbb{R}^{\tilde{D}}$. For instance, with an explicit Gaussian process representation for the latent vector $X\in\tilde{\mathcal{X}}$, Bayesian optimization algorithms or Gaussian process search algorithms \citep{wang2023gaussian} can be employed. These methods aim to identify prompts that may achieve higher mean scores than the prompt selected at the end of the evaluation and selection stage.

In general, suppose we select a set of $\mathcal{R} = \left\{\tilde{X}_1,\tilde{X}_2,\ldots,\tilde{X}_I\right\}$ as the additional latent vectors to evaluate with $I\geqslant 2$, and $\tilde{X}_i\neq\tilde{X}_j$ for $i\neq j$. We consider the \textit{cumulative uncertainty}
\begin{equation}
    \label{eq.cumulativeuncertainty}
    U_{I} = \sum_{i=1}^{I}\widehat{\sigma^2}\left(\tilde{X}_i\right),
\end{equation}
where $\widehat{\sigma^2}\left(\tilde{X}_i\right)$ denotes the uncertainty (\ref{eq.predictionuncertainty}) associated with the PSK predictor (\ref{eq.pskpredictor}). All terms use the same posterior after incorporating the initial observations and all $I$ additional observations, with the fitted projection mapping $\tilde{A}^*$ held fixed. To bound the cumulative uncertainty, we make the following assumptions.
\begin{assumption}
    \label{assumption.2}
    \begin{enumerate}
        \item []
        \item The norm of the refined projection mapping is finite, that is, $\sup_{X\in \tilde{\mathcal{X}}}\left\|\tilde{A}^*\left(X\right)\right\|<\infty$.
        \item The uncertainty in observed scores satisfies $\sigma^2_{\epsilon}\left(X\right)\in \left(\underline{\sigma^2_{\epsilon}},\overline{\sigma^2_{\epsilon}}\right), \forall X \in\tilde{\mathcal{X}}$, where $0<\underline{\sigma^2_{\epsilon}},\overline{\sigma^2_{\epsilon}}<\infty$.
    \end{enumerate}
\end{assumption}

\begin{theorem}
\label{thm.refinement}
Under Assumption \ref{assumption.2}, the cumulative uncertainty of the PSK predictor satisfies
    \begin{equation*}
        U_{I}\leqslant \overline{\sigma^2_{\epsilon}}D^{*}.
    \end{equation*}
Here $U_{I}$ is the cumulative uncertainty defined in (\ref{eq.cumulativeuncertainty}); $D^*$ is the dimension of the refined projected space; and $I$ denotes the number of additional evaluations in the refinement stage.
\end{theorem}

Theorem \ref{thm.refinement} bounds the cumulative prediction uncertainty independently of $I$. For fixed $D^*$, the mean uncertainty at these additional evaluation points satisfies $\bar{U}_I=U_I/I\leqslant\overline{\sigma^2_{\epsilon}}D^*/I$ and therefore shrinks to zero as $I$ approaches infinity under the stated assumptions. The bound depends explicitly on the dimension of the refined projected space $D^*$ rather than the original latent dimension $\tilde{D}$.

Given that the PSK model approximates the mean score from the latent space $\tilde{\mathcal{X}}$, an alternative strategy for the prompt selection problem is employing the PSK model as a surrogate to facilitate sequential evaluation in selecting prompts within the latent space, rather than using the prompt representation set. In Section \ref{sec.e3}, we conduct numerical experiments to compare our proposed two-stage framework with the approach that evaluates and selects the prompt using the PSK model.

\section{Experiments}
\label{sec.newexp}{
We conduct experiments to study how candidate expansion, evaluation allocation, and refinement affect prompt performance under limited budgets. Section \ref{sec.expso} compares M-UCB with classical simulation optimization methods for anomaly detection, varying the number of candidate prompts and the evaluation budget. Section \ref{sec.expbbo} examines how allocating part of the budget to refinement improves prompts for solving optimization problems. Section \ref{sec.expgpt} compares prompt generation and evaluation strategies under a shared token budget for judging human-preferred answers. The score functions used to evaluate the prompts are described in each set of experiments.

The target LLMs are accessed through text-in, text-out APIs; our procedure does not require access to their weights or gradients. The original example prompts (20 for each task) are obtained manually by checking the relevant online resources. Computing details and baseline descriptions are provided in Appendix \ref{sec:exp_settings}. Additional computational studies examine the surrogate model, acquisition maximization, and direct search within the continuous latent space; see Appendix \ref{sec.ablation}, which also reports the surrogate and warm-up settings.

\subsection{Candidate Breadth and Evaluation Allocation}
\label{sec.expso}
In the first set of experiments, we compare the proposed modified upper confidence bound (M-UCB) method for the second stage with classical simulation optimization (SO) methods. We first implement the search stage (Section \ref{sec.searching}) to fix a set of prompt representations, then compare evaluation and selection over this common set. The fixed-budget ranking and selection (R\&S) baselines are Optimal Computing Budget Allocation (OCBA) \citep{chen2011stochastic}, Expected Value of Information (EVI) \citep{alban2021expected}, and Knowledge Gradient (KG) \citep{frazier2009knowledge}. When the number of candidates exceeds the budget available for their initialization, the R\&S baselines use a uniformly sampled subset. M-UCB uses a surrogate model to share information across the full candidate set. Comparisons with Gaussian-process surrogates are included in Appendix \ref{sec.ablation}.

For the applications, we instruct LLMs to detect anomalies, a task relevant to monitoring business and equipment data \citep{su2024large,li2024anomaly,gu2025argos}. We consider two datasets: 1) credit card transactions, including both normal and fraud transactions\footnote{\url{https://www.kaggle.com/datasets/mlg-ulb/creditcardfraud}}, and 2) sensor data monitoring a water-pump system failure\footnote{\url{https://www.kaggle.com/datasets/nphantawee/pump-sensor-data}}. Both datasets have ground-truth anomaly labels. Anomaly detection is cast as binary classification, and the score function is the F1 score. In each LLM call, we append ten data instances as a group and calculate the F1 score by comparing the LLM results with the ground-truth labels to obtain the observed score $\widehat{v}$ for the prompt.

\textbf{Information sharing under a limited budget.} Figure \ref{fig:newexp1} compares the F1 scores of the selected prompts as the total budget increases, using a fixed set of 100 prompt representations and Claude 3.5 Sonnet. M-UCB generally outperforms the classical R\&S methods across the budgets considered. By constructing a surrogate model of the score function, M-UCB uses relationships among numerical prompt representations to guide evaluation. In comparison, the R\&S baselines allocate evaluations to individual alternatives without using these prompt representations. Among the R\&S methods, OCBA performs best: its allocation concentrates resources on high-variance or near-best alternatives, fitting the setting with many candidates and heterogeneous score noise. M-UCB achieves higher mean F1 scores with larger standard deviations, reflecting the variability of selecting prompts through a learned surrogate.

\begin{figure}[htbp]
\centering
\begin{minipage}[t]{0.49\textwidth}
  \centering
  \includegraphics[width=\textwidth]{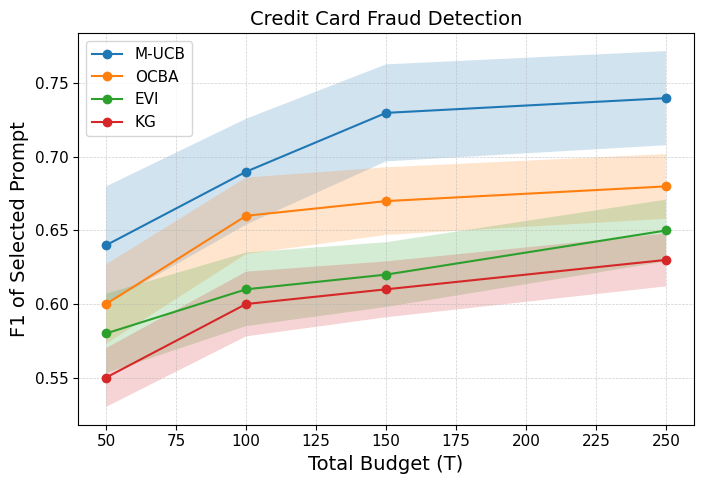}%
  \vspace{0pt}
\end{minipage}\hfill
\begin{minipage}[t]{0.49\textwidth}
  \centering
  \includegraphics[width=\textwidth]{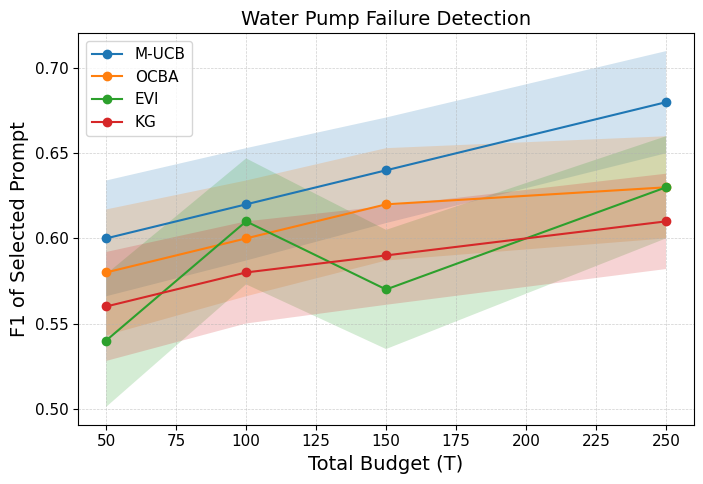}%
  \vspace{0pt}
\end{minipage}
\caption{Experimental results for comparing M-UCB with R\&S algorithms on time series anomaly detection. The employed LLM is Claude 3.5 Sonnet with $N=100$ prompt representations. The left figure is for credit card fraud detection, and the right figure is for water-pump system failure detection. The solid lines represent the mean values of the F1 score while the shadows are standard deviations.}
\label{fig:newexp1}
    \Description{Two line charts compare mean F1 scores against evaluation budget for M-UCB, OCBA, EVI, and KG on credit-card fraud detection and water-pump status prediction. M-UCB has the highest mean across the plotted budgets. Shaded bands show one standard deviation.}
\end{figure}

\textbf{Matching candidate breadth to the evaluation budget.} Table \ref{tab:abl_softprompts_budget_main} varies the number of prompt representations $N$ and the total budget $T$. When the budget is large relative to the candidate set, classical R\&S methods can match or outperform M-UCB: they can sample broadly and allocate replications to promising alternatives. For $N=20$, OCBA attains the highest mean F1 score at each budget. For $N\geqslant100$, M-UCB achieves the highest mean scores, showing the value of sharing information when direct evaluation of every candidate is costly. Increasing $N$ at a fixed budget initially offers better candidates, but the selected prompt's performance levels off or declines as the set becomes harder to evaluate. Increasing the budget at a fixed $N$ generally improves performance. The water-pump results in Table \ref{tab:abl_softprompts_budget_main_noisy_mixed} of Appendix \ref{sec:exp_anomaly} show the same need to match candidate breadth to evaluation capacity.

\begin{table}[t]
\centering
\caption{Joint effects of number of prompt representations ($N$) and total budget ($T$) on the selected prompt performance for M-UCB and R\&S algorithms based on the credit card fraud detection dataset. Values show the mean F1 score with standard deviation in parentheses. The employed LLM is Claude 3.5 Sonnet.}
\label{tab:abl_softprompts_budget_main}
\begin{threeparttable}
\begin{tabular}{@{}lS[table-format=1.2]@{\,}l@{\quad}S[table-format=1.2]@{\,}l@{\quad}S[table-format=1.2]@{\,}l@{\quad}S[table-format=1.2]@{\,}l@{}}
\toprule
$(N, T)$ & \multicolumn{2}{c}{M-UCB} & \multicolumn{2}{c}{OCBA} & \multicolumn{2}{c}{EVI} & \multicolumn{2}{c}{KG} \\
\midrule
(20, 50)    & 0.57 & (0.031) & 0.58 & (0.025) & 0.53 & (0.022) & 0.56 & (0.024) \\
(20, 100)    & 0.60 & (0.032) & 0.61 & (0.026) & 0.56 & (0.022) & 0.59 & (0.020) \\
(20, 150)    & 0.61 & (0.029) & 0.62 & (0.022) & 0.59 & (0.020) & 0.61 & (0.022) \\
(20, 250)    & 0.62 & (0.027) & 0.63 & (0.021) & 0.60 & (0.019) & 0.62 & (0.018) \\
\\[0.5em]
(100, 50)    & 0.64 & (0.040) & 0.60 & (0.027) & 0.58 & (0.027) & 0.55 & (0.020) \\
(100, 100)    & 0.69 & (0.036) & 0.66 & (0.026) & 0.61 & (0.025) & 0.60 & (0.022) \\
(100, 150)    & 0.73 & (0.033) & 0.67 & (0.023) & 0.62 & (0.022) & 0.61 & (0.019) \\
(100, 250)    & 0.74 & (0.032) & 0.68 & (0.022) & 0.65 & (0.021) & 0.63 & (0.018) \\
\\[0.5em]
(200, 50)    & 0.65 & (0.042) & 0.62 & (0.030) & 0.56 & (0.030) & 0.57 & (0.024) \\
(200, 100)    & 0.72 & (0.040) & 0.66 & (0.029) & 0.58 & (0.027) & 0.62 & (0.020) \\
(200, 150)    & 0.74 & (0.039) & 0.69 & (0.027) & 0.63 & (0.025) & 0.64 & (0.022) \\
(200, 250)    & 0.83 & (0.037) & 0.73 & (0.025) & 0.66 & (0.024) & 0.67 & (0.021) \\
\\[0.5em]
(500, 50)    & 0.64 & (0.043) & 0.60 & (0.032) & 0.54 & (0.031) & 0.55 & (0.026) \\
(500, 100)    & 0.70 & (0.045) & 0.63 & (0.030) & 0.59 & (0.029) & 0.60 & (0.022) \\
(500, 150)    & 0.73 & (0.033) & 0.68 & (0.028) & 0.62 & (0.028) & 0.61 & (0.021) \\
(500, 250)    & 0.82 & (0.037) & 0.70 & (0.026) & 0.65 & (0.027) & 0.64 & (0.020) \\
\bottomrule
\end{tabular}
\end{threeparttable}
\end{table}

Figure \ref{fig:f1comparison} helps explain this tradeoff by plotting the best attainable F1 score within nested candidate sets, all constructed from the original 20 example prompts. Increasing the number of prompt representations improves the best available score, particularly for credit card fraud detection; the gains are smaller for water-pump failure detection and diminish as the sets grow. Taken together with Table \ref{tab:abl_softprompts_budget_main}, these results distinguish the value of expanding the candidate set from the budget required to identify a good prompt within it.

\begin{figure}
    \centering
    \includegraphics[width=0.8\linewidth]{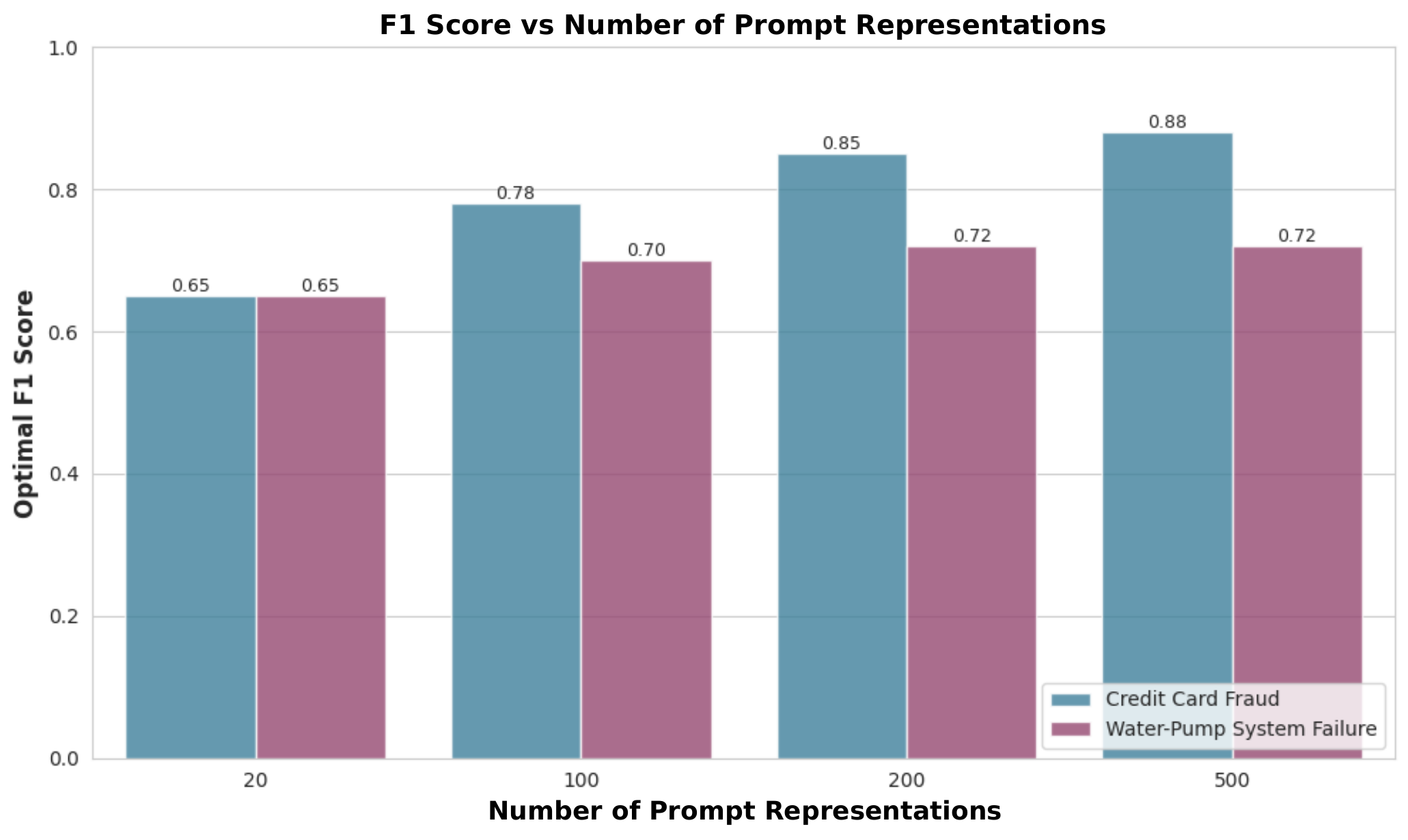}
    \caption{Best attainable F1 score within candidate sets of different sizes on the two anomaly detection datasets, using Claude 3.5 Sonnet. The smaller sets are fully included in the larger sets. Each prompt's F1 score is estimated using all data points in the dataset; the plotted value is the highest score within the set.}
    \label{fig:f1comparison}
    \Description{Two line charts show the best attainable F1 score as the candidate set grows for credit-card fraud detection and water-pump status prediction. Results are compared across nested sets of prompt representations.}
\end{figure}

The advantage of M-UCB over OCBA also appears across Claude 3.5 Sonnet, Gemini 2.0 Flash, GPT-4o, and Qwen2.5 32B, with $N=500$ and $T=250$; Figures \ref{fig:differentllms} and \ref{fig:differentllms2} in Appendix \ref{sec:exp_anomaly} report the results for both datasets.

\subsection{Improving Prompts through Staged Refinement}
\label{sec.expbbo}
The second set of experiments examines how the evaluation budget should be divided between selection from a finite candidate set and subsequent refinement. We compare M-UCB, M-UCB with refinement (M-UCB-r), and direct latent-space search using projection stochastic kriging (PSK-only). We also include two methods that optimize prompt tokens directly: RLPROMPT \citep{deng2022rlprompt} and Black-Box Discrete Prompt Learning (BDPL) \citep{diao2022black}. Baseline descriptions are given in Appendix \ref{sec:exp_settings}.

For this application, we select a prompt that instructs LLMs to solve optimization problems from textual descriptions, supporting decision-making in areas such as supply chain management, production planning, and logistics \citep{orlm2024,jiang2025llmopt}. We use the \emph{IndustryOR} dataset\footnote{\url{https://huggingface.co/datasets/CardinalOperations/IndustryOR}} proposed in \cite{orlm2024}, which consists of 100 optimization problems. Each problem includes a textual description and the numerical optimal value. Consistent with \cite{orlm2024}, the prompt's score is the accuracy of the optimal value output by the LLM: 100\% if it matches the ground truth and 0 otherwise. We append five problems to each LLM call, and the observed score $\widehat{v}$ is the average accuracy across them.

The refinement step (Section \ref{sec.refinment}) uses scores collected during evaluation and selection to fit a PSK surrogate and search beyond the finite candidate set. Under a fixed total budget of 1000 API calls, we reserve the last 100, 200, or 400 calls for refinement. Table \ref{tab:main_industryor_1000} compares these allocations using Claude 4 Sonnet and Claude 3.5 Sonnet. The reported accuracy is based on a held-out test dataset with problems that are not used to select the prompt.

\begin{table}[t]
\centering
\caption{Prompt improvement and refinement-budget allocation on IndustryOR under a fixed total budget of $T=1000$ API calls. Reported accuracy is mean$\pm$std over 3 repetitions. The last 100, 200, or 400 calls are reserved for refinement.}
\label{tab:main_industryor_1000}
\begin{subtable}[t]{0.48\linewidth}
\centering
\caption{Claude 4 Sonnet}
\begin{tabular}{lcc}
\toprule
Method & Accuracy (\%) $\uparrow$ \\
\midrule
RLPROMPT & $30.2 \pm 4.3$ \\
BDPL & $33.8 \pm 2.7$ \\
PSK-only & $35.8 \pm 2.1$ \\
M\text{-}UCB (no refinement) & $37.4 \pm 2.0$ \\
M\text{-}UCB\text{-}r (last 100)  & $38.1 \pm 2.2$ \\
\textbf{M\text{-}UCB\text{-}r (last 200)} & \bm{$41.7 \pm 3.1$} \\
M\text{-}UCB\text{-}r (last 400) & $40.6 \pm 3.5$ \\
\bottomrule
\end{tabular}
\end{subtable}\hfill
\begin{subtable}[t]{0.48\linewidth}
\centering
\caption{Claude 3.5 Sonnet}
\begin{tabular}{lcc}
\toprule
Method & Accuracy (\%)  $\uparrow$ \\
\midrule
RLPROMPT & $26.9 \pm 3.4$ \\
BDPL & $29.7 \pm 2.4$ \\
PSK-only & $30.5 \pm 2.1$ \\
M\text{-}UCB (no refinement) & $31.4 \pm 1.9$ \\
M\text{-}UCB\text{-}r (last 100)  & $32.5 \pm 2.2$ \\
\textbf{M\text{-}UCB\text{-}r (last 200)} & \bm{$35.1 \pm 2.7$} \\
M\text{-}UCB\text{-}r (last 400) & $34.6 \pm 2.6$ \\
\bottomrule
\end{tabular}
\end{subtable}
\end{table}

The experimental results in Table \ref{tab:main_industryor_1000} provide the following insights.
\begin{enumerate}
\item The refinement step improves the selected prompts by enlarging the search space. Allocating the last 200 calls to refinement raises accuracy from 37.4\% to 41.7\% on Claude 4 and from 31.4\% to 35.1\% on Claude 3.5. The budget split matters: reserving only 100 calls yields modest improvements, whereas reserving 400 leaves fewer observations for the preceding evaluation and selection stage. The best results come from allowing both stages sufficient budget. PSK-only achieves lower accuracy than staged refinement, further showing the value of first collecting informative observations over the finite candidate set. Appendix \ref{sec.e3} examines this use of previous evaluations in more detail.

\item Searching over a finite set of moderate-dimensional prompt representations and using a Bayesian surrogate yield sample efficiency. On IndustryOR, RLPROMPT and BDPL optimize discrete tokens with sparse accuracy rewards, whereas M-UCB shares score information across candidates before refining the search. Figure \ref{fig:tr} in Appendix \ref{sec:exp_industry} plots their score trajectories: M-UCB-r attains higher scores within 1000 calls, while RLPROMPT and BDPL are still improving at the end of this budget.

\item All methods achieve higher absolute scores on Claude 4 than on Claude 3.5. The accuracy advantage of M-UCB-r (last 200) over RLPROMPT and BDPL is also larger on Claude 4. Thus, staged refinement benefits both models and delivers larger gains on the more capable model in this comparison.
\end{enumerate}

Table \ref{tab:industryor_difficulty_stratified} in Appendix \ref{sec:exp_industry} further reports IndustryOR results by problem difficulty.

\subsection{Prompt Generation and Evaluation Budgets}
\label{sec.expgpt}
The third set of experiments considers the allocation of tokens between prompt generation and evaluation. We compare M-UCB with Automatic Prompt Engineer (APE) \citep{zhou2022large} and Optimization by PROmpting (OPRO) \citep{yang2023large}, which use LLMs to generate and revise prompts. This comparison accounts for the tokens consumed by both activities under a shared budget.

We use the WebGPT human-feedback dataset\footnote{\url{https://huggingface.co/datasets/openai/webgpt_comparisons}}, in which each user query is paired with two existing responses and a human preference label. The task is to select a prompt that guides the LLM to choose the human-preferred response. The observed score is the log-probability of selecting that response minus the log-probability of selecting the other response; higher scores indicate stronger agreement with human preferences. Appendix \ref{sec:exp_preference} gives the score formula and API implementation.

LLM usage comes from two activities: prompt generation and prompt evaluation. Our method constructs candidates through the latent-space search without additional generative LLM calls. For comparison, we record the score of the selected prompts under total token budgets of 2, 5, and 10 million, counting both generation and evaluation. APE and OPRO use Claude 3.5 Sonnet to generate and revise prompts; all methods use GPT-4o for evaluation.

Table \ref{tab:token-budgets} shows that M-UCB achieves the highest score at each total token budget. For APE, about 65\% of the tokens are used for prompt generation, whereas for OPRO the ratio is over 70\%. Consequently, with the same token budget, M-UCB can obtain more evaluation results to guide selection. OPRO also includes prompt histories and evaluation results in its inputs, increasing the tokens required for generation. At the largest budget, APE and OPRO narrow the performance gap; APE achieves 2.07 compared with 2.11 for M-UCB at 10 million tokens. These results show how the cost of generating candidates affects the budget available to evaluate them.

\begin{table}[ht]
  \centering
  \caption{Preference-judging scores under total token budgets covering both prompt generation and evaluation. Higher scores indicate stronger preference for the human-preferred response.}
  \begin{tabular}{lccc}
    \toprule
    Method & Score @ 2M Tokens &  Score @ 5M Tokens&  Score @ 10M Tokens \\
    \midrule
    M-UCB & $1.42$  & $1.73$  & $2.11$  \\
    APE  & $1.17$  & $1.38$  & $2.07$  \\
    OPRO  & $1.08$  & $1.35$  & $1.96$  \\
    \bottomrule
  \end{tabular}
  \label{tab:token-budgets}
\end{table}

The comparison at a fixed number of LLM calls in Table \ref{tab:call-limits} (Appendix \ref{sec:exp_preference}) complements this result. At 50 calls, APE and OPRO achieve higher scores than M-UCB, using approximately 8.6 and 15.5 million tokens, respectively, compared with 3.1 million for M-UCB. The evaluation therefore changes with the resource being constrained: calls alone do not capture the token cost of generation and longer prompt histories.

Another advantage of LLM-generated prompts is \emph{readability}. Across our experiments, these prompts are generally human-readable and grammatically correct. Our method starts from an initial set of human-readable prompts and retains candidates whose cosine similarity to the text decoded from their perturbation anchor lies within the prescribed band (Section \ref{sec.searching}). Some grammatical errors remain, and latent-space refinement can make prompts less readable, as also reported by \cite{deng2022rlprompt}. Representative prompts are provided in Appendix \ref{sec:exp_preference}.

These complementary strengths motivate future work that alternates LLM-driven prompt generation with surrogate-guided evaluation, using selected prompts to seed the next generation round.

}

\section{Conclusion}

{

In this work, we facilitate prompt selection for language models using simulation optimization methods, focusing on scenarios with limited budgets for calling language models to evaluate prompts. We conclude by pointing out limitations and some future work of this work. We prove the theoretical results for the first search stage under some stylized assumptions. Our theoretical results focus on the consistency of prompt selection, while for sample efficiency, we empirically demonstrate it via numerical experiments. Additionally, in our experiments, we observe that the proposed refinement procedure improves the performance of our two-stage framework when additional budgets are available for prompt evaluation. This raises the question of how to allocate the total budget between the evaluation and selection stage and the refinement procedure. Lastly, our experimental results show that employing language models themselves to generate and revise prompts exhibits relative advantages, especially when the budget for calling language models is not limited. Therefore, integrating both language model-driven methods and our cost-effective simulation-optimization-based method has the potential to achieve a balance between performance and efficiency. We defer these discussions to future work.

}

\clearpage

\bibliographystyle{ACM-Reference-Format}

\label{start:references}
\bibliography{references}

\clearpage

\appendix
\label{start:appendix}

\section{Implementation Details}
\label{sec.implementation}

\subsection{Prompt Representation}
\label{sec.promptimplementation}

\paragraph{Text autoencoder.}
We employ a pre-trained text autoencoder \citep{montero2021sentence} and fine-tune both its encoder and decoder on the initial set of prompts using a token-level cross-entropy reconstruction loss. Each input prompt supplies the target token sequence, and the decoder predicts these tokens from the encoded latent vector. The target language model remains fixed during this training. After fine-tuning, the initial prompts are encoded once, and the decoder is held fixed and used deterministically to generate candidate text from perturbed latent vectors, as described in Section \ref{sec.embed}.

\paragraph{Dimensionality reduction.}
The latent vectors lie in $\tilde{\mathcal{X}}=[-1,1]^{\tilde{D}}$. We compute their sample covariance matrix without standardizing the vectors and form the PCA projection from the eigenvectors corresponding to the $D$ largest eigenvalues. Perturbations in Section \ref{sec.sampling} also use a sample covariance matrix, aligning the perturbation directions with the PCA projection rather than using a pre-specified covariance structure.

\subsection{Warm-up and Noise Variance Estimation}
\label{sec.warm}
In this section, we describe the warm-up step that observes scores for some prompt representations $z_n$'s in preparation for the sequential evaluation step. Recall that we have the set of prompts $\mathcal{Z} = \left\{z_1,z_2,\ldots,z_N\right\}$. In the warm-up stage, we select $N_W (<N)$ prompt representations $z_{W;n}\in \mathcal{Z}$ and for each we observe $R$ scores as in (\ref{eq.generating}). The selection of $z_{W;n}$'s is flexible and user-specified. Here we provide a practical approach. As described in Section \ref{sec.sampling}, each prompt representation $z_n\in \mathcal{Z}$ is attained by transforming a latent vector $X_n\in\mathcal{X}$ with a dimension reduction algorithm. Furthermore, this latent vector set $\mathcal{X}$ is expanded by an initial latent vector set $\mathcal{X}^{(0)}$, which contains the latent vectors $X^{(0)}_n$'s that are attained from the initial example prompts $\left\{\p_1^{(0)},\p_2^{(0)},\ldots,\p_{N^{(0)}}^{(0)}\right\}$. In other words, the prompt representations $z_n$'s (as well as the human-readable prompts $\p\left(z_n\right)$'s) are attained by perturbing the initial example prompts since the latent vectors are attained by perturbation on the initial latent vectors as the sampling procedure in Section \ref{sec.sampling}. Therefore, the prompt representations are generated directly or indirectly from the prompt representations that represent the initial example prompts. Thus, the prompt representations representing the initial example prompts are among the most representative ones, and we therefore select a subset of these prompt representations in the warm-up stage to evaluate. That is, we set $\mathcal{Z}_{W}= \left\{z_{W;1},z_{W;2},\ldots,z_{W;N_W}\right\}\subset\mathcal{Z}$, where $N_W=5$ in our experiments and each $z_{W;n}$ corresponds to an initial latent vector in $\mathcal{X}^{(0)}$.

After deciding the set $\mathcal{Z}_W$, we collect $R=3$ observations for each $z_{W;n}\in \mathcal{Z}_W$. The warm-up therefore uses $T_W=N_WR=15$ evaluations, which are included in the total budget $T$. We then approximate the variance of the observed scores $\sigma^2_{n} = \operatorname{Var}\left[\widehat{v} _{n,m}\right]$ by $\widehat{\sigma}^2_n = \frac{1}{R-1}\sum_{r=1}^R \left ( \widehat{v}_{n,r} - \bar{v}_{n} \right ) ^2, \forall n \in \left\{n: z_n\in \mathcal{Z}_W\right\},$ where $\bar{v}_n =\frac{1}{R} \sum_{r=1}^R \widehat{v}_{n,r}$. With the set $\mathcal{D}_W\doteq\left \{ \left ( z_{W;1},\widehat{\sigma}^2_{W;1} \right ),\ldots,\left ( z_{W;N_{W}},\widehat{\sigma}^2_{W;N_{W}} \right )  \right \}$, we construct a predictive model for the log variance $g^*\left(z_n\right)\in \mathbb{R},z_n\in \mathcal{Z}$ by $
    g^*\in\arg\min_{g\in \mathcal{G} }\allowbreak\sum_{z_n\in\mathcal{Z}_W }\left ( g\left ( z_n \right )-\log\widehat{\sigma}^2_n  \right ) ^2,$ where $\mathcal{G}$ denotes a class of predictive models (e.g., a linear regression model with unknown parameters) and the selection of the predictive model is flexible and user-specified. In our experiments, we use the kriging method as suggested by \cite{ankenman2010stochastic}. After the predictive model $g^*$ is attained, we then approximate $\sigma^2_n$ for $\forall z_n \in \mathcal{Z}$ by $\exp\left(g^*\left(z_n\right)\right)$, ensuring positive variance estimates. In our work, we treat the variance $\sigma^2_n$ as a nuisance parameter, and the difference between the approximated variances $\exp\left(g^*\left(z_n\right)\right)$'s and the ground-truth variances $\sigma^2_n$'s will be ignored. These approximated variances will be incorporated into the surrogate model in the sequential evaluation step, which resembles classical surrogate models construction \citep{ankenman2010stochastic,chen2012effects,chen2013enhancing}.

\subsection{Monte Carlo Approximation of M-UCB}
\label{sec.mcimplementation}

For the Bayesian parametric model in Section \ref{sec.model}, we consider Hamiltonian Monte Carlo \citep{girolami2011riemann} for posterior sampling and variational inference \citep{blei2017variational} for constructing an approximate posterior.

\begingroup\emergencystretch=3em
When the posterior distribution of the unknown parameters $p\left(\bm{W}\mid \mathcal{S}_t\right)$ does not adopt a closed-form expression, it is challenging to obtain the closed-form expression of $\mu_t\left(z_n\right)$ and $\sigma_t\left(z_n\right)$ as well. In these scenarios, we simulate parameter samples $\widehat{\bm{W}}_k$ from the posterior or its variational approximation. Then, for each prompt representation $z_n$, we have a set of samples $
    \mathtt{Samp} _n \doteq\left\{\bm{f}\left (z_n ;\widehat{ \bm{W}}_1 \right ),\bm{f}\left (z_n ;\widehat{ \bm{W}}_2 \right ),\ldots,\bm{f}\left (z_n ;\widehat{ \bm{W}}_K \right )\right\} 
$ to infer the mean score $v\left(z_n\right)$. Thus, when optimizing the acquisition function M-UCB (\ref{eq.acquisition}), we instead consider \[
\arg\max_{z_n}\left\{\widehat{\alpha}_{t;K} \left ( z_n \right )  \doteq \widehat{ \mu}_{t;K}\left ( z_n \right ) +\beta_t\left(\widehat{\sigma}_{t;K}\left(z_n\right)+\gamma\left(r_{n}\left(t\right)\right)\right)\right\},
\] where 
\begin{equation}
\label{eq.approximate}
    \begin{aligned}
        \widehat{ \mu}_{t;K}\left ( z_n \right )  = \frac{1}{K}\sum_{k=1}^K \bm{f}\left (z_n ;\widehat{ \bm{W}}_k \right )\quad\text{and}\quad
        \widehat{ \sigma}_{t;K}\left ( z_n \right )      =\sqrt{\frac{1}{K-1}\sum_{k=1}^K\left(\bm{f}\left (z_n ;\widehat{ \bm{W}}_k \right ) - \widehat{ \mu}_{t;K}\left ( z_n \right )\right)^2}
    \end{aligned}
\end{equation}
are the sample mean and standard deviation associated with the simulated samples $\mathtt{Samp} _n$. Recall that $z^{(t+1)} \in\argmax_{z_n\in\mathcal{Z}}\alpha_t\left(z_n\right)$.\par\endgroup
\begin{proposition}
    \label{prop.mc}
    Fix $t$ and the history $\mathcal{S}_t$, with a finite candidate set $\mathcal{Z}$. Suppose $\widehat{\bm{W}}_1,\widehat{\bm{W}}_2,\ldots$ are i.i.d. draws from $p\left(\bm{W}\mid\mathcal{S}_t\right)$ and $\max_{z_n\in\mathcal{Z}}\mathbb{E}\left[\bm{f}^2\left(z_n;\bm{W}\right)\mid\mathcal{S}_t\right]<\infty$. For $K\geq2$, let
    \[
        \bar{z}_{t;K}^*\in\arg\max_{z_n\in\mathcal{Z}}\widehat{\alpha}_{t;K}(z_n).
    \]
    Then
    \begin{equation*}
        \lim_{K\rightarrow\infty}\widehat{\alpha}_{t;K}\left(\bar{z}_{t;K}^*\right)\stackrel{w.p.1.}{=}\max_{z_n\in\mathcal{Z}}\alpha_t(z_n).
    \end{equation*}
\end{proposition}
For HMC draws, the same conclusion holds if the empirical first and second moments of $\bm{f}(z_n;\widehat{\bm{W}}_k)$ converge almost surely to the corresponding posterior moments for every $z_n\in\mathcal{Z}$. For i.i.d. draws from a fixed variational approximation under which $\bm{f}(z_n;\bm{W})$ has a finite second moment for every candidate, the limit is the acquisition maximum formed from that approximation's moments. We refer to \cite{asmussen2007stochastic} for detailed discussions on practical implementations and theoretical support of algorithms for simulating samples $\widehat{\bm{W}}_{k}$.

\subsection{Optimization of PR-M-UCB}
\label{sec.primplementation}
PR-M-UCB is a continuous optimization problem, for which we use stochastic gradient ascent \citep{robbins1951stochastic}. Indeed, the differentiability of PR-M-UCB is supported by the differentiability of $p(z;\theta)$ with $\theta$, and is formalized into the following proposition.
\begin{proposition}
    \label{prop.gradient}
    For every $\theta\in\Theta$, the gradient of $\widetilde{\alpha}_t(\theta)$ is
    \begin{equation*}
        \nabla_{\theta}\widetilde{\alpha}_t(\theta)=\sum_{i=1}^{N}\alpha_t(z_i)\nabla_{\theta}p(z_i;\theta).
    \end{equation*}
    At boundary points of $\Theta$, derivatives are taken from the smooth extension of the same normalized-weight formula to the open half-space $\{\theta\in\mathbb{R}^{N}:\sum_{i=1}^{N}\theta^{(i)}>0\}$.
\end{proposition}
Furthermore, we provide an unbiased estimator of the gradient with the following proposition.
\begin{proposition}
    \label{prop.gradient_unbiased}
    For a fixed $\theta\in\Theta$ satisfying $p(z_n;\theta)>0$ for every $n=1,\ldots,N$, an unbiased estimator of the gradient $\nabla_{\theta}\widetilde{\alpha}_t\left(\theta\right)$ is
    \begin{equation}
    \label{eq.gradient}
        \widehat{\nabla_{\theta}}\widetilde{\alpha}_t(\theta) \doteq \frac{1}{I}\sum_{i=1}^{I}\alpha_{t}\left(\widehat{z}_i\right) \nabla _{\theta}\log\left(p\left ( \widehat{z}_i;\theta  \right ) \right),
    \end{equation}
    where $\widehat{z}_1,\ldots,\widehat{z}_I\stackrel{i.i.d.}{\sim}p(z;\theta)$.
\end{proposition}
In Algorithm \ref{alg.detail}, the posterior sample batch is drawn once per outer round and reused throughout the inner iterations. With this batch fixed, the approximations in (\ref{eq.approximate}) define the objective
\begin{equation*}
    \widehat{\widetilde{\alpha}}_{t;K}(\theta)\doteq\sum_{n=1}^{N}p(z_n;\theta)\widehat{\alpha}_{t;K}(z_n).
\end{equation*}
Conditional on this batch, the starting points, and past inner-iteration draws, the current iterate $\theta$ and the approximations $\widehat{\alpha}_{t;K}(z_n)$ are fixed. If all $p(z_n;\theta)>0$, replacing $\alpha_t$ in (\ref{eq.gradient}) by $\widehat{\alpha}_{t;K}$ gives a conditionally unbiased estimator of $\nabla_{\theta}\widehat{\widetilde{\alpha}}_{t;K}(\theta)$ using fresh, conditionally i.i.d. candidate draws from $p(\,\cdot\,;\theta)$. We optimize this approximate objective by stochastic gradient ascent with multiple starts, as detailed in Algorithm \ref{alg.detail}.
\newpage

\begin{algorithm}[H]
\caption{Detailed Procedure of Sequential Evaluation Step with PR-M-UCB}
\label{alg.detail}
\begin{algorithmic}[1]
\Require{
The feasible set of prompts $\mathcal{Z} = \left\{z_1,\ldots,z_N\right\}$, a parametric model $\bm{f}(\,\cdot\,;\bm{W})$, a prior distribution of unknown parameters $\pi(\bm{W})$, the total budget of evaluating prompts $T$, the collected observed scores after the warm-up stage $\mathcal{S}_{T_W}$, the estimated evaluation uncertainty $\sigma_n^2, n\in \left\{1,2,\ldots,N\right\}$, the number of iterations $\mathfrak{T}$ for gradient ascent, the sequence of positive learning rates $\ell_{\mathfrak{t}},\mathfrak{t}\in \left\{0,1,\ldots,\mathfrak{T}-1\right\}$, and the number of the starting points $\mathfrak{M}$ for gradient ascent.}
\Ensure{The selected prompt representation $\widehat{z}^*_T$.}
\State{Let $ t= T_W$}.
\While{$t< T$}
\State Update the posterior distribution $p\left(\bm{W}\mid \mathcal{S}_t\right)$ as in (\ref{eq.posterior}).
\State Sample $\widehat{\bm{W}}_k\sim p\left(\bm{W}\mid \mathcal{S}_t\right)$ for $k\in \left\{1,2,\ldots,K\right\}$.
\State Let $\mathfrak{t} = 0$ and uniformly select $\theta^{(0)}_{\mathfrak{m}}\in (0,1)^N$ in random for $\mathfrak{m} = \left\{1,2,\ldots,\mathfrak{M}\right\}$.
\While{$\mathfrak{t}<\mathfrak{T}$}
\For{$\mathfrak{m} = \left\{1,2,\ldots,\mathfrak{M}\right\}$}
\For{$i=1,2,\ldots,I$}
\State Sample $\widehat{z}_i\sim p\left(z;\theta^{(\mathfrak{t})}_{\mathfrak{m}}\right)$.
\State Estimate $a_{\mathfrak{t},\mathfrak{m},i}\doteq\widehat{\alpha}_{t;K}\left(\widehat{z}_i\right)$ with $\left\{\bm{f}\left (\widehat{z}_i ;\widehat{ \bm{W}}_1 \right ),\bm{f}\left (\widehat{z}_i ;\widehat{ \bm{W}}_2 \right ),\ldots,\bm{f}\left (\widehat{z}_i ;\widehat{ \bm{W}}_K \right )\right\}$ as in (\ref{eq.approximate}).
\EndFor

\State Calculate the estimated gradient $\eta  = \frac{1}{I}\sum_{i=1}^{I}a_{\mathfrak{t},\mathfrak{m},i} \nabla _{\theta}\log\left(p\left ( \widehat{z}_i;\theta  \right ) \right)$.
\State If needed, repeatedly halve $\ell_{\mathfrak{t}}$ until $\theta^{(\mathfrak{t})}_{\mathfrak{m}}+\ell_{\mathfrak{t}}\eta\in(0,1)^N$.
\State Let $\theta^{(\mathfrak{t}+1)}_{\mathfrak{m}} = \theta^{(\mathfrak{t})}_{\mathfrak{m}}+\ell_{\mathfrak{t}}\eta$.
\EndFor
\State Let $\mathfrak{t}\to \mathfrak{t}+1$.
\EndWhile
\State For each $\mathfrak{m}$, draw $I$ fresh candidates from $p(\,\cdot\,;\theta_{\mathfrak{m}}^{(\mathfrak{T})})$ and compute $a_{\mathfrak{T},\mathfrak{m},i}$ as above.
\State Select $\mathfrak{m}^* =\arg\max_{\mathfrak{m}\in\left \{ 1,2,\ldots,\mathfrak{M} \right \} }\sum_{i=1}^{I}a_{\mathfrak{T},\mathfrak{m},i}$.
\State Let $\theta^*_t = \theta_{\mathfrak{m}^*}^{\mathfrak{T}}$.

\State Sample $\widehat{z}^{(t+1)}\sim p\left(z;\theta^*_t\right)$.

\State{Observe $\widehat{v}^{(t+1)}$ with $\widehat{z}^{(t+1)}$ as in (\ref{eq.generating}).}
\State{Update the historical dataset $\mathcal{S}_{t+1} = \left\{\left(\widehat{z}^{(t+1)},\widehat{v}^{(t+1)}\right)\right\}\cup\mathcal{S}_t$.}
\State{Let $t\to t+1$.}
\EndWhile
\vspace{-1.2mm}
\State
\Return{$n^* = \argmax_{n\in\{1,\ldots,N\}:\,r_n(T)>0}\frac{1}{r_n(T)}\sum_{m=1}^{r_n(T)}\widehat{v}_{n,m}$ and $\widehat{z}^*_T = z_{n^*}$.}

\end{algorithmic}
\end{algorithm}

\subsection{PSK Fitting and Prediction}
\label{sec.pskimplementation}
This section gives the fitting and prediction details of the PSK model introduced in Section \ref{sec.refinment}.

\paragraph{Observed data and noise estimation.}

We now describe the preparation for learning the projection mapping from the observed scores. Recall that the set of observed scores upon completing the evaluation and selection stage is $\mathcal{S}_T =\left \{ \left ( z_n,\widehat{v}_{n,m}  \right )  \right \}$ for $m\in\left\{1,2,\ldots,r_n(T)\right\}$ and $n\in\left\{1,2,\ldots,N\right\}$, as defined in Section \ref{sec.evaluation}. Here $z_n\in\mathcal{Z}$ denotes the prompt representation, $\widehat{v}_{n,m}$ denotes the $m$-th observed score when evaluating the prompt associated with $z_n$, and $r_n(T)$ is the number of evaluations at $z_n$ when the evaluation budget $T$ allocated before refinement is reached. The formulas below are written for the case $r_n(T)\geqslant 1$ for $n\in \left\{1,2,\ldots,N\right\}$. If some candidates have not been evaluated, all vectors and matrices indexed by the initial candidates are restricted to the observed indices, with $N$ replaced by their number. Each prompt representation $z_n$ has an associated latent vector $X_n\in \mathcal{X}\subseteq \tilde{\mathcal{X}}$. In the remaining part, we consider the set of the observed scores as $\mathcal{S}_T =\left \{ \left ( X_n,\widehat{v}_{n,m}  \right )  \right \}$. That is, the prompt representations are replaced with the corresponding latent vectors. {\color{black}Regarding the observation uncertainty, we compute sample variances at the observed vectors $X_n$ with $r_n(T)\geqslant 2$ and use these to construct a predictive model to estimate $\sigma^2_{\epsilon}\left(X\right), X\in \tilde{\mathcal{X}}$, following the method described in Appendix \ref{sec.warm}.}

\paragraph{Learning the projection.}
For a candidate mapping $\tilde{A}\in\tilde{\mathcal{A}}$, the expression

\begin{equation}
\label{eq.likelihood}
\begin{aligned}
\ell\left(\mathcal{S}_T;\tilde{A}\right)=-\frac{1}{2} \ln \left[\left|\bm{K}_{N}\left(\tilde{A}\right)+\bm{\Sigma}_{\epsilon;N }\right|\right]-\frac{1}{2}\bar{\bm{V}}_N^{\top}\left[\bm{K}_{N}\left(\tilde{A}\right)+\bm{\Sigma}_{\epsilon;N }\right]^{-1}\bar{\bm{V}}_N
\end{aligned}
\end{equation}
is the log marginal likelihood of the observed mean scores under an $N$-dimensional Gaussian distribution, up to an additive constant independent of the mapping. Here $\bm{K}_N\left(\tilde{A}\right)\in\mathbb{R}^{N\times N}$ is the kernel matrix, of which the $(i,j)$-th entry is $\tilde{A}\left(X_i\right)^{\top}\tilde{A}\left(X_j\right)$; \[
\bm{\Sigma}_{\epsilon;N} = \operatorname{diag}\left\{\sigma^2_\epsilon\left(X_1\right)/r_1(T),\ldots,\sigma^2_\epsilon\left(X_N\right)/r_N(T)\right\}
\] denotes the observation uncertainty matrix; and $\bar{\bm{V}}_N = \left(\bar{v}_1,\ldots,\bar{v}_N\right)^{\top}$ is the mean vector of the observed scores, where $ \bar{v}_n = \frac{1}{r_n(T)}\sum_{m=1}^{r_n(T)} \widehat{v}_{n,m}$. In this way, the refined projection mapping is learned by maximizing the log-likelihood function through the kernel matrix $\bm{K}_N\left(\tilde{A}\right)$. We note that, instead of a pre-specified kernel function, our PSK model learns a data-driven kernel function using observed scores, allowing the projection to reflect observed task performance. {\color{black}Regarding the dimension of the refined prompt representation space $D^*$, we choose it using the cross-validation method \citep{stone1974cross}. That is, we first determine a set of potential $D^*$'s, denoted by $\mathcal{D}^* = \left\{D^*_1,\ldots,D^*_q\right\}$. For each $D^*_l\in\mathcal{D}^*$, we randomly select a proportion of the observations $\left(X_n,\widehat{v}_{n,m}\right)\in\mathcal{S}_T$, say 70\%, to construct a PSK model with dimension $D^*_l$. We then use the remaining 30\% of the observations to evaluate the prediction accuracy for the constructed PSK model \citep{bastos2009diagnostics}. The selected $D^*$ is the one that achieves the highest prediction accuracy among those in $\mathcal{D}^*$.

}

\paragraph{Prediction during refinement.}

Suppose that $I\geqslant 0$ additional evaluations have been observed associated with different latent vectors, say $\left\{\left(\tilde{X}_1,\tilde{v}_1\right),\ldots,\left(\tilde{X}_I,\tilde{v}_I\right)\right\}_{i=1}^I$. Here $\tilde{X}_i\notin \mathcal{X}$ is an additional latent vector different from those in the latent vector set decided during the search stage in Section \ref{sec.searching}, and $\tilde{v}_i = v\left(\tilde{X}_i\right)+\epsilon\left(\tilde{X}_i\right)$ is the observed score associated with $\tilde{X}_i$. We denote the set of additional latent vectors selected to evaluate during the refinement procedure as $\mathcal{R} = \left\{\tilde{X}_1,\tilde{X}_2,\ldots,\tilde{X}_I\right\}$.

Based on (\ref{eq.skmodel}) and (\ref{eq.skmodel2}), for any latent vector in the latent space $\forall X\in\tilde{\mathcal{X}}$, we have a PSK predictor for the mean score $v\left(X\right)$ as
\begin{equation}
\label{eq.pskpredictor}
    \widehat{\mu}\left(X\right) = \left(\bm{K}_N\left ( X\right )^{\top},\tilde{\bm{K}}_I\left(X\right)^{\top}\right) \left(\begin{pmatrix}
\bm{K}_N  & \tilde{\bm{K}}_{N,I}\\
 \tilde{\bm{K}}^{\top}_{N,I} & \tilde{\bm{K}}_I
\end{pmatrix}+\bm{\Sigma}_{\epsilon;N+I }\right)^{-1} \begin{pmatrix}
\bar{\bm{V}}_N \\
\tilde{\bm{V}}_I
\end{pmatrix},
\end{equation}
with the associated prediction uncertainty quantified by 
\begin{equation}
\label{eq.predictionuncertainty}
    \widehat{\sigma^2}\left(X\right) = \tilde{A}^*\left(X\right)^{\top}\tilde{A}^*\left(X\right) - \left(\bm{K}_N\left ( X\right )^{\top},\tilde{\bm{K}}_I\left(X\right)^{\top}\right) \left(\begin{pmatrix}
\bm{K}_N  & \tilde{\bm{K}}_{N,I}\\
 \tilde{\bm{K}}^{\top}_{N,I} & \tilde{\bm{K}}_I
\end{pmatrix}+\bm{\Sigma}_{\epsilon;N+I }\right)^{-1}\begin{pmatrix}
\bm{K}_N\left ( X\right ) \\
\tilde{\bm{K}}_I\left(X\right)
\end{pmatrix}.
\end{equation}
Here 
\begin{align*}
&\bm{K}_N(X) \in \mathbb{R}^{N} \text{ with } [\bm{K}_N(X)]_i = \tilde{A}^*(X)^{\top}\tilde{A}^*(X_i) \text{ for } X_i \in \mathcal{X}; \\
&\tilde{\bm{K}}_{I}(X) \in \mathbb{R}^{I} \text{ with } [\tilde{\bm{K}}_{I}(X)]_{i} = \tilde{A}^*(X)^{\top}\tilde{A}^*(\tilde{X}_i) \text{ for } \tilde{X}_i \in \mathcal{R}; \\
&\bm{K}_N \in \mathbb{R}^{N \times N} \text{ with } [\bm{K}_N]_{i,j} = \tilde{A}^*(X_i)^{\top}\tilde{A}^*(X_j) \text{ for } X_i, X_j \in \mathcal{X}; \\
&\tilde{\bm{K}}_I \in \mathbb{R}^{I \times I} \text{ with } [\tilde{\bm{K}}_I]_{i,j} = \tilde{A}^*(\tilde{X}_i)^{\top}\tilde{A}^*(\tilde{X}_j) \text{ for } \tilde{X}_i, \tilde{X}_j \in \mathcal{R}; \\
&\tilde{\bm{K}}_{N,I} \in \mathbb{R}^{N \times I} \text{ with } [\tilde{\bm{K}}_{N,I}]_{i,j} = \tilde{A}^*(X_i)^{\top}\tilde{A}^*(\tilde{X}_j) \text{ for } X_i \in \mathcal{X} \text{ and } \tilde{X}_j \in \mathcal{R}.
\end{align*}
Also, $\bm{\Sigma}_{\epsilon;N+I } = \operatorname{Diag}\left\{\sigma^2_\epsilon\left(X_1\right)/r_1(T),\ldots,\sigma^2_\epsilon\left(X_N\right)/r_N(T),\sigma^2_\epsilon\left(\tilde{X}_1\right),\ldots,\sigma^2_\epsilon\left(\tilde{X}_I\right)\right\}$ denotes the observation uncertainty matrix; $\bar{\bm{V}}_N = \left(\bar{v}_1,\ldots,\bar{v}_N\right)^{\top}$ is the mean vector of the observed scores as defined in (\ref{eq.likelihood}); and $\tilde{\bm{V}}_I = \left(\tilde{v}_1,\ldots,\tilde{v}_I\right)^{\top}$ denotes the observation vector of the additional evaluations.

\paragraph{Random latent inputs.}

Additionally, the proposed PSK model can quantify the uncertainty of observed scores when the input latent vector $\widehat{X}$ to the PSK model is itself a random vector. For detailed procedures on employing the stochastic kriging model for input uncertainty analysis, we refer to \cite{xie2014bayesian,barton2014quantifying}.

\subsection{Hyperparameter Selection in the Acquisition Function}
\label{sec.hyperparameters}

Recall that the proposed acquisition function, the modified upper confidence bound (M-UCB), is
\begin{equation*}
    \alpha_t\left(z_n\right) =\mu_{t}\left ( z_n \right ) +\beta_t\left(\sigma_{t}\left(z_n\right)+\gamma\left(r_{n}\left(t\right)\right)\right)
\end{equation*}
where $\mu_{t}\left ( z_n \right )$ and $\sigma_t\left(z_n\right)$ are the posterior mean and standard deviation of the surrogate model at $z_n$, and $r_n(t)$ denotes the number of evaluations at $z_n$ up to time $t$. In addition, $\beta_t$ is a sequence of hyperparameters and $\gamma\left(\,\cdot\,\right)$ is a user-specified non-negative decreasing function. In this section, we provide a practical procedure to decide the sequence of hyperparameters $\beta_t$ and $\gamma\left(\,\cdot\,\right)$. The procedure involves constructing a stochastic kriging (SK) model to approximate the scores of the prompt representations \citep{ankenman2010stochastic}, using the observations collected during the warm-up stage (Appendix \ref{sec.warm}). This SK model, once constructed, serves as a synthetic observation generative model. That is, when evaluating each selection of the hyperparameter in the acquisition function before the sequential evaluation step, we use the synthetic observation generated by the SK model instead of the score observations collected from the language model. Specifically, $\forall z_n\in\mathcal{Z}$, the constructed SK model generates 
\begin{equation*}
    \tilde{v}\left(z_n\right)\sim\mathcal{N}\left(\mu_{\text{SK}}\left(z_n\right), \sigma^2_{\text{SK}}\left(z_n\right)+\exp\left(g^*\left(z_n\right)\right)\right),
\end{equation*}
where $\mu_{\text{SK}}\left(z_n\right)$ and $ \sigma^2_{\text{SK}}\left(z_n\right)$ are the predicted value and mean squared error of the SK predictor, and $\exp\left(g^*\left(z_n\right)\right)$ is the estimated observation variance obtained from the log-variance predictor in Appendix \ref{sec.warm}. Regarding the explicit form of $\mu_{\text{SK}}\left(z_n\right)$ and $ \sigma^2_{\text{SK}}\left(z_n\right)$, we refer to \cite{ankenman2010stochastic}. The procedure shares similar spirits with parametric bootstrap \citep{kirk2009gaussian}.

To implement the procedure, we denote the set of the observations as $\mathcal{S}_W \doteq \left\{\left(z_i,\hat{v}_{i}\right)\right\}_{i=1}^{I_W}$, where $I_W$ is the number of the observations collected during the warm-up step. Furthermore, we assume that the pair of sequence of hyperparameters $\beta_t$ and the function $\gamma\left(\,\cdot\,\right)$ are selected from a pre-defined set $\mathfrak{C}  = \left \{ \left ( \beta_t^{(1)} ,\gamma^{(1)}\left(\,\cdot\,\right)\right ) , \left ( \beta_t^{(2)} ,\gamma^{(2)}\left(\,\cdot\,\right)\right ),\ldots, \left ( \beta_t^{(Q)} ,\gamma^{(Q)}\left(\,\cdot\,\right)\right )\right \}$. With the notation set up, the procedure works as follows:
\begin{enumerate}
    \item Construct a SK model using the dataset $\mathcal{S}_W$;
    \item For each pair $\left ( \beta_t^{(q)} ,\gamma^{(q)}\left(\,\cdot\,\right)\right )\in\mathfrak{C}$
    \begin{enumerate}
        \item Perform the sequential evaluation step (Section \ref{sec.acquisition}) using the synthetic observations $\tilde{v}\left(z_n\right)$ generated by the constructed SK model instead of using the language model to collect score observations;
        \item Record the selected prompt representation $z^{(q)}_{n^*} $ when the iteration ends;
    \end{enumerate}
    \item Select the pair $\left ( \beta_t^{(q)} ,\gamma^{(q)}\left(\,\cdot\,\right)\right )$ with the maximum value of $\mu_{\text{SK}}\left ( z^{(q)}_{n^*} \right ) $.
\end{enumerate}
In addition to a finite set of candidate hyperparameters $\mathfrak{C}$, the proposed procedure can also be adapted to the scenarios where $\beta_t$ and $\gamma\left(\,\cdot\,\right)$ are parametrized by continuous parameters. For more methods of hyperparameter selection, we refer to \cite{snoek2012practical}.

For a data-farming approach to studying how simulation-optimization solver parameters affect finite-time performance, see \cite{shashaani2024data}.

\section{Proofs}
In this section, we present the proofs of the theoretical results in the main text and Appendix \ref{sec.implementation}.
\subsection{Proof of Proposition \ref{proposition.search}}
\label{sec.proof_search}

Write $S=\tilde{\mathcal{X}}=[-1,1]^{\tilde{D}}$, let $\lambda$ denote Lebesgue measure restricted to $S$, and write $\varphi=\varphi_{\bm{\Sigma}_p}$. Under Assumption \ref{assumption.search}, each shell $\mathcal{A}(x)$ is nonempty and relatively open in $S$, so it has positive Lebesgue measure. Define
\[
a(x,y)=\mathbf{1}_{\mathcal{A}(x)}(y),\qquad
q(x,y)=a(x,y)\varphi(y-x),\qquad
\alpha(x)=\int_S q(x,y)\,\mathrm{d}y.
\]
The transition kernel is $K(x,\mathrm{d}y)=q(x,y)\,\mathrm{d}y+(1-\alpha(x))\delta_x(\mathrm{d}y)$. Let $Q(x,\mathrm{d}y)=q(x,y)\,\mathrm{d}y$ be the substochastic kernel of accepted moves.

\paragraph{Measurability and symmetry.}
Symmetry and section continuity make $F$ separately continuous, hence Borel measurable on the product of Euclidean boxes. For example, continuous partitions of unity $\{\psi_{n,j}\}_j$ in the first coordinate, with supports shrinking to their centers $x_{n,j}$, give jointly continuous functions $F_n(x,y)=\sum_j\psi_{n,j}(x)F(x_{n,j},y)$ converging pointwise to $F(x,y)$. Thus $a$ and $q$ are measurable. The symmetry of $F$ and the centered Gaussian density gives $a(x,y)=a(y,x)$ and $q(x,y)=q(y,x)$. For each fixed $y$, the map $x\mapsto a(y,x)$ is lower semicontinuous, since it is the indicator of a relatively open set.

\paragraph{Two-step transitions.}
The kernel $Q^2$ has density
\[
h(x,z)=\int_S q(x,y)q(y,z)\,\mathrm{d}y
=\int_S a(y,x)a(y,z)\varphi(y-x)\varphi(z-y)\,\mathrm{d}y.
\]
For each fixed $y$, the integrand is nonnegative and lower semicontinuous in $(x,z)$. Fatou's lemma therefore gives joint lower semicontinuity of $h$. The density $h$ is finite because $q$ is bounded on the finite-volume box. On the diagonal,
\[
h(x,x)=\int_{\mathcal{A}(x)}\varphi(y-x)^2\,\mathrm{d}y>0.
\]
This follows from the positive measure of $\mathcal{A}(x)$ and strict positivity of the Gaussian density.

\paragraph{Uniform local bound.}
The lower semicontinuous function $x\mapsto h(x,x)$ attains a positive minimum $m$ on the compact set $S$. Set $c=m/2>0$. There exists $\eta>0$ such that
\[
\|x-z\|<\eta\quad\Longrightarrow\quad h(x,z)\geq c
\qquad(x,z\in S).
\]
Otherwise there would be sequences $x_n,z_n\in S$ with $\|x_n-z_n\|\to0$ and $h(x_n,z_n)<c$. Along a subsequence, compactness gives $x_n,z_n\to x_*$. Lower semicontinuity would imply $h(x_*,x_*)\leq c<m$, a contradiction.

\paragraph{Global minorization.}
Choose an integer $M\geq2$ such that $\operatorname{diam}(S)/M<\eta/2$, and choose $0<r\leq\min\{\eta/8,1/2\}$. For every $v\in S$,
\[
\lambda(\mathcal{B}_r(v)\cap S)\geq v_r:=2^{-\tilde{D}}\omega_{\tilde{D}}r^{\tilde{D}}>0,
\]
where $\omega_{\tilde{D}}$ is the volume of the unit Euclidean ball. Indeed, choosing the direction toward the box midpoint in each coordinate gives an orthant of the ball lying inside $S$. This bound also holds at faces and corners.

For any endpoints $x,z\in S$, let $v_i=(1-i/M)x+(i/M)z$, for $i=0,\ldots,M$. Convexity gives $v_i\in S$. Set $w_0=x$, $w_M=z$, and integrate each $w_i$ over $U_i=\mathcal{B}_r(v_i)\cap S$, for $i=1,\ldots,M-1$. Consecutive points satisfy
\[
\|w_i-w_{i-1}\|\leq\operatorname{diam}(S)/M+2r<\eta,
\]
so $h(w_{i-1},w_i)\geq c$. The density of $Q^{2M}=(Q^2)^M$ consequently satisfies
\[
\begin{aligned}
h^{(*M)}(x,z)
&\geq\int_{U_1\times\cdots\times U_{M-1}}\prod_{i=1}^M h(w_{i-1},w_i)\,\mathrm{d}w_1\cdots\mathrm{d}w_{M-1}\\
&\geq c^M v_r^{M-1}=:b>0.
\end{aligned}
\]
Since $K\geq Q$, we have $K^{2M}\geq Q^{2M}$. Thus, for every $x\in S$ and Borel set $B\subseteq S$,
\[
K^{2M}(x,B)\geq b\lambda(B).
\]

\paragraph{Infinitely many visits and density.}
Fix a Borel set $B\subseteq S$ with $\lambda(B)>0$, and put $L=2M$ and $p_B=b\lambda(B)\in(0,1]$. From any initial state $x_0\in S$, the Markov property and minorization bound give
\[
\mathbb{P}_{x_0}(X_{(k+1)L}\in B\mid\mathcal{F}_{kL})\geq p_B,
\]
where $\mathcal{F}_n$ is the history through step $n$. Iterated conditioning yields, for every $k\geq0$ and $n\geq1$,
\[
\mathbb{P}_{x_0}\left(\bigcap_{j=1}^n\{X_{(k+j)L}\notin B\}\right)\leq(1-p_B)^n.
\]
Letting $n\to\infty$ and taking a countable union over $k$ shows that $B$ is visited infinitely often almost surely. A countable basis for the relative topology on $S$ consists of its nonempty intersections with rational-center, rational-radius open balls. Each such set has positive Lebesgue measure. With probability one, all these sets are visited infinitely often, so $\{X_n:n\geq0\}$ is dense in $S$. Every neighborhood in (\ref{eq.set}) is therefore reached with probability one, giving $\mathcal{X}_{\infty}(x_0)=S$.

\paragraph{Accepted subsequence.}
By symmetry,
\[
\alpha(x)=\int_S a(y,x)\varphi(y-x)\,\mathrm{d}y.
\]
Fatou's lemma again gives lower semicontinuity. Since $\alpha(x)>0$ for every $x\in S$, compactness gives $\alpha_*:=\min_{x\in S}\alpha(x)>0$. The probability of no acceptance in the next $n$ proposal steps is at most $(1-\alpha_*)^n$, so accepted proposals occur infinitely often almost surely. Let $\tau_0=0$ and let $\tau_1<\tau_2<\cdots$ be the accepted-proposal times. Rejections leave the state unchanged, so
\[
\{X_n:n\geq0\}=\{X_{\tau_j}:j\geq0\}.
\]
The accepted-only chain has kernel $J(x,\mathrm{d}y)=q(x,y)\,\mathrm{d}y/\alpha(x)$, whereas $K$ also records rejections. Their visited sets, including the initial point, have the same closure. Removing the initial point preserves density because $S$ has no isolated points. Thus the accepted subsequence is also dense almost surely. \qed

\subsection{Proof of Theorem \ref{thm.infinite}}
\label{sec.proof_consistency}
We consider one run of Algorithm \ref{alg.1} with a fixed finite warm-up, hyperparameter sequence $\{\beta_t\}$, and count-bonus function $\gamma$; $\widehat{z}^*_T$ is its recommendation after $T$ evaluations, including the warm-up.
Let $V_n=\bm{f}(z_n;\bm{W})=v(z_n)$, let $\mathcal{F}_t$ be the $\sigma$-algebra generated by the observed history through round $t$, and let $\mathcal{F}_\infty=\sigma(\bigcup_t\mathcal{F}_t)$. Assumption \ref{assumption.1} makes $V_n$ square-integrable. By L\'evy's upward convergence theorem, the posterior moments satisfy
\begin{equation*}
    \begin{aligned}
        \mu_t(z_n)=\mathbb{E}[V_n\mid\mathcal{F}_t]&\longrightarrow\mathbb{E}[V_n\mid\mathcal{F}_\infty],\\
        \mathbb{E}[V_n^2\mid\mathcal{F}_t]&\longrightarrow\mathbb{E}[V_n^2\mid\mathcal{F}_\infty]
    \end{aligned}
\end{equation*}
almost surely. Both limits are finite almost surely. Hence $\mu_t(z_n)$ and $\sigma_t(z_n)=\{\mathbb{E}[V_n^2\mid\mathcal{F}_t]-\mu_t(z_n)^2\}^{1/2}$ have finite limits, with the latter nonnegative.

For each $n$, define $A_n=\{r_n(t)\to\infty\}$. Since the counts are observed and nondecreasing,
\begin{equation*}
    A_n=\bigcap_{m=1}^{\infty}\bigcup_{t=1}^{\infty}\{r_n(t)\geqslant m\}\in\mathcal{F}_\infty.
\end{equation*}
The strong law of large numbers for the independent and identically distributed noise sequence $\{\epsilon_{n,m}\}_{m\geqslant1}$ gives $m^{-1}\sum_{\ell=1}^{m}\widehat{v}_{n,\ell}\to V_n$ almost surely \citep{van2000asymptotic}. Composing this convergence with the adaptive count $r_n(t)$ shows that the observed average
\begin{equation*}
    \bar{v}_t(z_n)=\frac{1}{r_n(t)}\sum_{m=1}^{r_n(t)}\widehat{v}_{n,m}
\end{equation*}
converges to $V_n$ almost surely on $A_n$; set $\bar{v}_t(z_n)=0$ when $r_n(t)=0$. Each average is $\mathcal{F}_t$-measurable. Its limit on $A_n$ therefore gives an $\mathcal{F}_\infty$-measurable version of $\mathbf{1}_{A_n}V_n$, and squaring gives one for $\mathbf{1}_{A_n}V_n^2$. For $q\in\{1,2\}$, the conditional-expectation identity
\begin{equation*}
    \mathbf{1}_{A_n}\mathbb{E}[V_n^q\mid\mathcal{F}_\infty]
    =\mathbb{E}[\mathbf{1}_{A_n}V_n^q\mid\mathcal{F}_\infty]
    =\mathbf{1}_{A_n}V_n^q
\end{equation*}
holds almost surely. Thus $\mu_t(z_n)\to V_n$ and $\sigma_t(z_n)\to0$ almost surely on $A_n$.

Work on the probability-one event where these conclusions hold for all finitely many candidates. Since $\beta_t\to\infty$, it is eventually positive, and maximizing M-UCB is then equivalent to maximizing
\begin{equation*}
    \alpha_{\text{res};t}(z_n)=\frac{\mu_t(z_n)}{\beta_t}+\sigma_t(z_n)+\gamma(r_n(t)).
\end{equation*}
At least one candidate, say $z_k$, is selected infinitely often because $\mathcal{Z}$ is finite. Its count diverges, so $\alpha_{\text{res};t}(z_k)\to0$. If another candidate $z_j$ were evaluated only finitely often, its nondecreasing integer count would eventually equal some finite $c$. Writing $\sigma_\infty(z_j)\geqslant0$ for its limiting posterior standard deviation, we would have
\begin{equation*}
    \alpha_{\text{res};t}(z_j)\longrightarrow\sigma_\infty(z_j)+\gamma(c)>0.
\end{equation*}
At every sufficiently late round in which $z_k$ is selected, exact maximization requires $\alpha_{\text{res};t}(z_k)\geqslant\alpha_{\text{res};t}(z_j)$. There are infinitely many such rounds, contradicting these limits. Hence every candidate is evaluated infinitely often.

It follows that $\bar{v}_T(z_n)\to v(z_n)$ simultaneously for all candidates, almost surely. Eventually all candidates have been evaluated, and Algorithm \ref{alg.1} selects one with the largest observed average. Therefore,
\begin{equation*}
    0\leqslant v(z^*)-v(\widehat{z}^*_T)
    \leqslant2\max_{z_n\in\mathcal{Z}}\left|\bar{v}_T(z_n)-v(z_n)\right|\longrightarrow0
\end{equation*}
almost surely. Including the fixed finite warm-up cost in $T$ does not affect this limit. \qed

\subsection{Proof of Theorem \ref{thm.end2end}}
\label{sec.proof_end2end}
{
With the candidate locations and representations included in the initial information, and with $\mathbb{E}[v(X_n)^2]<\infty$ supplying square integrability, the proof of Theorem \ref{thm.infinite} gives, for each fixed $N$, $r_n(t)\to\infty$ and $\sigma_t(z_n)\to0$ almost surely for every $n\leqslant N$. Thus $\max_{n\leqslant N}\{\sigma_t(z_n)+\gamma(r_n(t))\}\to0$, so the first qualifying round after the finite warm-up exists. Taking a countable intersection over $N$ gives $T_N<\infty$ for all $N$ on one probability-one event.

At $T_N$, nonnegativity of the posterior standard deviation implies $2/\sqrt{r_n(T_N)+1}\leqslant\varsigma_N$. Hence, for all sufficiently large $N$ and every $n\leqslant N$,
\begin{equation*}
    r_n(T_N)\geqslant\frac{4}{\varsigma_N^2}-1\geqslant k_N,
    \qquad k_N:=\left\lfloor\frac{1}{\varsigma_N^2}\right\rfloor\geqslant1.
\end{equation*}
The rate condition gives $k_N/\log(N+1)\to\infty$.

Let $e_{N,n,m}=m^{-1}\sum_{\ell=1}^m\epsilon_{N,n,\ell}$ be the mean of the first $m$ evaluation noises in run $N$. For $\varepsilon>0$ and $m\geqslant1$, the Gaussian tail bound gives $\mathbb{P}(|e_{N,n,m}|\geqslant\varepsilon)\leqslant2\exp(-m\varepsilon^2/(2b))$. Set $c=\varepsilon^2/(2b)>0$. A union bound over all $N$ candidates and all sample counts $m\geqslant k_N$ yields
\begin{equation*}
    \mathbb{P}\left(\exists n\leqslant N,\ m\geqslant k_N:\ |e_{N,n,m}|\geqslant\varepsilon\right)
    \leqslant\frac{2N\exp(-c k_N)}{1-\exp(-c)}.
\end{equation*}
Eventually $k_N\geqslant(4/c)\log(N+1)$, so the right-hand side is bounded by a constant times $(N+1)^{-3}$ and is summable in $N$. By the first Borel--Cantelli lemma, the displayed event occurs only finitely often almost surely. Intersecting these probability-one events for $\varepsilon=1/j$, $j\geqslant1$, gives $\sup_{n\leqslant N,\,m\geqslant k_N}|e_{N,n,m}|\to0$ almost surely. The observed average in run $N$ satisfies $\bar{v}_{T_N}(z_n)-v(X_n)=e_{N,n,r_n(T_N)}$ whenever $r_n(T_N)>0$. Since each adaptive count $r_n(T_N)$ is at least $k_N\geqslant1$ eventually,
\begin{equation*}
    E_N:=\max_{n\leqslant N}\left|\bar{v}_{T_N}(z_n)-v(X_n)\right|\longrightarrow0
    \qquad\text{almost surely},
\end{equation*}
The bound over all sample counts allows this conclusion at the data-dependent stopping time.

Proposition \ref{proposition.search} gives almost-sure density of $D_p=\{X_n:n\geqslant1\}$ in $\mathcal{X}_{\infty}$. On a path satisfying this density and the score assumptions, let $v^\star=\sup_{X\in\mathcal{X}_{\infty}}v(X)$. For any $a<v^\star$, lower semicontinuity makes $\{X\in\mathcal{X}_{\infty}:v(X)>a\}$ a nonempty relatively open set. It contains some candidate $X_j$, so every prefix with $N\geqslant j$ has maximum value greater than $a$. The prefix maxima are nondecreasing and bounded above by $v^\star$, proving
\begin{equation}
\label{eq.le}
    \lim_{N\to\infty}\max_{1\leqslant n\leqslant N}v(X_n)=v^\star
    \qquad\text{almost surely}.
\end{equation}

For all sufficiently large $N$, every candidate has been evaluated at $T_N$. Let $z_N^*$ maximize the true score over $\mathcal{Z}^{(N)}$. Since Algorithm \ref{alg.1} recommends a candidate with the largest observed average,
\begin{equation*}
    \begin{aligned}
    v(z_N^*)-v(\widehat{z}^*_{T_N})
    &\leqslant\left|v(z_N^*)-\bar{v}_{T_N}(z_N^*)\right|
      +\left|\bar{v}_{T_N}(\widehat{z}^*_{T_N})-v(\widehat{z}^*_{T_N})\right|\\
    &\leqslant2E_N.
    \end{aligned}
\end{equation*}
Combining this bound with (\ref{eq.le}) gives
\begin{equation*}
    0\leqslant v^\star-v(\widehat{z}^*_{T_N})
    \leqslant v^\star-\max_{1\leqslant n\leqslant N}v(X_n)+2E_N\longrightarrow0
\end{equation*}
almost surely under the joint law. \qed
}

\subsection{Proofs for Acquisition Function Computation}
{\color{black}
In this section, we prove Theorem \ref{thm.pr} and Propositions \ref{prop.mc}, \ref{prop.gradient}, and \ref{prop.gradient_unbiased} in Appendices \ref{sec.mcimplementation} and \ref{sec.primplementation}.

Regarding Proposition \ref{prop.mc}, fix $t$ and $\mathcal{S}_t$. The finite second moments allow the strong law of large numbers \citep{van2000asymptotic} to be applied to both $\bm{f}(z_n;\widehat{\bm{W}}_k)$ and its square. For each $z_n\in\mathcal{Z}$, almost surely,
\begin{equation*}
\begin{aligned}
    \widehat{\mu}_{t;K}(z_n)&\longrightarrow\mu_t(z_n),\\
    \frac{1}{K}\sum_{k=1}^{K}\bm{f}^2(z_n;\widehat{\bm{W}}_k)&\longrightarrow\mathbb{E}\left[\bm{f}^2(z_n;\bm{W})\mid\mathcal{S}_t\right].
\end{aligned}
\end{equation*}
For $K\geq2$, the sample variance satisfies
\begin{equation*}
    \widehat{\sigma}_{t;K}^{2}(z_n)
    =\frac{K}{K-1}\left(\frac{1}{K}\sum_{k=1}^{K}\bm{f}^2(z_n;\widehat{\bm{W}}_k)-\widehat{\mu}_{t;K}^{2}(z_n)\right)
    \longrightarrow\sigma_t^{2}(z_n)
\end{equation*}
almost surely. Continuity of the square root gives $\widehat{\sigma}_{t;K}(z_n)\to\sigma_t(z_n)$, including when $\sigma_t(z_n)=0$. Since $t$ and the history are fixed, $\widehat{\alpha}_{t;K}(z_n)\to\alpha_t(z_n)$ almost surely. Finiteness of $\mathcal{Z}$ then gives
\begin{equation*}
    \delta_K\doteq\max_{z_n\in\mathcal{Z}}\left|\widehat{\alpha}_{t;K}(z_n)-\alpha_t(z_n)\right|\longrightarrow0
\end{equation*}
almost surely. Finally,
\begin{equation*}
    \left|\max_{z_n\in\mathcal{Z}}\widehat{\alpha}_{t;K}(z_n)-\max_{z_n\in\mathcal{Z}}\alpha_t(z_n)\right|\leq\delta_K.
\end{equation*}
Because $\widehat{\alpha}_{t;K}(\bar{z}_{t;K}^*)=\max_{z_n\in\mathcal{Z}}\widehat{\alpha}_{t;K}(z_n)$, this proves Proposition \ref{prop.mc}.

Regarding Theorem \ref{thm.pr}, fix $t>0$ and write $a_i=\alpha_t(z_i)$ and $M=\max_{1\leq i\leq N}a_i$. For any $\theta\in\Theta$, let $s=\sum_{j=1}^{N}\theta^{(j)}>0$ and $p_i=p(z_i;\theta)=\theta^{(i)}/s$. Since $p_i\geq0$ and $\sum_i p_i=1$,
\begin{equation*}
    M-\widetilde{\alpha}_t(\theta)=\sum_{i=1}^{N}p_i(M-a_i)\geq0.
\end{equation*}
For every $z_j\in\mathcal{H}_t^*$, the standard basis vector $e_j\in\Theta$ attains $\widetilde{\alpha}_t(e_j)=M$ and induces a point mass at $z_j$. Hence the maximum over $\Theta$ is $M$. Equality in the display holds if and only if every index with $p_i>0$ satisfies $a_i=M$. Thus each optimal distribution is supported on $\mathcal{H}_t^*$, while each point in $\mathcal{H}_t^*$ is the support of an optimal point mass. Taking the union over all optimal parameters gives $\widehat{\mathcal{H}}_t^*=\mathcal{H}_t^*$.

Regarding Proposition \ref{prop.gradient}, on the open half-space $s=\sum_{i=1}^{N}\theta^{(i)}>0$, the functions $p(z_i;\theta)=\theta^{(i)}/s$ are smooth. Differentiating the finite sum gives
\begin{equation*}
    \nabla_{\theta}\widetilde{\alpha}_t(\theta)=\sum_{i=1}^{N}\alpha_t(z_i)\nabla_{\theta}p(z_i;\theta).
\end{equation*}
In particular, the quotient rule yields, for $j=1,\ldots,N$,
\begin{equation*}
    \frac{\partial\widetilde{\alpha}_t(\theta)}{\partial\theta^{(j)}}
    =\frac{s\alpha_t(z_j)-\sum_{i=1}^{N}\theta^{(i)}\alpha_t(z_i)}{s^2}
    =\frac{\alpha_t(z_j)-\widetilde{\alpha}_t(\theta)}{s}.
\end{equation*}
This proves Proposition \ref{prop.gradient}, including at boundary points of $\Theta$ under the stated extension.

For Proposition \ref{prop.gradient_unbiased}, fix $\theta\in\Theta$ with $p(z_n;\theta)>0$ for every candidate. Expanding the expectation over the finite set $\mathcal{Z}$ gives
\begin{equation*}
\begin{aligned}
    \mathbb{E}_{p(z;\theta)}\left[\alpha_t(z)\nabla_{\theta}\log p(z;\theta)\right]
    &=\sum_{n=1}^{N}p(z_n;\theta)\alpha_t(z_n)\frac{\nabla_{\theta}p(z_n;\theta)}{p(z_n;\theta)}\\
    &=\sum_{n=1}^{N}\alpha_t(z_n)\nabla_{\theta}p(z_n;\theta)
    =\nabla_{\theta}\widetilde{\alpha}_t(\theta),
\end{aligned}
\end{equation*}
where the last equality follows from Proposition \ref{prop.gradient}. Averaging over $I$ i.i.d. draws preserves this expectation, proving Proposition \ref{prop.gradient_unbiased}. Conditioning on the posterior sample batch, the starting points, and past inner-iteration draws fixes both the current iterate $\theta$ and the coefficients $\widehat{\alpha}_{t;K}(z_n)$. The same finite-sum calculation then establishes the stated conditional unbiasedness for $\nabla_{\theta}\widehat{\widetilde{\alpha}}_{t;K}(\theta)$.

The estimator in (\ref{eq.gradient}) is also known as REINFORCE \citep{williams1992simple} and the likelihood ratio estimator \citep{glynn2019likelihood}.

In addition, we note that the maximization of PR-M-UCB is reformulated as a linear programming problem when we impose the constraint $\sum_{i=1}^N\theta^{(i)}=1$. On the other hand, the coefficients $\left ( \alpha_t\left(z_1\right),\alpha_t\left(z_2\right),\ldots,\alpha_t\left(z_N\right) \right )$ in the objective function are unknown and require estimation, which is time-consuming if $N$ is large. Therefore, we employ stochastic gradient ascent to optimize PR-M-UCB. We refer to \cite{louveaux2003stochastic} for detailed discussions on stochastic linear programming.

}

\subsection{Proof of Theorem \ref{thm.refinement}}
Let $d=D^*$, $b=\overline{\sigma^2_{\epsilon}}$, and $\phi_i=\tilde{A}^*(\tilde{X}_i)$. Define the feature matrices
\begin{equation*}
    \bm{A}_I=[\phi_1,\ldots,\phi_I],\qquad
    \bm{B}=[\tilde{A}^*(X_1),\ldots,\tilde{A}^*(X_N)],
\end{equation*}
where $X_1,\ldots,X_N$ index the initially observed candidates as in Appendix \ref{sec.pskimplementation}. Let $\bm{D}_0=\bm{\Sigma}_{\epsilon;N}$ be the noise covariance matrix of their observed mean scores, and let
\begin{equation*}
    \bm{D}_I=\operatorname{Diag}\{\sigma^2_{\epsilon}(\tilde{X}_1),\ldots,\sigma^2_{\epsilon}(\tilde{X}_I)\}.
\end{equation*}
Both noise covariance matrices have positive diagonal entries. The initial coefficient precision and the final coefficient covariance are
\begin{equation*}
    \bm{P}=\bm{I}_d+\bm{B}\bm{D}_0^{-1}\bm{B}^{\top},\qquad
    \bm{S}_I=(\bm{P}+\bm{A}_I\bm{D}_I^{-1}\bm{A}_I^{\top})^{-1},
\end{equation*}
which are positive definite. Writing $\bm{F}=[\bm{B},\bm{A}_I]$, the kernel matrix in (\ref{eq.predictionuncertainty}) is $\bm{F}^{\top}\bm{F}$ and its noise matrix is block diagonal with blocks $\bm{D}_0$ and $\bm{D}_I$. Applying the Woodbury identity to that equation gives $\widehat{\sigma^2}(\tilde{X}_i)=\phi_i^{\top}\bm{S}_I\phi_i$.

Assumption \ref{assumption.2} gives $\bm{D}_I\preceq b\bm{I}_I$, so $\bm{A}_I\bm{A}_I^{\top}\preceq b\bm{A}_I\bm{D}_I^{-1}\bm{A}_I^{\top}$, where $\preceq$ denotes the positive-semidefinite order. Since $\bm{S}_I$ is positive definite, trace monotonicity gives
\begin{equation*}
    \begin{aligned}
        U_I&=\operatorname{tr}(\bm{S}_I\bm{A}_I\bm{A}_I^{\top})\\
        &\leqslant b\operatorname{tr}(\bm{S}_I\bm{A}_I\bm{D}_I^{-1}\bm{A}_I^{\top})\\
        &=b\bigl[d-\operatorname{tr}(\bm{S}_I\bm{P})\bigr]\leqslant bd.
    \end{aligned}
\end{equation*}
The equality uses $\bm{S}_I(\bm{P}+\bm{A}_I\bm{D}_I^{-1}\bm{A}_I^{\top})=\bm{I}_d$, and the final inequality follows from $\operatorname{tr}(\bm{S}_I\bm{P})=\operatorname{tr}(\bm{S}_I^{1/2}\bm{P}\bm{S}_I^{1/2})\geqslant0$. Substituting $d=D^*$ and $b=\overline{\sigma^2_{\epsilon}}$ proves the bound.

\section{Surrogate Models}
\label{sec.surrogates}
This appendix describes four score surrogates for the framework in Section \ref{sec.model}: Bayesian linear regression (BLR), a finite-rank Gaussian process (GP), a Bayesian neural network (BNN), and a variational autoencoder (VAE). Appendix \ref{sec.e1} compares their predictive performance; we use the BNN in the other experiments, including those in Section \ref{sec.newexp}.
\subsection{Bayesian Linear Regression}
Bayesian linear regression \citep{minka2000bayesian} models the observed score as 
\begin{equation*}
    \widehat{v}(z)=\boldsymbol{W}^{\top} z+\epsilon,
\end{equation*}
where $\widehat{v}(z)$ is the observed score, $z$ is the input vector, $\boldsymbol{W}$ is the vector of unknown weights or parameters, and $\epsilon$ is the noise term, typically assumed to be Gaussian with zero mean and variance $\sigma^2$.

We calculate the posterior distribution of $\boldsymbol{W}$ given the data using Bayes' theorem:
$$
P(\boldsymbol{W} \mid \boldsymbol{Z}, \boldsymbol{V})=\frac{P(\boldsymbol{V} \mid \boldsymbol{Z}, \boldsymbol{W}) \pi\left(\bm{W}\right)}{P(\boldsymbol{V} \mid \boldsymbol{Z})}
$$
where $\boldsymbol{Z}$ is the matrix of input data, $\boldsymbol{V}$ is the vector of output scores, $\pi\left(\bm{W}\right)$ is the prior distribution, $P(\boldsymbol{V} \mid \boldsymbol{Z}, \boldsymbol{W})$ is the likelihood, and $P(\boldsymbol{V} \mid \boldsymbol{Z})$ is the marginal likelihood. Assuming the noise variance $\sigma^2$ is known, a standard prior is the Gaussian distribution $P(\boldsymbol{W})=\mathcal{N}\left(\boldsymbol{W} \mid \boldsymbol{\mu}_0, \boldsymbol{\Sigma}_0\right)$, with $\boldsymbol{\mu}_0$ and $\boldsymbol{\Sigma}_0$ specified by the user.

\subsection{Finite-Rank Gaussian Process}
The general GP formulation \citep{quan2013simulation,wang2023gaussian} assumes $\bm{f}(z) \sim \mathcal{GP}\left(0,\bm{K}\left(z,z'\right)\right)$, where $\bm{K}\left(z,z'\right)$ is a pre-specified kernel function. A common selection is the radial basis function (RBF) kernel $\bm{K}_{\operatorname{RBF}}\left(z,z'\right) = \exp\left\{-\frac{\left\|z-z'\right\|^2}{\ell^2}\right\},$ where $\left\|\,\cdot\,\right\|$ denotes the Euclidean norm and $\ell^2$ is the kernel bandwidth. Given the observed points, inference for the function value $f$ at an arbitrary point $\tilde{z}$ is based on the conditional distribution of a multivariate normal distribution. That is, $\bm{f}\left(\tilde {z}\right)\mid \tilde {S}_t \sim \mathcal{N}\left(\mu_{GP;t}\left(\tilde {z}\right),\sigma^2_{GP;t}\left(\tilde {z}\right)\right)$. Here, $\tilde {S}_t = \left \{ \left(z^{(1)},y^{(1)}  \right),\ldots,\left(z^{(t)},y^{(t)}  \right)\right \}$ denotes the pairs of inputs and observations up to time $t$, where $y^{(\tau)} = \bm{f}\left (z^{(\tau)}  \right ) +\epsilon^{(\tau)} $ denotes the noisy observation, and $\epsilon^{(\tau)} \stackrel{i.i.d.}{\sim}\mathcal{N}\left(0,\sigma^2\right)$ denotes the noise. The conditional mean and variance adopt explicit expressions
\begin{equation}
\label{eq.gp}
    \begin{aligned}
        \mu_{GP;t}\left(\tilde {z}\right) = \bm{K}_t\left ( \tilde{z}\right )^{\top} \left(\tilde{K}_t+\sigma^2\bm{I}_t\right)^{-1} \tilde{\bm{y}}_t,\quad
        \sigma^2_{GP;t}\left(\tilde {z}\right) = \bm{K}\left ( \tilde{z}, \tilde{z}\right )   - \bm{K}_t\left ( \tilde{z}\right )^{\top} \left(\tilde{K}_t+\sigma^2\bm{I}_t\right)^{-1}\bm{K}_t\left ( \tilde{z}\right ) ,
    \end{aligned}
\end{equation}
where $\tilde{\bm{y}}_t = \left ( y^{(1)},\ldots,y^{(t)} \right )^{\top}\in \mathbb{R}^t$ is the vector of observations; $\tilde{K}_t\in\mathbb{R}^{t\times t}$ is the kernel matrix with $\left(\tau,\tau'\right)$-th entry $\bm{K}\left(z^{(\tau)},z^{\left(\tau'\right)}\right )$; and $\bm{K}_t\left ( \tilde{z}\right )   = \left ( \bm{K}\left(\tilde{z},z^{\left(1\right)}\right ),\bm{K}\left(\tilde{z},z^{\left(2\right)}\right ),\ldots,\bm{K}\left(\tilde{z},z^{\left(t\right)}\right ) \right )^{\top}\in \mathbb{R}^t$ is the vector of kernel function values between $\tilde{z}$ and $\left\{z^{(\tau)}\right\}_{\tau=1}^{t}$.

The GP model is generally regarded as a Bayesian nonparametric model. However, when the selected kernel function has a finite rank, that is $\bm{K}\left(z,z'\right) = \phi\left(z\right)^{\top}\phi\left(z'\right)$ for $\phi\left(z\right)\in \mathbb{R}^{p}$, the GP model has an alternative parametric representation $\bm{f}\left(z;\bm{W}\right) = \phi\left(z\right)^{\top} \bm{W}.$ Here, $\bm{W}\in \mathbb{R}^{p}$ denotes the unknown parameter with the prior distribution $\pi(\bm{W}) \stackrel{D}{ = } \mathcal{N}\left(\bm{0}_{p},\bm{I}_p\right)$, and the posterior distribution of the unknown parameters is \[
p\left(\bm{W}\mid\tilde{S}_t\right)\stackrel{D}{=}\mathcal{N}\left(\left(\Phi_t^\top \Phi_t+\sigma^2\bm{I}_p\right)^{-1} \Phi_t^\top \tilde{\bm{y}}_t,\sigma^2\left(\Phi_t^\top \Phi_t+\sigma^2\bm{I}_p\right)^{-1}\right),
\] where $\Phi_t=\left[\phi\left(z^{(1)}\right), \ldots, \phi\left(z^{(t)}\right)\right]^{\top}\in \mathbb{R}^{t\times p}$. In this way, inference for $\bm{f}\left(\tilde{z};\widehat{\bm{W}}\right) = \phi\left(\tilde{z}\right)^{\top} \widehat{\bm{W}} $ with the posterior distribution $\widehat{\bm{W}}\sim p\left(\bm{W}\mid\tilde{S}_t\right)$ matches the results in (\ref{eq.gp}). That is, GP is equivalent to a Bayesian parametric model, parametrized by $\bm{W}$ when the kernel function has a finite rank.

\subsection{Bayesian Neural Network}

Let $\bm{f}(z;\bm{W})$ be a neural network \citep{peng2022new,wang2023large} with $L$ hidden layers. Suppose the $l$-th layer contains $m_l$ neurons, where $m_0=d^{\text{in}}$ and $m_{L+1}=d^{\text{out}}$ are consistent with the dimensions of the inputs and outputs of the mapping to be approximated. In this way, the neural network $\bm{f}(z;\bm{W})$ is represented as
$$
\bm{f}(z;\bm{W})=W_{L}\cdot \varphi \left (  W_{L-1}\cdots\varphi\left ( W_0z +b_0 \right )\cdots +b_{L-1}  \right ) +b_L.
$$
Here, $W_l$ is the $m_{l+1}\times m_l$ weight matrix, and $b_{l}$ is the $m_{l+1}$-dimensional intercept. All these weight matrices and intercepts compose the weight parameters $\bm{W}$ of a neural network. The activation function $\varphi(\,\cdot\,)$ is defined on each entry (neuron) respectively and imposes non-linearity to the neural network. Common selections of activation functions include the rectified linear unit (ReLU) function and the tanh function. When the weight parameters are regarded as random variables and adopt a prior distribution, the neural network $\bm{f}(z;\bm{W})$ is known as a \textit{Bayesian neural network}.

\subsection{VAE-based Score Surrogate}
The VAE in the comparison in Appendix \ref{sec.e1} predicts scores from prompt representations $z$. Its encoder maps each input $z$ to a latent distribution $p\left(y\mid z,\bm{W}_{enc}\right)$, where $y$ is a random latent vector and $\bm{W}_{enc}$ denotes the encoder parameters. The decoder maps the sampled latent vector to an output score. For a fixed input $z$, latent samples $\left\{y_1,y_2,\ldots,y_K\right\}$ therefore yield a set of score predictions.

When the encoder and decoder are neural networks, their weights form the unknown model parameters. A Bayesian treatment places a prior on these parameters and updates their posterior using the observed scores, with variational inference providing an approximation. An alternative training approach minimizes the combined loss 
\begin{equation*}
\mathcal{L}_{\text {total }}=\mathcal{L}_{\text {regression }}+\beta \times \mathcal{L}_{K L},\end{equation*}
where $\mathcal{L}_{\text {regression }}$ is the loss between the observed scores and the predicted outputs, and $\mathcal{L}_{K L}$ is the Kullback--Leibler divergence between the learned latent distribution and a predefined distribution, usually standard normal. We refer to \cite{kingma2019introduction} for further details on VAE training.

\section{Additional Experimental Results \& Details}
\label{sec:additionalexp}
{
We provide implementation details and additional results for Section~\ref{sec.newexp}.

\subsection{Computing Environment and Baseline Details}
\label{sec:exp_settings}
Our experiments were conducted with PyTorch and Python 3.9 on a computer equipped with two AMD Ryzen Threadripper 3970X 32-Core Processors, 256 GB memory, and two Nvidia GeForce RTX 3090 GPUs with 24GB of RAM each.

\paragraph{Ranking and selection baselines.}
\begin{enumerate}

\item \textbf{Optimal Computing Budget Allocation (OCBA):} A fixed-budget R\&S algorithm that allocates more simulation resources to high-variance and near-best alternatives to asymptotically maximize the probability of correct selection; the allocation ratios are computed from current samples and updated sequentially \citep{chen2011stochastic}.

\item \textbf{Expected Value of Information (EVI):} A Bayesian allocation rule that approximately solves a finite-horizon sampling problem by choosing the next samples to maximize the expected value of information under a loss/budget model; commonly implemented in two-stage or fully sequential forms for R\&S \citep{alban2021expected}. 

\item \textbf{Knowledge Gradient (KG):} A sequential Bayesian policy that myopically samples the alternative with the largest expected one-step gain in the stop-and-implement value, i.e., the payoff from choosing the posterior-mean best under the current belief \citep{frazier2009knowledge}.
\end{enumerate}

Our implementations of these R\&S methods initialize each alternative under consideration with repeated evaluations to estimate its mean and variance. In our experiments, when the budget is insufficient to initialize all prompt representations (alternatives), we implement a subsampling stage. That is, we first randomly select a subset of the prompt representations with equal probability. Then the SO methods are implemented to choose from the subset of prompt representations. Specifically, the subsampling strategy is $N' = \min\left ( N, \left \lfloor \frac{T}{3}  \right \rfloor  \right ) $, where $T$ is the total budget, $N$ is the number of prompt representations, and $N'$ is the number of prompt representations in the subset. The proposed M-UCB procedure does not require a subsampling procedure. By constructing a surrogate model of the numerical prompt representations, the mean and variance of each prompt representation are approximated in the warm-up step (Appendix \ref{sec.warm}) and updated in the sequential evaluation step (Section \ref{sec.acquisition}).

\paragraph{Token optimization baselines.}
\begin{enumerate}
    \item \textbf{RLPROMPT:} \cite{deng2022rlprompt} formulate prompt selection as a reinforcement learning problem. A policy network generates tokens one at a time and receives a reward when a language model evaluates the entire prompt. The framework treats the large language model as a black‑box environment and allows flexible reward functions defined from task outputs.
    \item \textbf{Black‑Box Discrete Prompt Learning (BDPL):} \cite{diao2022black} frame prompt selection as optimizing a categorical distribution over discrete token indices. Each prompt position samples a token from a distribution parameterized by a vector. Because the language model is a black box and gradients are inaccessible, BDPL uses a variance‑reduced policy‑gradient estimator to update these distributions.
\end{enumerate}

\paragraph{LLM-driven prompt generation baselines.}
\begin{enumerate}
    \item \textbf{Automatic Prompt Engineer (APE):} \cite{zhou2022large} formulates prompt engineering as a black-box optimization guided by LLMs. APE iterates between proposal and evaluation: it generates instruction candidates using LLMs, evaluates them with score functions based on a validation dataset, filters the top-performing ones, and uses these high-scoring instructions to generate new similar candidates for the next iteration.
    \item \textbf{Optimization by PROmpting (OPRO):} \cite{yang2023large} treats LLMs as an iterative optimizer for prompts. OPRO operates by iteratively alternating between prompt generation and evaluation, where the language model generates new prompt candidates based on past results to improve the instructions progressively. Additionally, OPRO includes an ``optimization trajectory'' as part of its input, consisting of previous prompts and their scores.
\end{enumerate}

\subsection{Additional Anomaly Detection Results}
\label{sec:exp_anomaly}
For the applications, we instruct LLMs for time series anomaly detection. This task is routinely needed to monitor business metrics such as revenue, inventory, traffic, and customer support volumes, and is supported by enterprise services that detect and diagnose anomalies in operational evaluation criteria. These use cases are documented in provider materials for Microsoft Azure Anomaly Detector and Metrics Advisor \citep{Microsoft_AI_Anomaly_Detector, Microsoft_Anomaly_Detector_TN}, as well as in documentation for Amazon Lookout for Metrics \citep{AWS_Lookout_Anomalies}, which emphasizes business-metric anomalies and root-cause attribution. 

Table \ref{tab:abl_softprompts_budget_main_noisy_mixed} supplements Table \ref{tab:abl_softprompts_budget_main} with water-pump results. Classical R\&S methods are competitive when the budget is large relative to the candidate set, while the value of information sharing becomes clearer for larger sets. Figures \ref{fig:differentllms} and \ref{fig:differentllms2} compare M-UCB with OCBA across four LLMs on both datasets; M-UCB achieves higher F1 scores in each comparison. F1-score estimates may vary across experimental runs, even with the same candidate-set size and evaluation budget.

\begin{table}[!htbp]
\centering
\caption{Joint effects of number of prompt representations ($N$) and total budget ($T$) on the selected prompt performance for M-UCB and R\&S algorithms based on the water pump system failure dataset. Values show the mean F1 score with standard deviation in parentheses. The employed LLM is Claude 3.5 Sonnet.}
\label{tab:abl_softprompts_budget_main_noisy_mixed}
\begin{threeparttable}
\begin{tabular}{@{}lS[table-format=1.2]@{\,}l@{\quad}S[table-format=1.2]@{\,}l@{\quad}S[table-format=1.2]@{\,}l@{\quad}S[table-format=1.2]@{\,}l@{}}
\toprule
$(N, T)$ & \multicolumn{2}{c}{M-UCB} & \multicolumn{2}{c}{OCBA} & \multicolumn{2}{c}{EVI} & \multicolumn{2}{c}{KG} \\
\midrule
(20, 50)	 & 0.59 & (0.038) & 0.57 & (0.029) & 0.56 & (0.034) & 0.55 & (0.023) \\
(20, 100)	 & 0.60 & (0.031) & 0.60 & (0.032) & 0.57 & (0.030) & 0.57 & (0.022) \\
(20, 150)	 & 0.60 & (0.027) & 0.61 & (0.022) & 0.57 & (0.025) & 0.61 & (0.020) \\
(20, 250)	 & 0.62 & (0.013) & 0.65 & (0.019) & 0.60 & (0.021) & 0.61 & (0.022) \\

\\[0.5em]
(100, 50)	 & 0.60 & (0.034) & 0.58 & (0.037) & 0.54 & (0.039) & 0.56 & (0.032) \\
(100, 100)	 & 0.62 & (0.033) & 0.60 & (0.034) & 0.61 & (0.037) & 0.58 & (0.030) \\
(100, 150)	 & 0.64 & (0.031) & 0.62 & (0.033) & 0.60 & (0.035) & 0.59 & (0.029) \\
(100, 250)	 & 0.68 & (0.030) & 0.63 & (0.030) & 0.63 & (0.030) & 0.61 & (0.028) \\

\\[0.5em]
(200, 50)	 & 0.60 & (0.054) & 0.57 & (0.031) & 0.59 & (0.039) & 0.60 & (0.036) \\
(200, 100)	 & 0.65 & (0.045) & 0.60 & (0.017) & 0.59 & (0.032) & 0.63 & (0.038) \\
(200, 150)	 & 0.66 & (0.049) & 0.63 & (0.038) & 0.63 & (0.036) & 0.63 & (0.034) \\
(200, 250)	 & 0.67 & (0.035) & 0.64 & (0.037) & 0.64 & (0.033) & 0.64 & (0.026) \\

\\[0.5em]
(500, 50)	 & 0.59 & (0.051) & 0.56 & (0.047) & 0.55 & (0.032) & 0.56 & (0.032) \\
(500, 100)	 & 0.61 & (0.041) & 0.60 & (0.044) & 0.59 & (0.043) & 0.58 & (0.035) \\
(500, 150)	 & 0.62 & (0.033) & 0.61 & (0.042) & 0.63 & (0.042) & 0.59 & (0.027) \\
(500, 250)	 & 0.65 & (0.050) & 0.62 & (0.046) & 0.63 & (0.034) & 0.60 & (0.033) \\
\bottomrule
\end{tabular}
\end{threeparttable}
\end{table}

\begin{figure}[!htbp]
    \centering
    \includegraphics[width=0.7\linewidth]{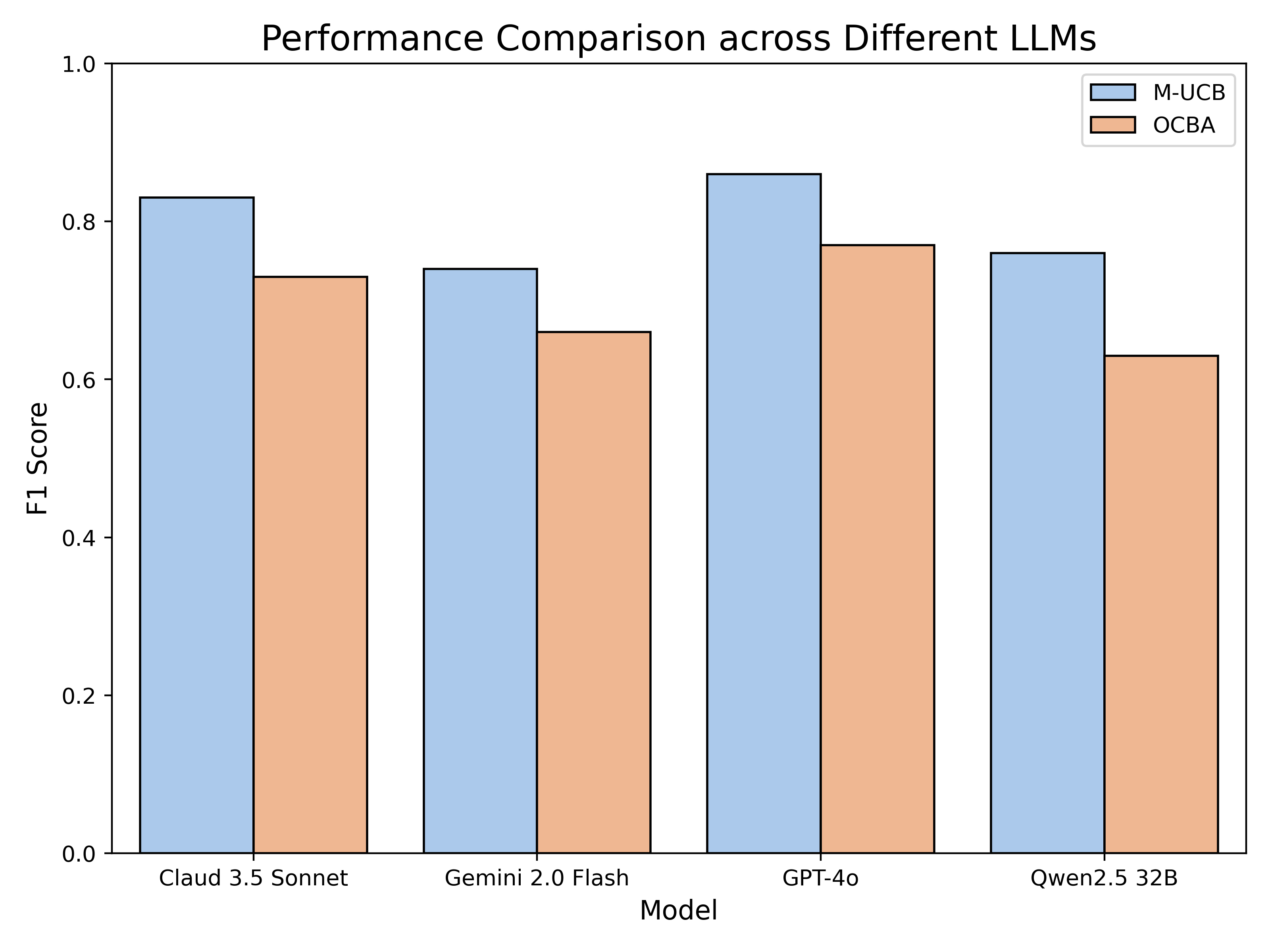}
    \caption{The F1 scores of M-UCB and OCBA based on the credit card fraud detection with $N=500$ prompt representations and $T=250$ evaluation calls. The employed LLMs include 1) Claude 3.5 Sonnet, 2) Gemini 2.0 Flash, 3) GPT-4o, and 4) Qwen2.5 32B.}
    \label{fig:differentllms}
    \Description{Grouped bars compare mean F1 scores of M-UCB and OCBA for credit-card fraud detection across Claude 3.5 Sonnet, Gemini 2.0 Flash, GPT-4o, and Qwen 2.5 32B. M-UCB has the higher mean for each model.}
\end{figure}

\begin{figure}[!htbp]
    \centering
    \includegraphics[width=0.7\linewidth]{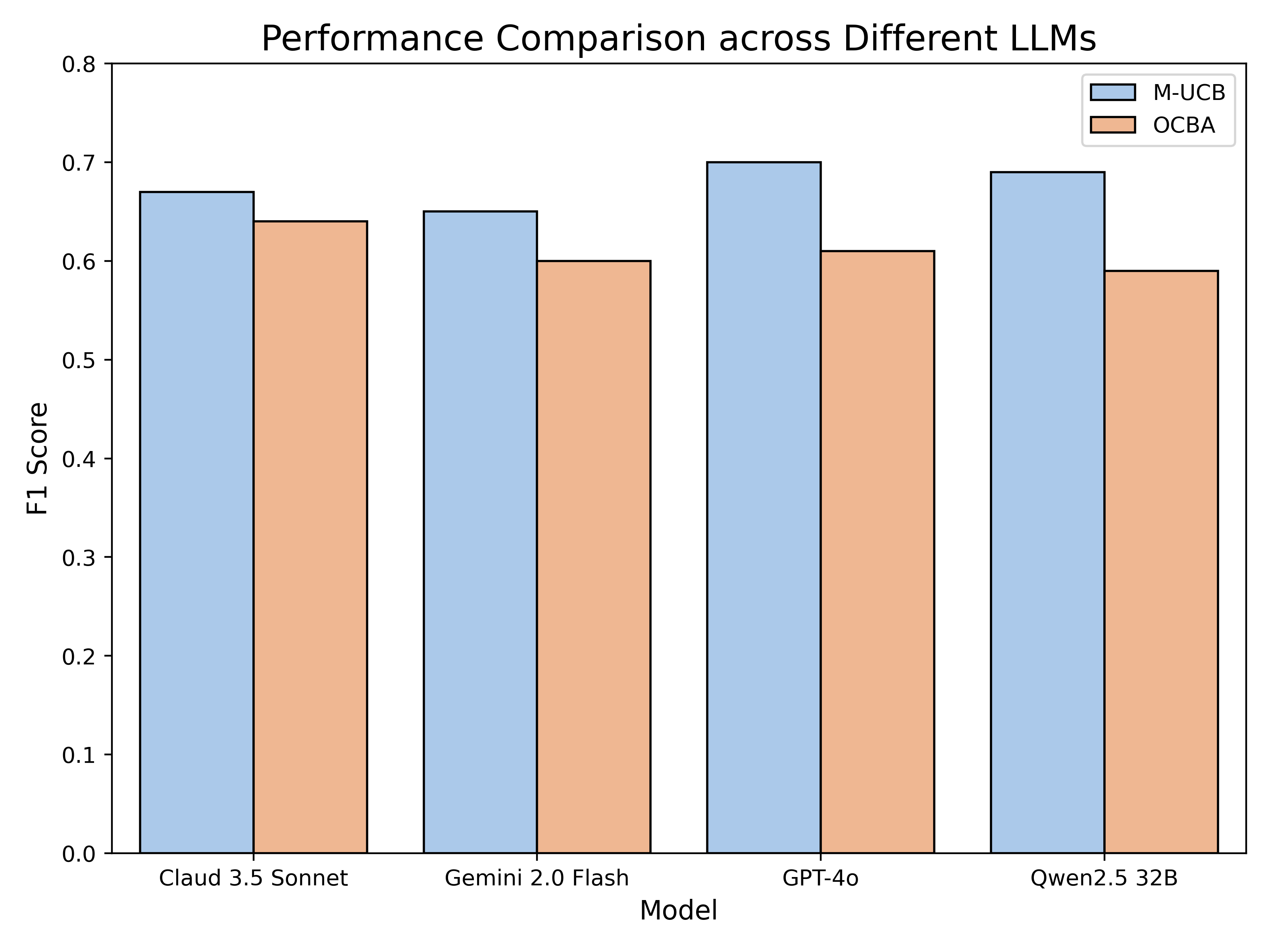}
    \caption{The F1 scores of M-UCB and OCBA based on the water-pump system failure dataset with $N=500$ prompt representations and $T=250$ evaluation calls. The employed LLMs include 1) Claude 3.5 Sonnet, 2) Gemini 2.0 Flash, 3) GPT-4o, and 4) Qwen2.5 32B.}
    \label{fig:differentllms2}
    \Description{Grouped bars compare mean F1 scores of M-UCB and OCBA for water-pump status prediction across Claude 3.5 Sonnet, Gemini 2.0 Flash, GPT-4o, and Qwen 2.5 32B. M-UCB has the higher mean for each model.}
\end{figure}

\clearpage
\subsection{Additional Optimization-Problem Results}
\label{sec:exp_industry}
Figure \ref{fig:tr} reports the score trajectories for M-UCB-r (last 200), RLPROMPT, and BDPL. M-UCB-r improves consistently and attains higher scores within the 1000-call budget, while the two token optimization methods continue to improve near the end of the budget.

\begin{figure}[!htbp]
    \centering
    \includegraphics[width=\linewidth]{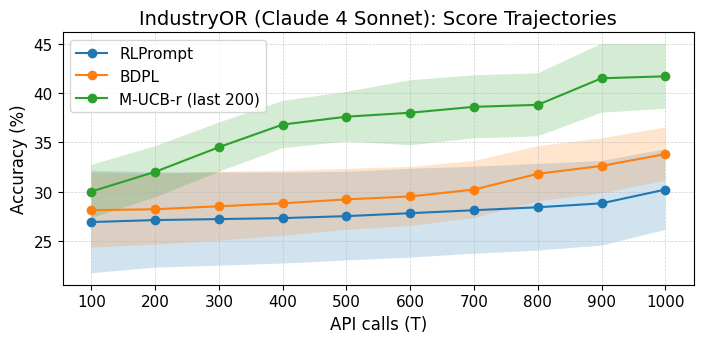}
    \caption{The score trajectories of the compared token-space black-box optimization methods. The solid line is the mean value of the prompt score calculated using the held-out test dataset, and the shadow represents the standard deviation.}
    \label{fig:tr}
    \Description{Line chart of accuracy against the number of API calls for the IndustryOR task with Claude 4 Sonnet. The curves compare RLPrompt, BDPL, and M-UCB-r, with shaded variability bands. M-UCB-r has the highest mean accuracy at the plotted budgets.}
\end{figure}

For the second set of experiments in Section \ref{sec.expbbo} for instructing LLMs to solve optimization problems, we present the accuracies of the LLMs' output in Table \ref{tab:industryor_difficulty_stratified}, where the optimization problems are stratified by difficulty \citep{orlm2024}. The results reveal that the performance of the selected prompt is highly dependent on the optimization problem itself, and our experiments are consistent with the difficulty levels pre-specified in the dataset: the more difficult the optimization problems are, the worse the selected prompts perform. These experimental results further show that solely relying on LLMs' textual reasoning capabilities to solve optimization problems is currently unsatisfactory. Instead, instructing LLMs to collaborate and apply external tools (e.g., model context protocol) for optimization showcases strong potential \citep{jiang2025llmopt}. This further underscores the importance of prompts, which explicitly guide LLMs' complex workflows. We leave this discussion to future work.

\begin{table}[t]
\centering
\small  %
\caption{IndustryOR results stratified by difficulty (Easy/Medium/Hard) under a fixed budget of 1000 API calls. The recorded metric is the accuracy of LLMs' output under the instructions from the selected prompt.}
\label{tab:industryor_difficulty_stratified}
\vspace{4pt}
\setlength{\tabcolsep}{4pt}  %

\begin{subtable}{\textwidth}
\centering
\caption{Easy subset}
\label{tab:industryor_easy}
\vspace{2pt}
\begin{tabular}{l*{7}{c}}
\toprule
LLM & RLPROMPT & BDPL & M-UCB & \makecell{M-UCB-r\\(100)} & \makecell{M-UCB-r\\(400)} & \makecell{\textbf{M-UCB-r}\\(200)} & PSK-only \\
\midrule
Claude 4 & 48.1 & 51.7 & 55.2 & 55.7 & 55.9 & \textbf{57.2} & 53.9 \\
Claude 3.5 & 44.5 & 48.3 & 49.3 & 49.9 & 50.3 & \textbf{53.1} & 48.2 \\
\bottomrule
\end{tabular}
\end{subtable}
\vspace{8pt}

\begin{subtable}{\textwidth}
\centering
\caption{Medium subset}
\label{tab:industryor_medium}
\vspace{2pt}
\begin{tabular}{l*{7}{c}}
\toprule
LLM & RLPROMPT & BDPL & M-UCB & \makecell{M-UCB-r\\(100)} & \makecell{M-UCB-r\\(400)} & \makecell{\textbf{M-UCB-r}\\(200)} & PSK-only \\
\midrule
Claude 4 & 28.3 & 31.9 & 35.6 & 35.9 & 36.1 & \textbf{38.6} & 33.7 \\
Claude 3.5 & 25.2 & 27.8 & 27.4 & 30.1 & 30.6 & \textbf{32.1} & 28.5 \\
\bottomrule
\end{tabular}
\end{subtable}
\vspace{8pt}

\begin{subtable}{\textwidth}
\centering
\caption{Hard subset}
\label{tab:industryor_hard}
\vspace{2pt}
\begin{tabular}{l*{7}{c}}
\toprule
LLM & RLPROMPT & BDPL & M-UCB & \makecell{M-UCB-r\\(100)} & \makecell{M-UCB-r\\(400)} & \makecell{\textbf{M-UCB-r}\\(200)} & PSK-only \\
\midrule
Claude 4 & 8.1 & 11.7 & 12.2 & 14.7 & 15.4 & \textbf{19.8} & 13.9 \\
Claude 3.5 & 6.7 & 7.6 & 8.3 & 8.6 & 10.3 & \textbf{10.7} & 8.5 \\
\bottomrule
\end{tabular}
\end{subtable}
\end{table}

\clearpage
\subsection{Preference-Judging Implementation and Additional Results}
\label{sec:exp_preference}
The WebGPT data take the form $\left(x_i, y_i^{(1)}, y_i^{(2)}\right)$, where $x_i$ is a user query and $y_i^{(1)}$ and $y_i^{(2)}$ are two responses. Without loss of generality, let $y_i^{(1)}$ be the response preferred by humans.

For this format of human-feedback data, we consider the observed score: 
\[
\hat{v}_i(\p)=\log p\left(y_i^{(1)} \mid x_i, \p\right)-\log p\left(y_i^{(2)}\mid x_i, \p\right) ,
\]
where the LLM is required to choose its preferred answer from the two candidate responses
\(y_i^{(1)}\) and \(y_i^{(2)}\).  
Here, $\log p\left(y_i^{(1)} \mid x_i, \p\right)$ denotes the log-probability that the model selects the preferred response. For the implementation, we use two tokens ``A'' and ``B'' to represent the two responses, and we force a one-token answer (``A'' or ``B'') via API. The log-probabilities for both tokens are then given via the API output; see details in \cite{hills_anadkat_using_logprobs_2023}. This evaluation criterion can be generalized to human preference datasets with a sequence of preferences instead of pairwise preferences \citep{rafailov2023direct}.

In business applications, organizations routinely maintain proprietary datasets of human feedback. For example, quality-assurance (QA) software for contact centers lets operations teams systematically score, annotate, and coach on agent–customer interactions and run pairwise preference evaluations of alternative replies \citep{boyer_quality_2025}; specialized deployments also leverage private human-rated data when aligning models to domain needs (e.g., finance-specific systems such as BloombergGPT trained on proprietary data \citep{wu2023bloomberggpt}). Although human feedback data is more often used to fine-tune language models, leveraging it to select prompts offers a more cost-effective strategy for applications with limited dataset sizes.

Table \ref{tab:call-limits} compares the methods under fixed call limits and reports the tokens consumed. At 10 calls, M-UCB achieves the highest score with the fewest tokens. At 50 calls, APE and OPRO achieve higher scores while consuming more tokens, complementing the shared-token-budget comparison in Table \ref{tab:token-budgets}.

\begin{table}[ht]
  \centering
  \caption{Comparison of methods when each is allowed a fixed number of LLM calls.}
  \begin{tabular}{lcccc}
    \toprule
    & \multicolumn{2}{c}{10 calls} & \multicolumn{2}{c}{50 calls} \\
    \cmidrule(lr){2-3}\cmidrule(lr){4-5}
    Method & Score & Tokens Used &  Score & Tokens Used \\
    \midrule
    M-UCB & $1.21$ & $\sim 0.6\,\mathrm{M}$ & $1.55$ & $\sim 3.1\,\mathrm{M}$ \\
    APE   & $1.13$ & $\sim 1.7\,\mathrm{M}$ & $1.86$ & $\sim 8.6\,\mathrm{M}$ \\
    OPRO  & $1.10$ & $\sim 2.1\,\mathrm{M}$ & $2.02$ & $\sim 15.5\,\mathrm{M}$ \\
    \bottomrule
  \end{tabular}
  \label{tab:call-limits}
\end{table}

For the third set of experiments (Section \ref{sec.expgpt}), we present representative prompts generated by different methods to illustrate their distinct prompting strategies. Specifically, \emph{OPRO produces structured, formal instructions} that emphasize systematic assessment of answer quality, since the optimization trajectory (pairs of prompts and performances) is fed into LLMs to learn and make decisions. Representative prompts follow.
\begin{quote}
\small\itshape
\begin{itemize}[leftmargin=*,topsep=5pt,itemsep=3pt]

\item Evaluate both answers to this question carefully by examining their accuracy, relevance, completeness, and informativeness. Consider the quality of the cited sources and how well each answer is supported by the web search evidence provided. Determine which answer better addresses the question while maintaining factual correctness and clarity. Assign a preference score indicating which response is superior overall.

\item Analyze the two candidate answers by assessing their factual correctness against the provided source citations. Evaluate the coherence and logical flow of each response. Consider whether each answer directly addresses the user's query with sufficient depth. Based on comprehensive evaluation criteria, determine which answer demonstrates higher quality and deserves a stronger preference score.

\item Compare the provided answers by examining the source material quality and relevance to each response. Assess how well each answer synthesizes information from the web search results. Evaluate clarity, accuracy, and informativeness of both responses. Determine overall answer quality and assign a preference rating that reflects which answer better satisfies the user's information needs.

\item Review both answers to the question with focus on factual accuracy supported by cited sources. Evaluate each answer's ability to provide comprehensive coverage of the topic. Assess the presentation quality and readability of each response. Based on these criteria, determine which answer is more helpful and assign an appropriate preference score reflecting relative quality.

\item Examine both answers systematically by assessing source utilization, factual accuracy, topical relevance, and response coherence. Determine whether citations properly support claims made in each answer. Evaluate the completeness of information provided. Calculate an overall quality assessment and provide a preference score indicating which answer better serves the user's information requirements.

\end{itemize}
\end{quote}

In comparison, M-UCB selects more \emph{exploratory, intuition-driven prompts}:
\begin{quote}
\small\itshape
\begin{itemize}[leftmargin=*,topsep=5pt,itemsep=3pt]
\item
Check both answers for factual correctness and see if they address user's request properly. Assess whether the operational steps are complete and thorough. How is the evidence being used though? Notice that wording in this section part part seems to repeat oddly while maintaining semantic meaning simultaneously. After your examination finishes, which answer is more strong? Provide a preference score between 1 to 10.

\item
Examine accuracy levels in both answers. Do they respond to what the user originally intended? Include whether critical understanding steps are present in each explanation. Sources should be inspected, but one sentence maybe not connecting connecting smoothly to the rest of the paragraph flow. Choose the better-performing reply when done and assign preference scoring accordingly based on criteria.

\item
Review clarity and informativeness across both responses. Can they guide user toward correct action paths effectively? Observe how cited evidence gets applied, though words misplaced in one line not fully making logic appears confusing here. Determine which answer is superior and output preference indicator now.

\item
Exam the reasoning strength and detail coverage in two answers provided. Do they offer actionable, usable instructions which people can follow without confusion? Citation usage requires observation carefully, but note one phrase phrase is duplicated and grammar not ideal throughout entire response section. When evaluation is complete, identify which performs better overall and specify preference mark score.

\item
Examine both answers, consider in consistency, correctness of core claims, and usefulness of recommended steps forward. How is evidence referenced by each response individually? Notice one sentence structure slightly broken—it's not reading readly smooth and lack of coherence may suffer considerable impactful effects. State which answer is preferable after full comparison analysis and include the preference score calculation result.

\item
Evaluate responses based on their ability to synthesize information correctly. Both answers should demonstrate understanding of concepts, although pronoun references may be ambiguous or unclear sometimes. The explanations lack sufficient depth in places where elaboration needed more details. Which response demonstrates superior comprehension overall?

\item
Asses and choose from two answers must address completeness of coverage, but some sections sections appear to have missing context that would benefit from additional explanation. Factual accuracy should be prioritized, yet inconsistencies in reasoning patterns can undermine credibility. Determine scoring preference metric based on overall performance evaluation holistically.

\end{itemize}
\end{quote}

The prompts selected by M-UCB include less structured sentences and some grammatical errors. Together with the token-budget results in Section \ref{sec.expgpt}, these examples illustrate the complementary strengths of surrogate-guided evaluation and LLM-driven prompt generation. Combining them is a direction for future work.

}

\section{Additional Computational Studies}
\label{sec.ablation}
In this section, we present the numerical experiments of our proposed framework for selecting prompts. Regarding the generative language models for evaluating the prompt selection procedure, we select 1) GPT-3.5-turbo and 2) text-davinci-003; details about the background of these two generative language models can be found at \url{https://platform.openai.com/docs/models}. We implement our prompt selection procedure to select the prompt for language models to accomplish following tasks: 1) Word sorting: Given a list of words as input, output the sorted list in alphabetical order; 2) Rhymes: Given a word $A$ and a list of words as input, output the word from the list that rhymes with $A$; 3) First letter: Given a word as input, output its first letter; 4) Largest animals: Given a list of animals as input, identify the largest animal; 5) Nums to verbal: Given a numerical number as input, output its English word form; 6) Count objects: Given a list of objects as input, output the number of objects in the list. Since all the tasks in this section are text-based, we select cosine similarity as the score function.

Recall that, in the first search stage of our proposed prompt selection procedure, the prompt in text form is first transformed into a latent vector $X_n\in\mathbb{R}^{\tilde{D}}$ and then a prompt representation $z_n\in\mathbb{R}^{D}$. In our experiments, the dimension of the latent vector is $\tilde{D} = 3584$, and the dimension of the prompt representation is $D=50$. Upon the completion of the search stage, we have a prompt representation set $\mathcal{Z} = \left\{z_1,z_2,\ldots,z_N\right\}$ which contains finite but sufficient vectors, serving as the feasible set of the prompt selection problem. In the subsequent evaluation and selection stage, we sequentially select prompts in the prompt representation set to evaluate. The sample size of the prompt representation set and the total budget for evaluation will be specified in each experiment. The computing environment is described in Appendix \ref{sec:exp_settings}. The experimental results presented here are mean performances based on 15 repetitions, with standard deviations represented by a shadow on the mean-value line.

We evaluate each step of the proposed prompt selection procedure in our experiments. Specifically, in Section \ref{sec.model}, we propose to use the Bayesian parametric model as the surrogate model to approximate the dependence of the mean score on the prompt representation. We then select four Bayesian parametric models and compare their approximation accuracy in Section \ref{sec.e1}. In Sections \ref{sec.acquisition} and \ref{sec.acqimplementation}, we describe two acquisition functions: 1) modified upper confidence bound (M-UCB) and 2) probabilistic reparametrization modified upper confidence bound (PR-M-UCB) to sequentially select the prompt to evaluate. We compare the optimization procedure associated with the acquisition functions in Section \ref{sec.e2}. In Section \ref{sec.refinment}, we also propose a projection stochastic kriging (PSK) model to refine the proposed two-stage framework with the observed scores after completing the evaluation and selection stage. Then in Section \ref{sec.e3}, we compare our two-stage framework with the direct search using the PSK model.

\subsection{Surrogate Model Comparison}
\label{sec.e1}
In this section, we conduct numerical experiments to compare the approximation accuracy of surrogate models on approximating the mean score with respect to prompt representations. Specifically, we choose 1) a Bayesian linear regression model (BLR), 2) a Gaussian process (GP) with a finite-rank kernel, 3) a Bayesian neural network (BNN), and 4) a variational autoencoder (VAE) as the surrogate model. The surrogate models are described in Appendix \ref{sec.surrogates}. These selected surrogate models are all represented by a general parametric form $\bm{f}\left(z;\bm{W}\right)$, where $z$ is the prompt representation and $\bm{W}\in\mathbb{R}^{p}$ is the unknown parameter of the surrogate model assigned with a prior distribution $\bm{W}\sim \pi\left(\bm{W}\right)$. In this set of experiments, the prior distribution for each surrogate model is set as the independent standard normal distribution.

For each set of experiments, we first fix a prompt representation set $\mathcal{Z} = \left\{z_1,z_2, \ldots,z_N\right\}$. We then randomly select $N_{\operatorname{tr}} = 0.7N$ prompt representations from $\mathcal{Z}$ with equal probabilities, and denote the set of selected prompt representations as $\mathcal{Z}_{\operatorname{tr}}$. For each $z_n\in\mathcal{Z}_{\operatorname{tr}}$, we observe the scores as in (\ref{eq.generating}) for $5$ times. We denote the training set as $\mathcal{S}_{\operatorname{tr}} = \left\{\left(z_n, \widehat{v}_{n,m}\right)\right\}$ for $m\in\left\{1,2,\ldots,5\right\}$ for $z_n\in\mathcal{Z}_{\operatorname{tr}}$, where $\widehat{v}_{n,m}$ denotes the $m$-th observation of the score evaluating with the prompt representation $z_n$. We let $\mathcal{Z}_{\operatorname{test}} = \mathcal{Z}\setminus \mathcal{Z}_{\operatorname{tr}}$. For each $z_n\in \mathcal{Z}_{\operatorname{test}}$, we evaluate it and observe the scores for $\tilde{r} = 50$ times. We let $\mathcal{S}_{\operatorname{test}} = \left\{\left(z_n,\bar{v}_{n}\right)\right\}$ for $z_n\in \mathcal{Z}_{\operatorname{test}}$ and $\bar{v}_n=\frac{1}{\tilde{r}}\sum_{m=1}^{\tilde{r}}\widehat{v}_{n,m}$, the sample mean of the observed score. In each set of experiments, we use the same training set $\mathcal{S}_{\operatorname{tr}}$ to calculate the posterior distribution of the unknown parameter $p\left(\bm{W}\mid \mathcal{S}_{\operatorname{tr}}\right)$ as in (\ref{eq.posterior}). We then generate $\left\{\widehat{\bm{W}}_1,\widehat{\bm{W}}_2,\ldots,\widehat{\bm{W}}_K\right\}$ from the posterior distribution $p\left(\bm{W}\mid \mathcal{S}_{\operatorname{tr}}\right)$, where $K=100$. Regarding the comparison metric of the surrogate models' performances, we consider 1) root mean squared error (RMSE) and 2) covering ratio (CR). That is 
\begin{equation*}
    \operatorname{RMSE} = \sqrt{\frac{1}{N_{\operatorname{test}}}\sum_{z_n\in\mathcal{Z}_{\operatorname{test}}}\left(\widehat{\mu}\left(z_n\right)-\bar{v}_{n}\right)^2},\quad\quad
    \operatorname{CR} = \frac{1}{N_{\operatorname{test}}}\sum_{z_n\in\mathcal{Z}_{\operatorname{test}}}\mathbb{I}\left\{\bar{v}_n\in\left[\mathcal{L}_n,\mathcal{U}_n\right]\right\},
\end{equation*}
where $N_{\operatorname{test}}$ is the sample size of the test dataset $\mathcal{Z}_{\operatorname{test}}$, $\widehat{\mu}\left(z_n\right) = \frac{1}{K}\sum_{k=1}^K\bm{f}\left(z_n;\widehat{\bm{W}}_k\right)$ and $\left[\mathcal{L}_n,\mathcal{U}_n\right]$ is a 90\% interval estimated from the posterior samples $\left\{\bm{f}\left(z_n;\widehat{\bm{W}}_k\right)\right\}_{k=1}^{K}$. We also consider the impact of the sample size on the surrogate models' performances, with $N\in \left\{50,100,200,500\right\}$.

\begin{figure}[htbp]
\centering
\begin{minipage}[t]{0.49\textwidth}
\centering
\includegraphics[width=\linewidth]{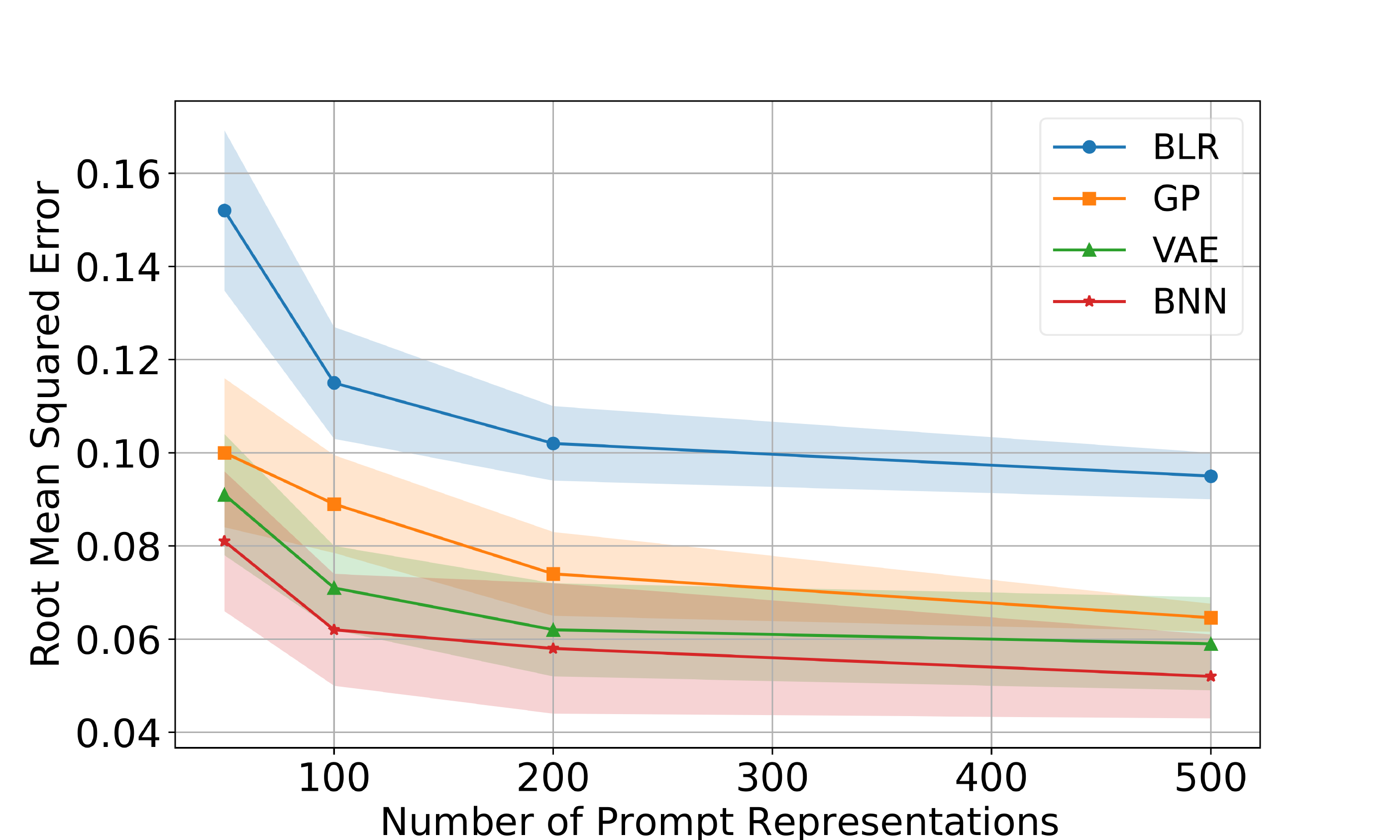}
\end{minipage}
\begin{minipage}[t]{0.49\textwidth}
\centering
\includegraphics[width=\linewidth]{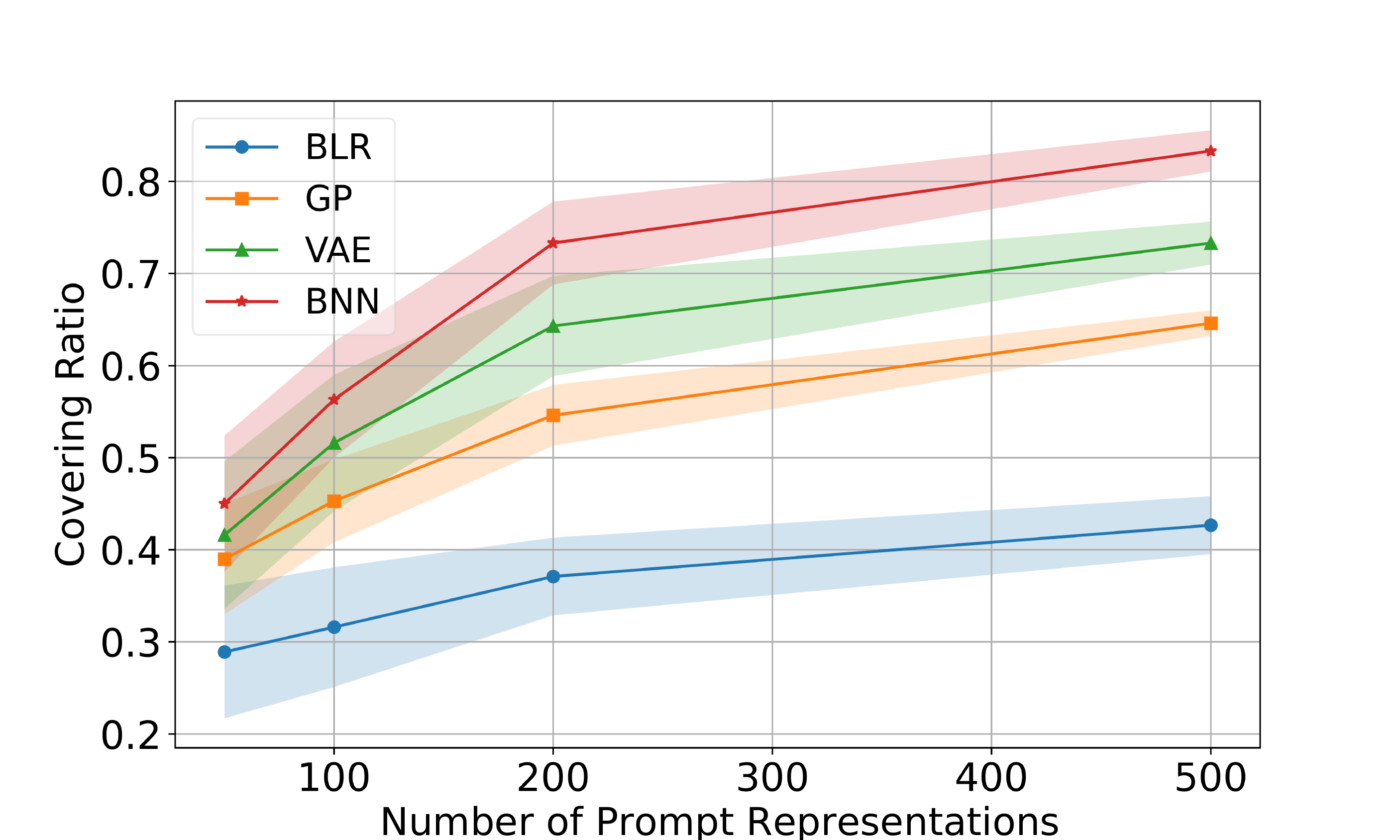}
\end{minipage}
\caption{Experimental results for approximating the mean score with the prompt representations using different Bayesian parametric models. The task is word sorting and the generative language model is text-davinci-003.}
\label{fig.exp1}
    \Description{Two line charts compare surrogate root mean squared error and coverage ratio as the number of prompt representations increases on the word-sorting task with text-davinci-003. The models are BLR, GP, VAE, and BNN.}
\end{figure}

\begin{figure}[htbp]
\centering
\begin{minipage}[t]{0.49\textwidth}
\centering
\includegraphics[width=\linewidth]{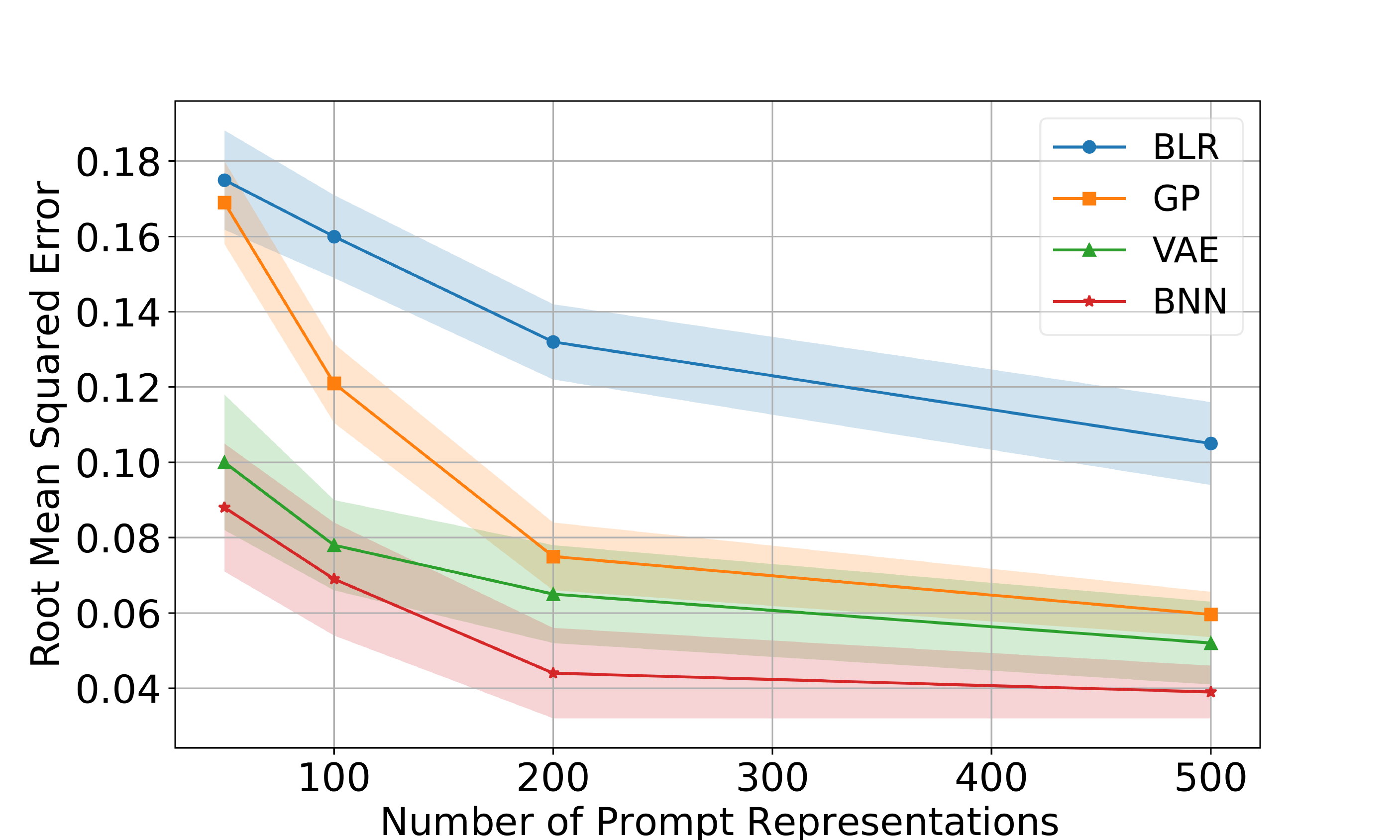}
\end{minipage}
\begin{minipage}[t]{0.49\textwidth}
\centering
\includegraphics[width=\linewidth]{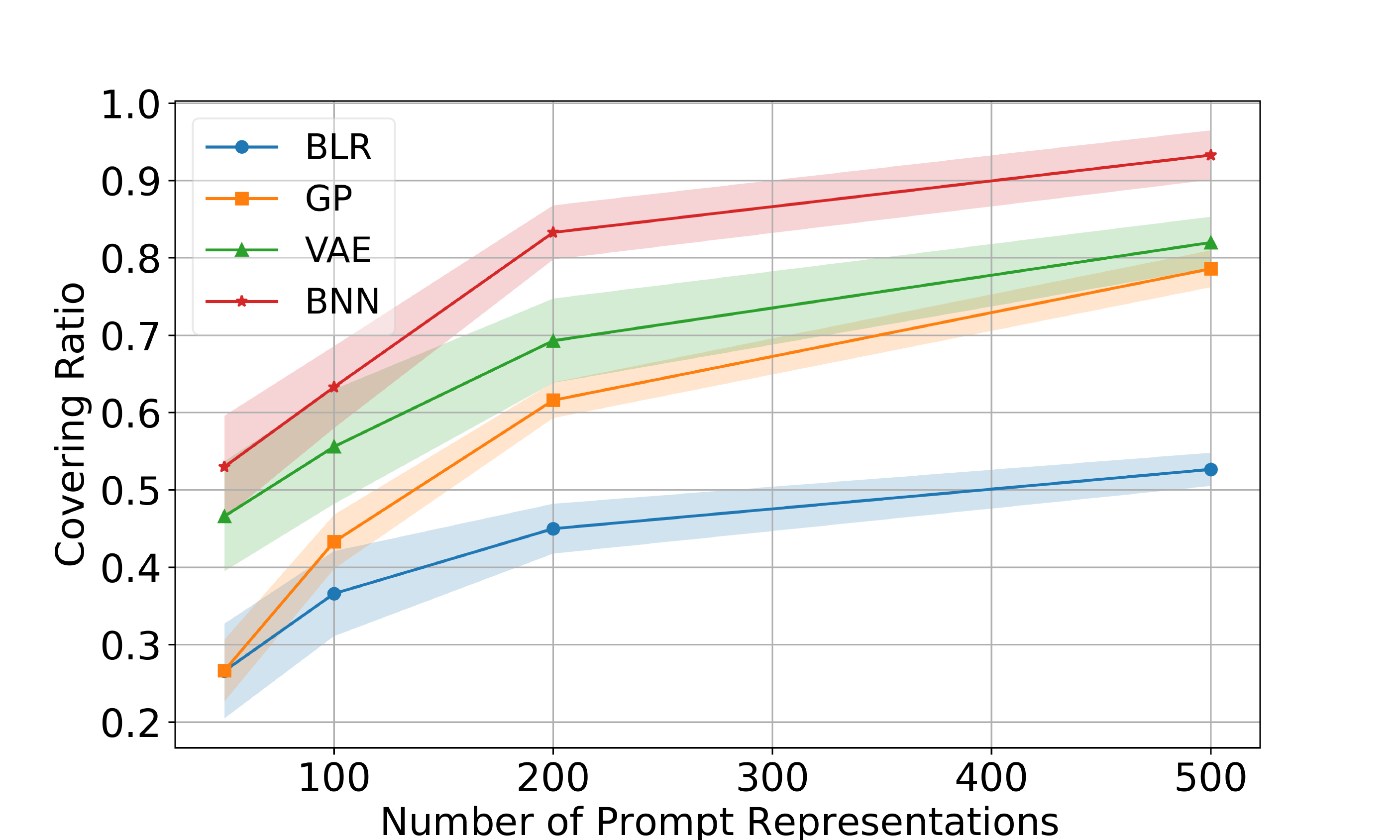}
\end{minipage}
\caption{Experimental results for approximating the mean score with the prompt representations using different Bayesian parametric models. The task is word sorting and the generative language model is GPT-3.5-turbo.}
\label{fig.exp11}
    \Description{Two line charts compare surrogate root mean squared error and coverage ratio as the number of prompt representations increases on the word-sorting task with GPT-3.5-turbo. The models are BLR, GP, VAE, and BNN.}
\end{figure}

The experiment results in {Figure \ref{fig.exp1}} and {Figure \ref{fig.exp11}} provide the following insights on the comparison among surrogate models. First, when there are not sufficient observations to construct the surrogate model ($N=50$), the difference in the performances of surrogate models on RMSE and CR is not significant. In general, these surrogate models all result in relatively high approximation error (indicated by high RMSE) and biased approximation confidence (indicated by low CR). This illustrates that constructing a surrogate model for appropriate use requires sufficient observations. Second, as the number of observed scores increases, all the surrogate models achieve better approximation performances, indicated by lower RMSE and higher CR. On the other hand, the improvement in the BLR model's performance is not as significant as in other models. That is, BLR cannot capture the dependence of the mean score on the prompt representation because of the lack of expressive power. Lastly, among all experiments, BNN achieves the more preferable performance with the lowest RMSE and highest CR. This suggests that the BNN configuration used here is well suited to approximating the dependence of the mean score on the prompt representation. On the other hand, VAE, as another deep learning model, does not perform as well as BNN in our experiments. For the VAE configuration used here, prediction accuracy remains below that of BNN even when $N=500$. For these reasons, we conclude that BNN achieves the more preferable performance among the selected surrogate models for approximating the mean score of the prompt representation, and we select BNN as the surrogate model for approximating the mean score function in other experiments, as well as those presented in Section \ref{sec.newexp}.

\subsection{Acquisition Function Optimization}
\label{sec.e2}
In this section, we compare the two acquisition functions used in the sequential evaluation step in our framework: M-UCB and PR-M-UCB in terms of their performances on 1) the mean score of the selected prompt and 2) implementation time. Recall that M-UCB is $\alpha_t\left(z_n\right) =\mu_{t}\left ( z_n \right ) +\beta_t\left(\sigma_{t}\left(z_n\right)+\gamma\left(r_{n}\left(t\right)\right)\right)$, where $\mu_{t}\left ( z_n \right )$ and $\sigma_t\left(z_n\right)$ are the posterior mean and standard deviation of the surrogate model at $z_n$, and $r_n(t)$ denotes the number of evaluations at $z_n$ up to time $t$. In this set of experiments, we set $\beta_t =\sqrt{ 2\log(t)}$ and $\gamma\left(m\right) = \frac{2}{\sqrt{m+1}}$. Since maximizing M-UCB requires calculating the acquisition value for each $z_n\in\mathcal{Z}$, when the total number of prompt representations $N$ is large, it is time-consuming to select the next prompt representation to evaluate. We also consider maximizing PR-M-UCB $\widetilde{\alpha}_t(\theta)\doteq\mathbb{E}_{p(z_n;\theta)}\left[\alpha_t\left(z_n\right)\right]$, where $p(z_n;\theta)$ is a categorical distribution as described in Section \ref{sec.acqimplementation}. Since it is a continuous optimization problem in $\theta$, we employ the method of gradient ascent with multiple starting points. Specifically, the algorithm using the gradient ascent method is represented by PR-M-UCB $(\mathfrak{M},\mathfrak{T})$, where $\mathfrak{M}$ is the number of starting points and $\mathfrak{T}$ is the number of gradient ascent iterations.

\begin{figure}[htbp]
\centering
\begin{minipage}[t]{0.49\textwidth}
\centering
\includegraphics[width=\linewidth]{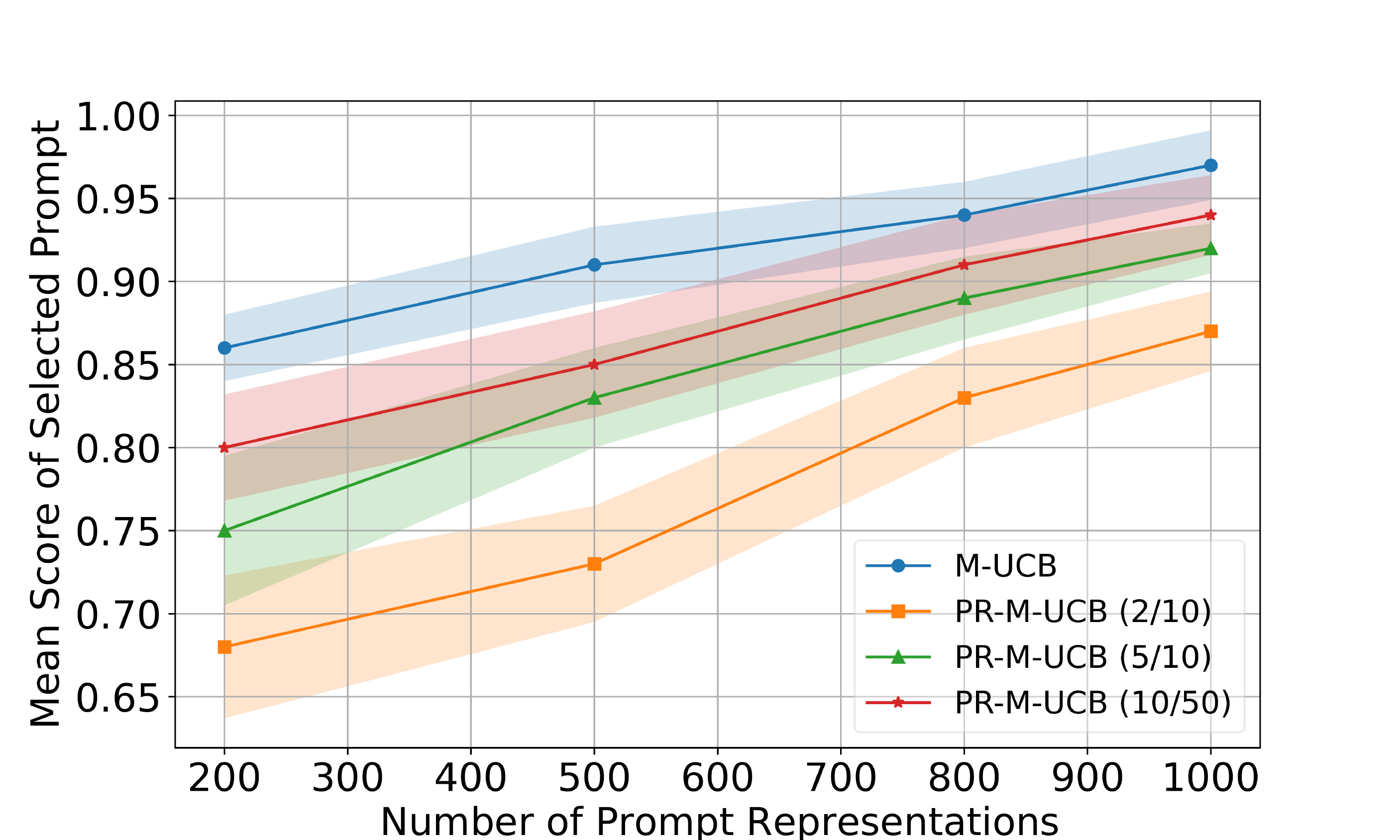}
\end{minipage}
\begin{minipage}[t]{0.49\textwidth}
\centering
\includegraphics[width=\linewidth]{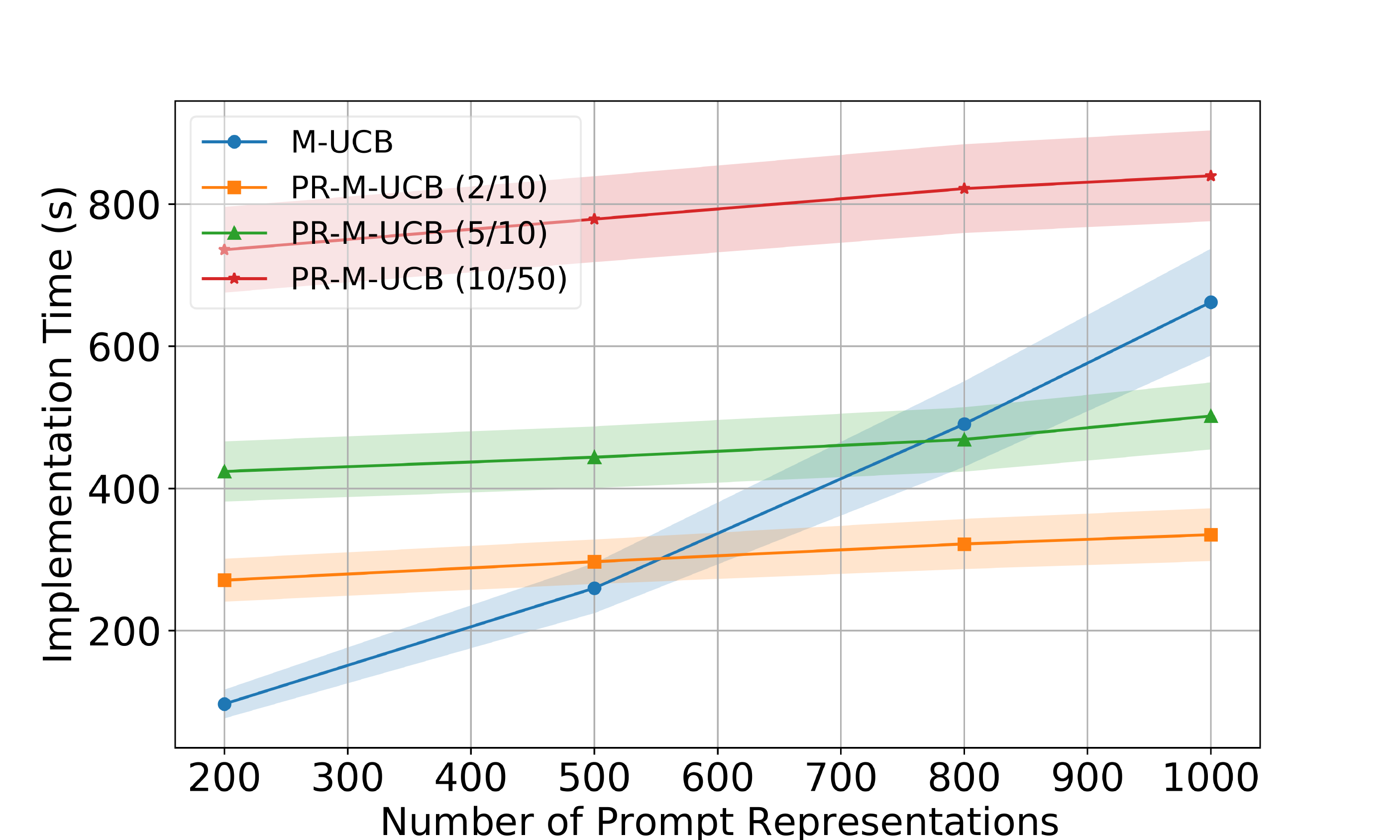}
\end{minipage}
\caption{Experimental results for comparison between two acquisition functions: M-UCB and PR-M-UCB (number of starting points, number of gradient ascent iterations). The task is finding the largest animals given the names and the generative language model is GPT-3.5-turbo.} 
\label{fig.exp2}
    \Description{Two line charts compare the mean score of the selected prompt and optimization time against candidate-set size on the largest-animals task with GPT-3.5-turbo. The curves show M-UCB and three PR-M-UCB configurations.}
\end{figure}

\begin{figure}[htbp]
\centering
\begin{minipage}[t]{0.49\textwidth}
\centering
\includegraphics[width=\linewidth]{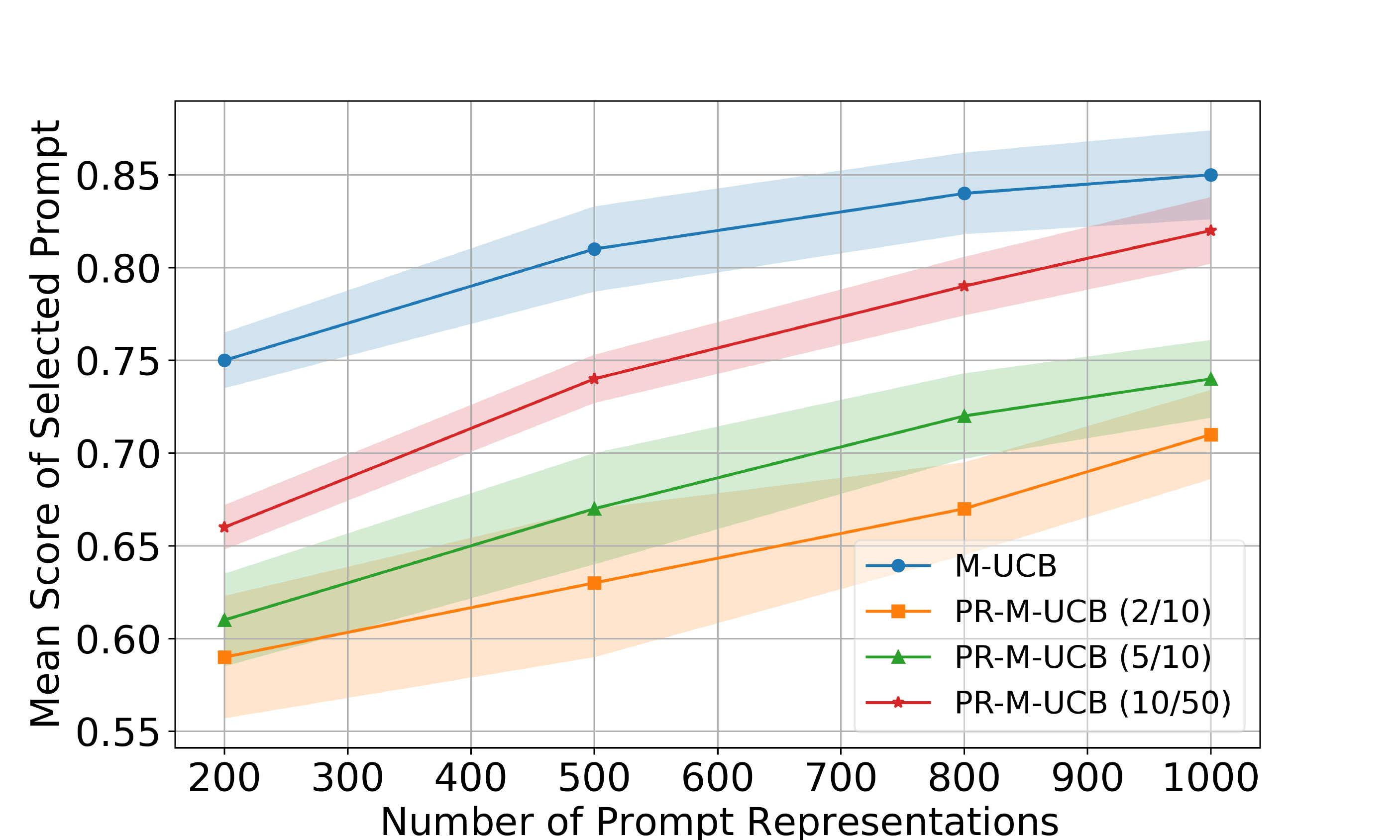}
\end{minipage}
\begin{minipage}[t]{0.49\textwidth}
\centering
\includegraphics[width=\linewidth]{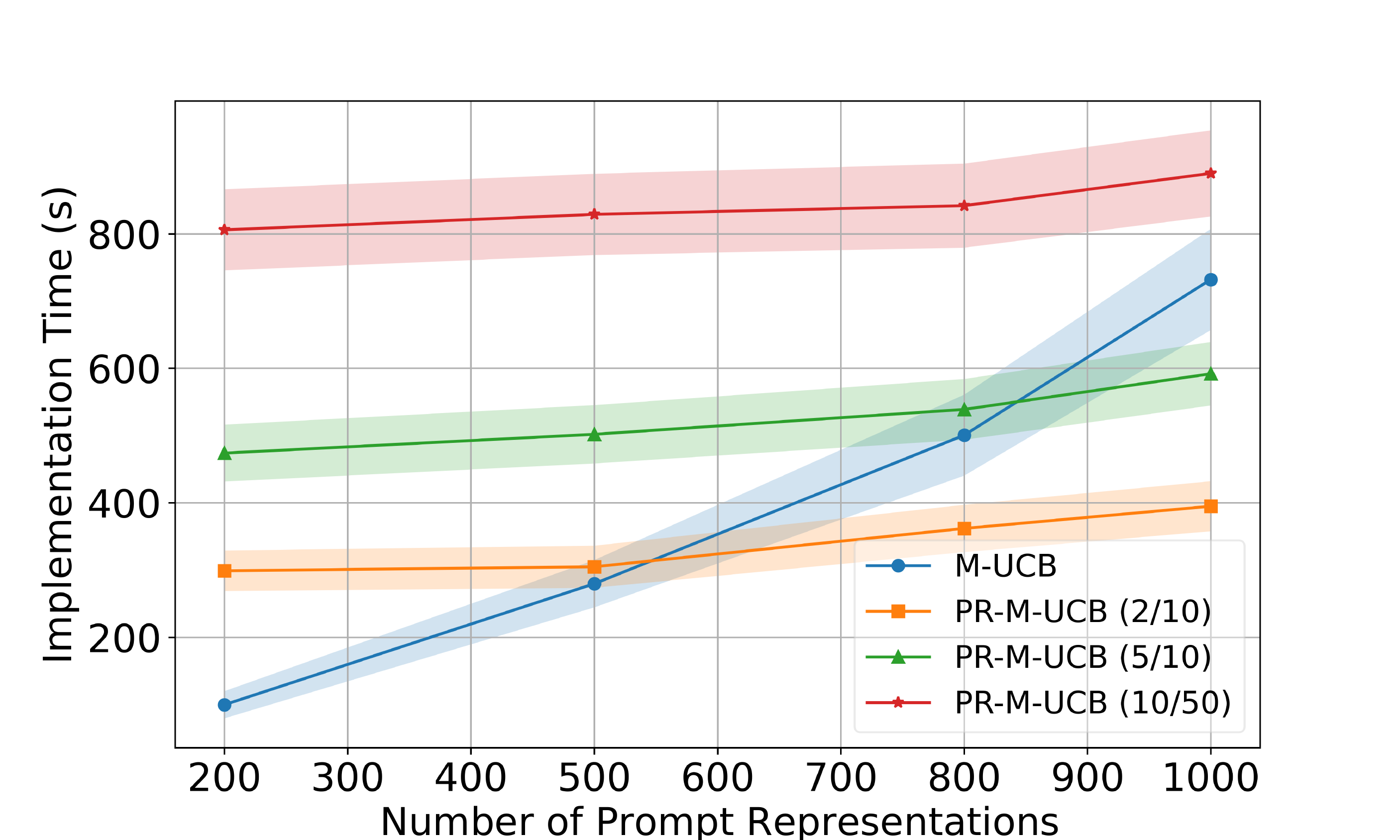}
\end{minipage}
\caption{Experimental results for comparison between two acquisition functions: M-UCB and PR-M-UCB (number of starting points, number of gradient ascent iterations). The task is word sorting and the generative language model is text-davinci-003.} 
\label{fig.exp22}
    \Description{Two line charts compare the mean score of the selected prompt and optimization time against candidate-set size on the word-sorting task with text-davinci-003. The curves show M-UCB and three PR-M-UCB configurations.}
\end{figure}

We consider different numbers of the prompt representations $N\in\left\{200,500,800,1000\right\}$. The total budget is fixed as $T=500$. We select BNN as the surrogate model and use variational inference (VI) to generate unknown parameters $\left\{\widehat{\bm{W}}_k\right\}_{k=1}^{100}$. In each set of experiments, the warm-up step uses $N_W=5$ prompts from the initial set with $R=3$ evaluations per prompt, as described in Appendix \ref{sec.warm}. We then construct a BNN model using these scores. The sequential evaluation step, the procedures associated with M-UCB and PR-M-UCB employ the same BNN model constructed in the warm-up step. Regarding the comparing metric, we record the mean score of the selected prompt when the total budget is used up. The mean score of the selected prompt is approximated by additionally evaluating it for 50 times. We also record the implementation time of the methods during the sequential evaluation step, which includes the time for updating the surrogate model, generating unknown parameters from the posterior distribution, and optimizing the acquisition function.

The experiment results presented in {Figure \ref{fig.exp2}} and {Figure \ref{fig.exp22}} provide following insights. First, M-UCB achieves higher scores than PR-M-UCB across all sets of experiments, since M-UCB evaluates the acquisition function at all prompt representations in each iteration. Second, the performance of PR-M-UCB improves as the number of iterations and starting points in the gradient ascent algorithm increase. Third, the implementation time of M-UCB increases almost linearly with the number of prompt representations. In contrast, the implementation time of PR-M-UCB is determined by the settings of the gradient ascent algorithm, specifically the number of iterations and starting points. As the number of prompt representations increases, the implementation time of PR-M-UCB does not increase significantly. Considering both the mean score of the selected prompt and implementation time, we recommend using M-UCB for a relatively small number of prompt representations, and switching to PR-M-UCB when the number of prompt representations is large.

\subsection{Two-stage Framework versus Direct Search in Latent Space}
\label{sec.e3}
Recall that a prompt in text form is transformed to a high-dimensional vector $X$ in the latent space $\tilde{\mathcal{X}}$, and we propose in Section \ref{sec.refinment} a projection stochastic kriging (PSK) model that provides an approximation from the latent vector to the mean score. In this section, we conduct numerical experiments to compare our proposed two-stage framework with the approach that directly evaluates and selects latent vectors for the prompt selection problem.

We here provide details of the direct search procedure in the latent space using PSK. As one of Gaussian process models, the PSK model is compatible with classical simulation optimization methods. In this set of experiments, we employ the acquisition function expected improvement (EI) (see \cite{frazier2018bayesian}), and regard every observation of the score as deterministic without noise. Specifically, in each iteration $t$, the latent vector to evaluate is selected by maximizing
\begin{equation}
\label{eq.ei}
     \operatorname{EI}_t(X)= \left(\widehat{\mu}\left(X\right)-v^{\star}_t\right) \Phi\left(\frac{\widehat{\mu}\left(X\right)-v^{\star}_t}{\sqrt{\widehat{\sigma^2}\left(X\right)}}\right)+\sqrt{\widehat{\sigma^2}\left(X\right)} \varphi\left(\frac{\widehat{\mu}\left(X\right)-v^{\star}_t}{\sqrt{\widehat{\sigma^2}\left(X\right)}}\right).
\end{equation}
Here $\Phi$ and $\varphi$ are the cumulative distribution function and the probability density function of the standard normal distribution; $v^{\star}_t$ is the highest observed score up to time $t$; $\widehat{\mu}(X)$ and $\widehat{\sigma^2}(X)$ denote the PSK predictor and the associated prediction uncertainty as described in (\ref{eq.pskpredictor}) and (\ref{eq.predictionuncertainty}). We denote the direct search in the latent space using the PSK model and the EI function as PSK-EI.

\begin{figure}[htbp]
\centering
\begin{minipage}[t]{0.49\textwidth}
\centering
\includegraphics[width=\linewidth]{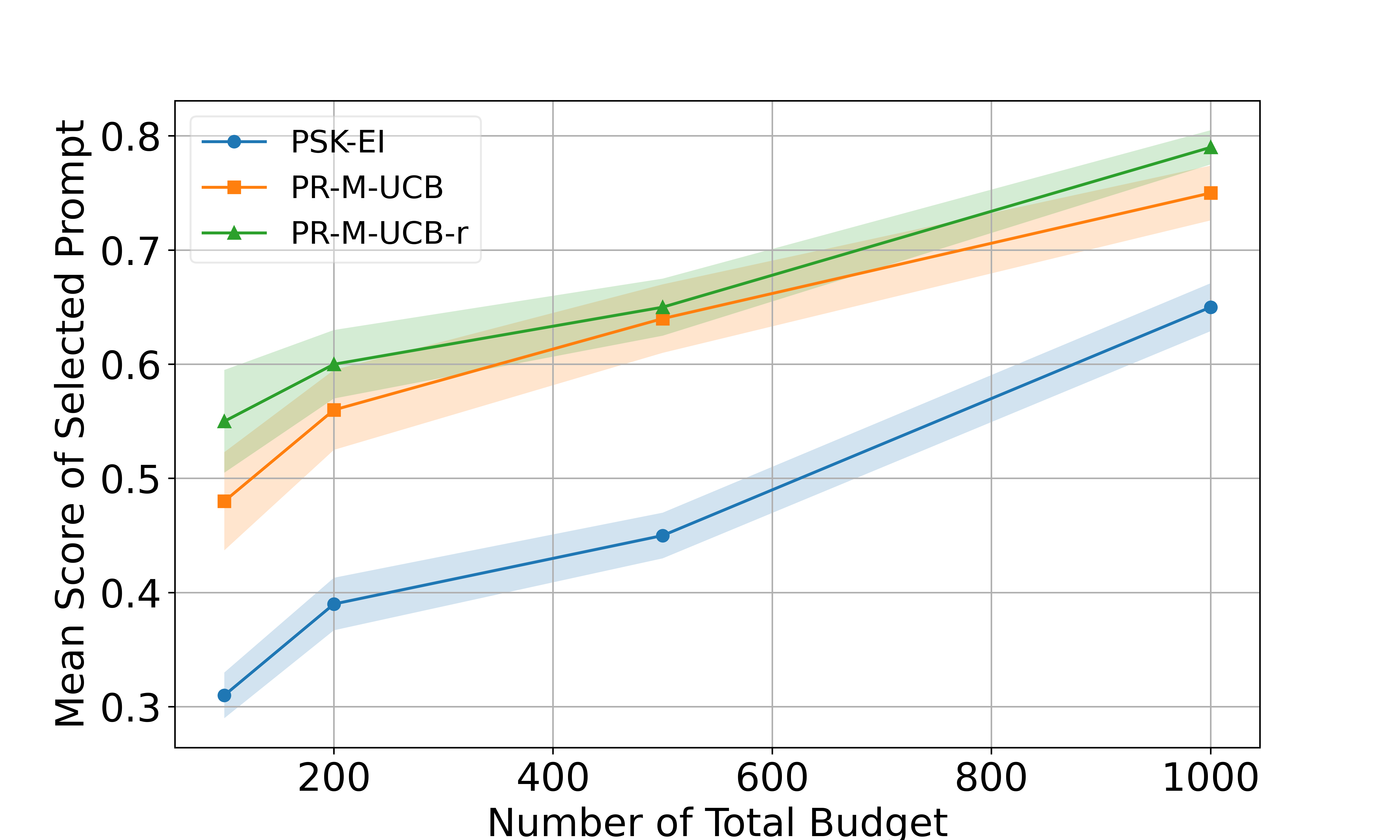}
\end{minipage}
\begin{minipage}[t]{0.49\textwidth}
\centering
\includegraphics[width=\linewidth]{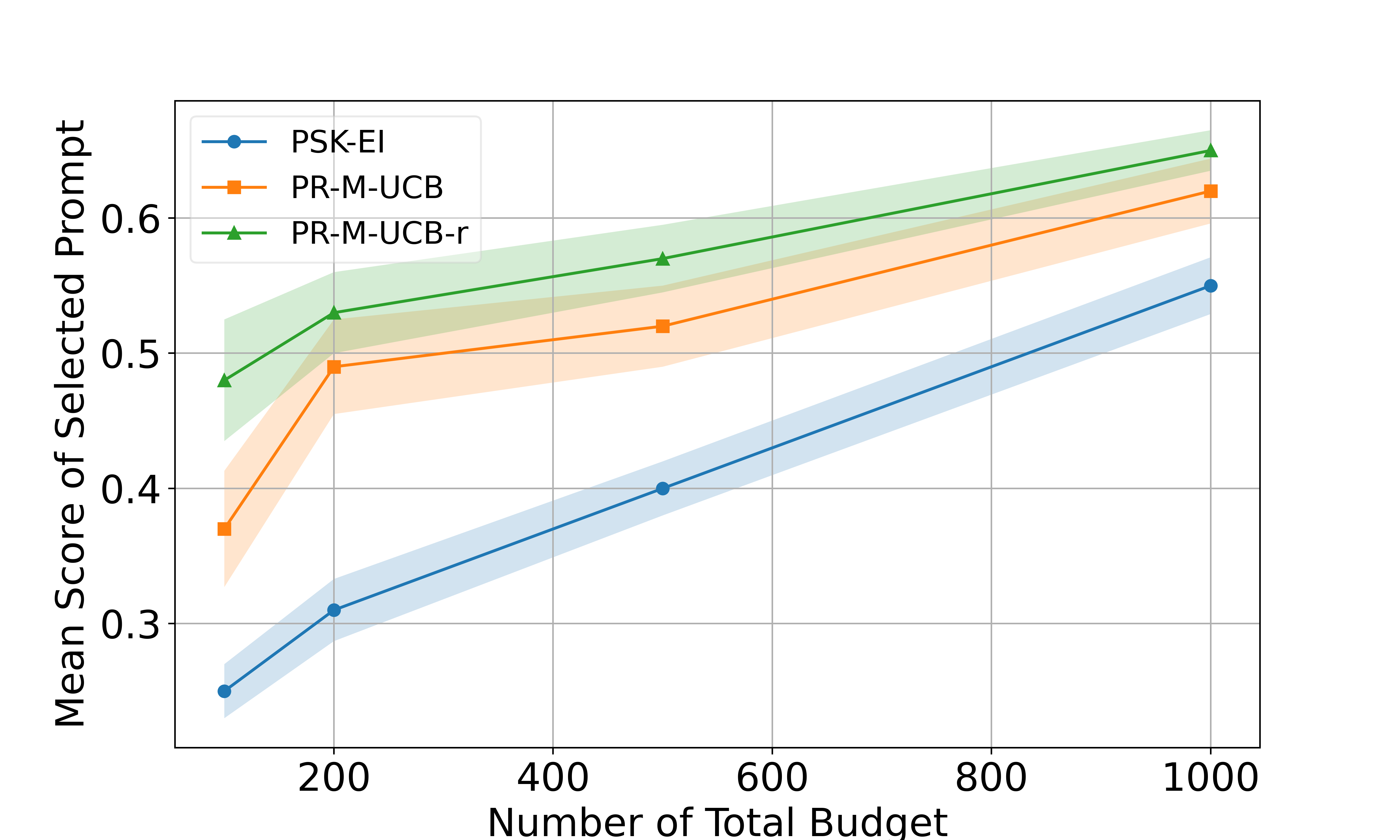}
\end{minipage}
\caption{Experimental results for comparison between the proposed two-stage framework and the direct search in the latent space with PSK. The pair of task and generative language model includes 1) counting objects with GPT-3.5-turbo (left) and 2) finding the rhymes with text-davinci-003 (right). PR-M-UCB-r uses 20 additional evaluation calls beyond $T$ for refinement.} %
\label{fig.exp3}
    \Description{Two line charts compare selected-prompt scores against evaluation budget for PSK-EI, PR-M-UCB, and PR-M-UCB-r. The tasks are counting objects with GPT-3.5-turbo and generating rhymes with text-davinci-003.}
\end{figure}

For the comparison in Figure \ref{fig.exp3}, we implement our proposed two-stage framework and employ the acquisition function PR-M-UCB. We consider different total budgets as $T\in\left\{50, 200, 500, 1000\right\}$. In terms of the search stage, we construct a set of $N=200$ prompt representations for all sets of experiments. In addition, we also consider the refinement procedure of our framework. That is, after the total budget is used up, we construct a PSK model using the observations and search within the latent space using the EI function (\ref{eq.ei}) with $20$ additional evaluations. We denote this method as PR-M-UCB-r in the experimental results. Regarding PSK-EI and PR-M-UCB, we record the mean score of the selected prompt when the total budget is used up. For PR-M-UCB-r, we record the mean score when the additional evaluations are used up as well. The mean score of the selected prompt is approximated by additionally evaluating it more 50 times.

The experimental results contained in {Figure \ref{fig.exp3}} provide following insights. First, in all sets of experiments, our proposed two-stage framework achieves higher scores than direct search in the high-dimensional latent space using PSK. This is because the PSK model requires sufficient observations to construct an accurate surrogate model. During the sequential evaluation without sufficient observations, the constructed PSK model often fails to provide a satisfactory approximation of the mean score for the latent vector, thereby not always selecting the prompts that are worth evaluating. Second, as the total budget increases, the improvement in direct search within the latent space is more significant than that of our two-stage framework. This is due to the increasingly accurate approximation provided by PSK, which depends on the number of observations. In comparison, since our two-stage framework is restricted to a finite set of constructed prompt representations, its improvement is relatively limited. Lastly, our two-stage framework benefits from the refinement step with additional evaluations. That is, the score of the selected prompt representation is enhanced after refinement using PSK. Therefore, when we have sufficient budgets to evaluate the prompt, leave some additional budget for a refinement procedure after the two-stage framework leads to a higher-score prompt. Section \ref{sec.expbbo} compares allocations between sequential evaluation and refinement under a fixed total budget.

At the end of this section, we provide a summary result of comparison between our two-stage framework and direct search in the latent space in {Figure \ref{fig:enter-label}}, which includes all six tasks implemented by the generative language model GPT-3.5-turbo. We also include the mean score of the random search as a baseline procedure. In each iteration, the random search procedure randomly selects a prompt representation from the constructed prompt representation set $z_n\in\mathcal{Z}$ to observe the score with equal probabilities. Both our proposed two-stage framework for prompt selection and the direct search procedure outperform the random search procedure with higher mean scores over six tasks. The experimental results also indicate the superiority of the proposed two-stage framework over the direct search in the latent space, and the selected prompts that achieve the highest mean score are included in Table \ref{tab:selectedprompts}.

\begin{figure}
    \centering
\begin{tikzpicture}
\begin{polaraxis}[
    xticklabels={word sorting, first letter, counting objects, rhymes, nums to verbal, largest animals},
    xtick={0,60,...,360},
    ymin=0, ymax=1, %
    ytick={0.2,0.4,0.6,0.8,1},
    grid=both,
    grid style={line width=1pt, draw=gray!50},
    major grid style={line width=1pt,draw=gray!50},
    minor tick num=1,
    every axis plot/.append style={line width=2pt, fill opacity=0.3},
    legend style={at={(1.05,1)}, anchor=north west},%
]

\addplot+[data cs=polar, mark=none, solid, fill=green] coordinates {
    (0,0.61) %
    (60,0.99) %
    (120,0.56) %
    (180,0.49) %
    (240,1) %
    (300,0.84) %
} -- cycle;
\addlegendentry{M-UCB}
\addplot+[data cs=polar, mark=none, solid, fill=blue] coordinates {
    (0,0.55)
    (60,0.81)
    (120,0.39)
    (180,0.31)
    (240,0.97)
    (300,0.79)
} -- cycle;\addlegendentry{PSK-EI}

\addplot+[data cs=polar, mark=none, solid, fill=orange] coordinates {
    (0,0.75)
    (60,0.98)
    (120,0.6)
    (180,0.53)
    (240,1)
    (300,0.89)
} -- cycle;\addlegendentry{M-UCB-r}

\addplot+[data cs=polar, mark=none, solid, fill=orange] coordinates {
    (0,0.41)
    (60,0.64)
    (120,0.33)
    (180,0.26)
    (240,0.75)
    (300,0.63)
} -- cycle;\addlegendentry{Random Search}

\end{polaraxis}
\label{fig.6}
\end{tikzpicture}
    \caption{The mean score of the selected prompts in six tasks. The generative language model used to implement the tasks is GPT-3.5-turbo. The budget before refinement is $T=500$; M-UCB-r uses 20 additional evaluation calls.}
    \label{fig:enter-label}
    \Description{Radar chart comparing M-UCB, PSK-EI, M-UCB-r, and random search on six task axes: word sorting, first letter, counting objects, rhymes, numbers to verbal, and largest animals.}
\end{figure}
In {Table \ref{tab:selectedprompts}}, we present the selected prompts using the algorithm M-UCB-r (described in Section \ref{sec.e3}) regarding all six tasks. The experimental setting is consistent with that in {Figure \ref{fig:enter-label}}.

\begin{table}[ht]
\caption{Selected prompts for six tasks using M-UCB-r.}
    \label{tab:selectedprompts}
\centering
\begin{tabularx}{\linewidth}{|>{\raggedright\arraybackslash}p{2cm}|>{\raggedright\arraybackslash}X|>{\raggedright\arraybackslash}X|}
\hline
\textbf{Task} & \textbf{GPT-3.5-turbo} & \textbf{text-davinci-003} \\ \hline
word sorting & Sort the following word list in alphabetic order. & Sort the words in alphabet order.\\
\hline
first letter & Return the first letter of the selected word. &Select the initial letter in the word.\\
\hline
counting objects & Count the total number of mentioned products. & Estimate the total quantity for all products.\\
\hline
rhymes &Choose the word with the closest pronunciation. &Select the word with most similar pronunciation.\\
\hline
nums to verbal & Turn number list to word list. &  Turn number list to word list. \\\hline
        largest animals & Choose the larger animal from the following animals. & Select the largest animals in size.  \\
\hline

\end{tabularx}
\end{table}

\end{document}